\pdfoutput=1

\documentclass[10pt,twocolumn,letterpaper]{article}

\usepackage[pagenumbers]{cvpr}%

\newcommand{\papertitle}{How Does Geometry Enter Generated Motion?}

\usepackage{siunitx}
\usepackage{colortbl}

\input{supp/style_supp}

\definecolor{cvprblue}{rgb}{0.21,0.49,0.74}
\usepackage[pagebackref,breaklinks,colorlinks,allcolors=cvprblue]{hyperref}
\hypersetup{pdftitle={\papertitle}, bookmarksnumbered=true}

\title{\vspace*{-24pt}\papertitle}

\newcommand{\paperauthors}{%
  Weihan Li\textsuperscript{1}\qquad
  Junhao Wu\textsuperscript{2}\qquad
  Yuhan Song\textsuperscript{1}\qquad
  Xiaofeng Lin\textsuperscript{1}\qquad
  Xinlei Chen\textsuperscript{3}\\[3pt]
  {\normalsize
  \textsuperscript{1}The University of Tokyo\qquad
  \textsuperscript{2}RWTH Aachen University\qquad
  \textsuperscript{3}Harbin Institute of Technology, Shenzhen}\\[3pt]
  {\tt\small liweihan1107@g.ecc.u-tokyo.ac.jp}%
}
\author{\paperauthors}
\hypersetup{pdfauthor={Weihan Li, Junhao Wu, Yuhan Song, Xiaofeng Lin, Xinlei Chen}}

\apptocmd{\bibsection}{\pdfbookmark[1]{\refname}{references}}{}{}

\begin{document}
\maketitle

\begin{abstract}
Under a fixed physical law, the visible geometry of a scene determines how motion must change.
We ask how video generators realize this relationship.
We fix the law and the initial state and change only the geometry drawn in the first frame, within matched families of tracks and deflectors, and compare each generated trajectory with the simulator prediction for that geometry.
Paired interventions change one thing at a time: a local bump, the height of a barrier, the words of the prompt, the length of the clip.
Across nine image-to-video models, geometry is preserved and shapes the motion: the speed of the ball follows the drawn undulation of a track.
A physical state would carry this response forward, and here the generated motion parts from the law.
The mean slope barely accelerates the ball, successive contacts fail to compose through a consistent state, an edit ahead of the ball alters its motion before it arrives, and the ball climbs over barriers higher than its release point.
Two global conditions organize the global trajectory: text strongly controls the destination, while clip length strongly controls timing in the open-weight models tested.
The pattern persists with photographed first frames.
Current video generation thus behaves as geometry-conditioned motion synthesis whose evolution of state differs systematically from that of a fixed physical law.
\newline The project page is available at\newline{\small\url{https://liweihan1107.github.io/samelaw/}}.
\end{abstract}

\section{Introduction}
\label{sec:intro}

\begin{figure}[t]
  \centering
  \includegraphics{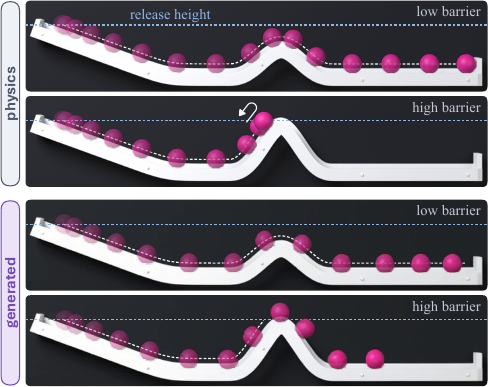}
  \caption{\textbf{Same law, different geometry.}
  One ball, one law, one starting point; only the height of a barrier changes.
  In the simulated motion (top) the ball crosses the lower barrier and turns back at the one that rises above its release height (dashed line).
  In the generated motion (bottom; MiniMax-H3, one seed) it crosses both, as it does with every seed.}
  \label{fig:teaser}
\end{figure}

\begin{figure*}[t]
  \centering
  \includegraphics{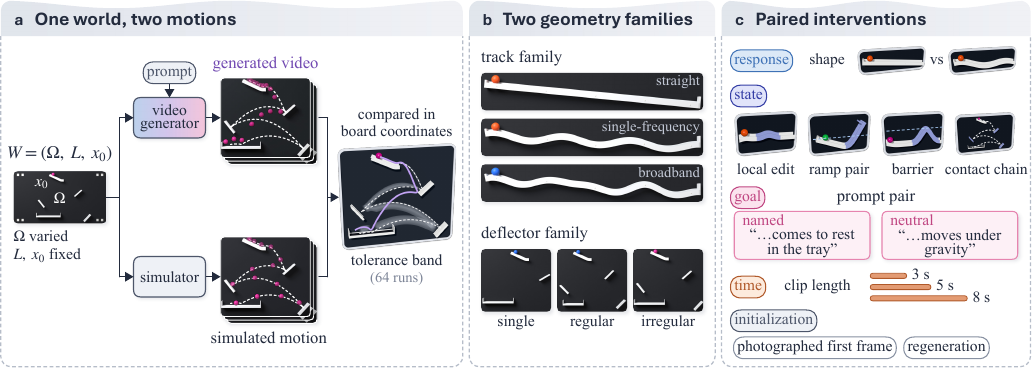}
  \caption{\textbf{Study design.}
  \emph{Left:} a world $W=(\Omega, L, x_0)$ is rendered as a first frame; the video generator produces a video and the simulator the physical motion with its tolerance band; both are compared in the coordinates of the board.
  \emph{Middle:} two families of geometry with matched global statistics.
  \emph{Right:} paired interventions, each changing one thing: shape (\cref{sec:response}); local edit, ramp pair, barrier and contact chain (\cref{sec:state}); prompt pair and clip length (\cref{sec:goal}); photographed first frame and regeneration (\cref{sec:robust}).}
  \label{fig:framework}
\end{figure*}

A ball on a track speeds up where the track falls and slows where it rises.
Tilt a plate in its path and it leaves in a different direction.
Under a fixed physical law and a fixed initial state, the visible geometry of a scene determines how the motion must change, and it does so through the state of the ball: each piece of geometry acts when the ball reaches it, and what it does depends on the speed and direction the ball brings.
This gives a direct test for a video generator.
When the geometry in the first frame changes, does the generated motion change in the way the same law requires?

The evaluation of physics in generated video has become steadily more quantitative.
Benchmarks first asked whether generated clips are physically plausible~\cite{bansal2025videophy,meng2025phygenbench,motamed2026physicsiq}.
Later work tests equations of motion and conserved quantities on trajectories extracted from the videos~\cite{tragoudaras2026morpheus}, compares them with simulator rollouts from the same first frame~\cite{jain2026rigidbench}, and reads relational quantities that need no calibration of scale or time~\cite{thozhiyoor2026slower,thozhiyoor2026principia}.
Controlled studies now vary one factor at a time: a physical parameter conveyed by the prompt or the conditioning frames~\cite{khanbayov2026physweep}, the length of a chain of dependent collisions~\cite{diazchao2026seriality}, or a counterfactual branch under shared noise~\cite{ma2026twinrollouts}.
Together, these studies make physical correctness measurable.
They leave open a complementary question: how visible geometry enters the generated dynamics when the law is held fixed.

We study this question through \emph{same law, different geometry}.
A world is a triple $W=(\Omega, L, x_0)$ of visible geometry, local law and initial state.
We fix $L$ and $x_0$ and vary $\Omega$ within families whose global statistics are matched, so that two first frames differ only in shape, and a simulator gives the physical motion for each.
Two families cover continuous and discrete geometry: a ball rolling on a track with a drawn undulation, and a ball bouncing through flat deflectors into a tray (\cref{fig:framework}).
Around them, paired interventions change one thing at a time: a local bump on the track, the height of a barrier, the words of the prompt, the length of the clip.
We generate more than 5{,}000 videos with nine image-to-video models, six open-weight and three closed, and track the ball in the coordinates of the board.
The design separates three questions that a plausibility score merges: whether the geometry is preserved, whether it changes the motion, and whether the change follows one local dynamics.

Geometry enters the motion.
Across all nine models, the drawn track is largely preserved and the ball remains constrained to it.
The speed of the ball rises and falls with the drawn undulation, at 0.5 to 1.2 of the physical gain and, in eight models, ahead of the physical phase by 15 to 41 degrees.
The mean slope, which supplies most of the speed in the physical motion, contributes little: the ball settles to a preferred speed, and the undulation modulates that speed.

A physical state would carry this response forward, and three tests ask for it.
Composition: with three deflectors, the physical order of contacts appears in at most a quarter of a model's videos, and each contact deviates on its own.
Locality: a bump changes the speed at the bump in one model, at a fifth of the physical size, while in three models an edit ahead of the ball alters its path before it arrives.
Energy: the ball rises above its release height in 63 of 72 videos and crosses 43 of 49 barriers that are physically impassable (\cref{fig:teaser}).
The response on the track is therefore a modulation of motion by shape, and the role of state is taken by something else.

Two global conditions instead organize the global trajectory.
The outcome named in the prompt controls the destination.
When the prompt says that the ball comes to rest in the tray, it does so in nearly every video, more often than its own motion, continued by the simulator, would bring it there.
With the same first frame and a prompt that leaves the outcome unnamed, arrival falls by 33 to 96 percentage points in all eight models tested, and the drawn tray by itself barely attracts the ball.
Clip length controls the timing in the open-weight models tested.
From 3 to 8 seconds the same event stretches almost in proportion, and one model reaches the end of the track at nine tenths of the clip at every length.

Taken together, current video generators use the visible geometry to shape motion locally, the text to choose its outcome, and the clip length to pace it.
A fixed law prescribes a different organization: a state that is updated where the ball is, composed across events and bounded by energy.
The pattern persists when the rendered first frames are replaced by photographs of the built scenes, and a regeneration probe returns each model to its own version of the motion.
These results motivate evaluating physics through how controlled changes propagate through generated motion, beyond whether the resulting clip appears physically plausible.

\section{Related Work}
\label{sec:related}

\paragraph{Physics in generated video.}
Benchmarks first scored generated clips for physical plausibility, with human or vision-language judges~\cite{bansal2025videophy,meng2025phygenbench} or against real footage of the same event~\cite{motamed2026physicsiq,radsch2026physicsiqverified}.
Trajectory-based evaluation tests the motion of tracked objects against equations of motion and conserved quantities~\cite{tragoudaras2026morpheus}, and relational readings remove the dependence on unknown scale and frame rate~\cite{thozhiyoor2026slower,thozhiyoor2026principia}.
Simulator rollouts from the same first frame separate errors of trajectory, contact and appearance~\cite{jain2026rigidbench}.
These studies make the gap between generated and physical motion measurable for a single scene.
We study how that gap depends on the visible geometry when the law and the initial state are held fixed.

\paragraph{Controlled response and composition.}
One-factor interventions reveal what a generator responds to.
Physical parameters conveyed by the prompt or by the conditioning frames barely move the generated dynamics~\cite{khanbayov2026physweep,zhu2026multiecho}.
Video diffusion models trained on hard-sphere dynamics degrade with the length of a chain of dependent collisions~\cite{diazchao2026seriality}.
Noise-coupled factual and counterfactual branches define locality for interactive world models~\cite{ma2026twinrollouts}, and single-factor pairs show that video-to-audio models take physics from the caption more than from the video~\cite{li2026flatsounds}.
This makes the response to a controlled change an object of measurement in its own right.
Our controlled variable is the shape of the tracks and deflectors, which appear in earlier suites as fixed stimuli~\cite{motamed2026physicsiq,sadeghi2026calibench}, and we ask of the response whether it composes across contacts, acts where the ball is, and respects energy.

\paragraph{What organizes generated motion.}
Generated motion carries an organization of its own.
Falls are slower than physical ones and the implied time scale is unstable~\cite{thozhiyoor2026slower,gao2026pulse}, trajectory spectra are weak at nonzero temporal frequencies~\cite{ni2026motionspec}, motion is planned in the early denoising steps~\cite{li2026violate} and its phase erodes during later refinement, so that two steps can be more physically consistent than fifty~\cite{han2026phaselock}, and events start late relative to their cause~\cite{arriagasantana2026foresight}.
We assign parts of the trajectory to conditions: the drawn geometry shapes the local speed, the outcome named in the prompt controls the destination, and the clip length controls the timing.
The contrast between simulating a state and applying a regularity has a long history in the study of human intuitive physics~\cite{battaglia2013simulation,mccloskey1980curvilinear,kozhevnikov2001impetus}, and the preferred-speed motion we observe resembles the impetus-like regularities described there.

\section{Same Law, Different Geometry}
\label{sec:setup}

\paragraph{Object.}
A world is $W=(\Omega, L, x_0)$: visible geometry, a local law and an initial state.
Physics defines a flow $\Phi_\Omega(x_0)$.
An image-to-video model defines a distribution over trajectories given a first frame that shows $\Omega$ and $x_0$, and a prompt.
We hold $L$ and $x_0$ fixed and vary $\Omega$.
Within a family the variants share their global statistics, such as the length of a track and its physical transit time or the drop and the end points of a deflector course, and the prompt is identical, so the variant is conveyed by the first frame alone.
All scenes share one visual world: a dark board, white geometry, one coloured ball, a fixed frontal camera and four corner markers that give the scale.

\paragraph{Two families.}
In the \emph{track family} a ball rolls on a twin-rail track whose height follows a mean slope plus an undulation.
The levels are a straight track, a single frequency of 3 or 4 cycles per metre at 15\,mm amplitude, and a broadband profile with three components of 10\,mm, matched in length and physical transit time (40 scenes).
In the \emph{deflector family} the ball leaves a short launch ramp, strikes flat deflectors and lands in a tray.
The levels are one deflector at four angles, three deflectors in a regular zigzag, and three with irregular angles and spacing, matched in drop, start, tray position and mean angle (24 scenes).
Smooth tracks and flat contacts keep the physical trajectory stable: the exit direction depends on the visible normal and the irregularity lies in the layout of the geometry.
The reference is therefore one trajectory with a narrow band.

\paragraph{Paired interventions.}
Each intervention changes one thing between generations that share everything else, including the seed where the model allows it.
\emph{Shape}: the levels of a family.
\emph{Local edit}: a 15\,mm bump or dip on one section of a track, with an edit of equal size placed off the track as a null.
\emph{Ramp pair}: a descent followed by an ascent of 0 to 60 degrees.
\emph{Barrier}: a hump whose height is 0.70 to 1.20 of the release drop.
\emph{Prompt pair}: the same first frame with a prompt that names the outcome (the ball comes to rest in the tray) or a neutral one.
\emph{Clip length}: 3, 5 and 8\,s.
\emph{Photographed first frame}: a photograph of the built scene in place of the render.
\emph{Regeneration}: a source video is partially noised and regenerated~\cite{meng2022sdedit}.

\begin{table*}[t]
  \centering
  \caption{\textbf{Nine models under the same law.}
  Medians over scenes; the first row is the physical oracle, the simulator read through the same measurement chain.
  Intact: share of intact videos in the track and the deflector family.
  Gain and phase are read on tracks with 3 cycles per metre, the trend ratio on straight tracks, $e_n$ at the first contact, the order of contacts on three deflectors, and the height ratio on ramp pairs.
  Tray arrival is given under the named and the neutral prompt, and $\gamma$ is the exponent of tempo against clip length ($\dagger$: twice as fast at 3\,s, unchanged from 5 to 8\,s).
  A dash marks a reading that was not taken for that model.}
  \label{tab:models}
  \small
  \setlength{\tabcolsep}{4.5pt}
  \definecolor{tabInteg}{HTML}{EEF1F6}
\definecolor{tabResp}{HTML}{DCE9F7}
\definecolor{tabState}{HTML}{E1E4FA}
\definecolor{tabGoal}{HTML}{FBE3EE}
\definecolor{tabTime}{HTML}{FDEBDD}
\definecolor{tabOracle}{HTML}{F0F2F5}
\definecolor{tabFaint}{HTML}{8790A1}
\providecommand*{\tabglyph}[1]{\raisebox{-0.14ex}{\includegraphics[height=0.72em]{figs/glyphs/#1}}\hspace{0.34em}}
\providecommand*{\tabpair}[2]{\makebox[1.6em][r]{#1}${}\to{}$\makebox[1.1em][r]{#2}}
\setlength{\aboverulesep}{0pt}\setlength{\belowrulesep}{0pt}\setlength{\extrarowheight}{1.6pt}
\sisetup{table-number-alignment=center, table-text-alignment=center}
\begin{tabular}{l S[table-format=3.0] S[table-format=3.0] S[table-format=1.2] S[table-format=+2.0] S[table-format=-1.2] S[table-format=1.2] S[table-format=3.0] S[table-format=1.2] c S[table-format=1.2]}
\toprule
 & \multicolumn{2}{>{\columncolor{tabInteg}}c}{Intact (\%)} & \multicolumn{3}{>{\columncolor{tabResp}}c}{Response} & \multicolumn{3}{>{\columncolor{tabState}}c}{State} & \multicolumn{1}{>{\columncolor{tabGoal}}c}{Goal: tray (\%)} & \multicolumn{1}{>{\columncolor{tabTime}}c}{Time} \\
Model & \multicolumn{1}{>{\columncolor{tabInteg}}c}{track} & \multicolumn{1}{>{\columncolor{tabInteg}}c}{deflector} & \multicolumn{1}{>{\columncolor{tabResp}}c}{gain $|H|$} & \multicolumn{1}{>{\columncolor{tabResp}}c}{phase ($^\circ$)} & \multicolumn{1}{>{\columncolor{tabResp}}c}{trend} & \multicolumn{1}{>{\columncolor{tabState}}c}{$e_n$} & \multicolumn{1}{>{\columncolor{tabState}}c}{order (\%)} & \multicolumn{1}{>{\columncolor{tabState}}c}{height} & \multicolumn{1}{>{\columncolor{tabGoal}}c}{named $\to$ neutral} & \multicolumn{1}{>{\columncolor{tabTime}}c}{$\gamma$} \\
\midrule
\rowcolor{tabOracle}\tabglyph{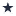}Physical oracle & {--} & {--} & 1.00 & 0 & 1.00 & 0.85 & 100 & 0.96 & \tabpair{91}{91} & 0 \\
\specialrule{0.8pt}{0pt}{0pt}
\multicolumn{1}{l}{\footnotesize\itshape Open-weight} & & & & & & & & & & \\[-1pt]
\tabglyph{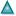}LTX-2.3 & 87 & 1 & 1.24 & +38 & 0.15 & 0.35 & 0 & 1.23 & \tabpair{100}{25} & 0.85 \\
\tabglyph{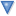}LTX-2.5~(distilled) & 100 & 75 & 0.97 & +35 & -0.10 & 0.21 & 0 & 2.36 & \tabpair{100}{56} & 0.91 \\
\tabglyph{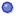}LTX-2.5 & 83 & 6 & 0.98 & +36 & 0.02 & {--} & {--} & 1.60 & -- & {--} \\
\tabglyph{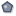}Cosmos 3 Nano & 88 & 67 & 0.57 & +41 & -0.02 & 0.10 & 0 & 1.15 & \tabpair{100}{7} & {--} \\
\tabglyph{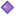}MiniMax-H3 & 100 & 88 & 0.73 & -3 & 0.22 & 0.45 & 0 & 1.56 & \tabpair{98}{2} & 1.04 \\
\tabglyph{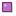}Wan 2.2~(A14B) & 96 & 81 & 0.54 & +15 & 0.22 & 0.15 & 0 & 1.76 & \tabpair{47}{3} & 0.89{\rlap{$^{\dagger}$}} \\
\arrayrulecolor{tabFaint}\specialrule{0.35pt}{0pt}{0pt}\arrayrulecolor{black}
\multicolumn{1}{l}{\footnotesize\itshape Closed} & & & & & & & & & & \\[-1pt]
\tabglyph{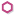}Wan 3.0 & 100 & 88 & 1.11 & +27 & 0.25 & 0.67 & 12 & 1.94 & \tabpair{100}{67} & {--} \\
\tabglyph{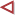}Seedance 2.0 & 100 & 72 & 0.60 & +19 & 0.62 & 0.74 & 24 & 1.66 & \tabpair{100}{33} & {--} \\
\tabglyph{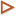}Seedance 2.0 mini & 100 & 86 & 0.59 & +36 & 0.65 & 0.67 & 14 & 1.82 & \tabpair{100}{33} & {--} \\
\bottomrule
\end{tabular}

\end{table*}

\paragraph{Reference and readings.}
A simulator integrates $L$ on the drawn geometry, and a 64-member ensemble over material parameters and the starting point gives a tolerance band.
We track the ball into board coordinates.
On a track, the ball's speed along the arc is regressed on the height profile, split into mean slope and undulation.
This yields the \emph{trend ratio}, the acceleration along the mean slope relative to the acceleration implied by the ball's own transit time, and a complex \emph{gain} $H$ for the undulation.
Both are normalized by the \emph{tempo} $\alpha$, the ratio of generated to physical transit time, so that any time-rescaled Newtonian motion has trend ratio 1, $|H|=1$ and phase 0.
At a deflector we compare the exit direction with the simulator's prediction for the generated incoming direction, and fit the apparent normal \emph{restitution} $e_n$.
We also read the order of contacts, whether the ball comes to rest in the tray, and, on ramps, the \emph{height ratio} of the highest point reached to the release height.
Exact renders passed through the same chain give each reading a measurement floor, and differences are read against the floor and the band.

\paragraph{Models.}
We study nine image-to-video models: six open-weight models (LTX-2.3, LTX-2.5 and its distilled version, Wan~2.2, MiniMax-H3, Cosmos~3 Nano) and three closed models accessed through their APIs (Seedance~2.0, Seedance~2.0 mini, Wan~3.0)~\cite{hacohen2026ltx2,lightricks2026ltx23,lightricks2026ltx25,wan2025wan,wan2025wan22,minimax2026h3,nvidia2026cosmos3,nvidia2026cosmos3nano,seedance2026seedance2,volcengine2026seedancemini,alibaba2026wan3}.
Unless otherwise stated, each model generates 5\,s clips at its native resolution and frame rate, with three seeds per scene.
Interventions that pair videos by seed use the open-weight models.
A video is \emph{intact} when the scene stays what the first frame shows: one ball, static apparatus, steady camera.
It stays \emph{readable} when a part changes only after the ball has left it.
Dynamics are read on readable videos, and the intact share is reported with them.

\section{Results}
\label{sec:results}

Each subsection takes up the question left by the one before.
\Cref{tab:models} collects the readings per model, and the supplement reports every reading with its interval, floor and denominator.

\subsection{How does visible geometry modulate motion?}
\label{sec:response}

\begin{figure*}[t]
  \centering
  \includegraphics{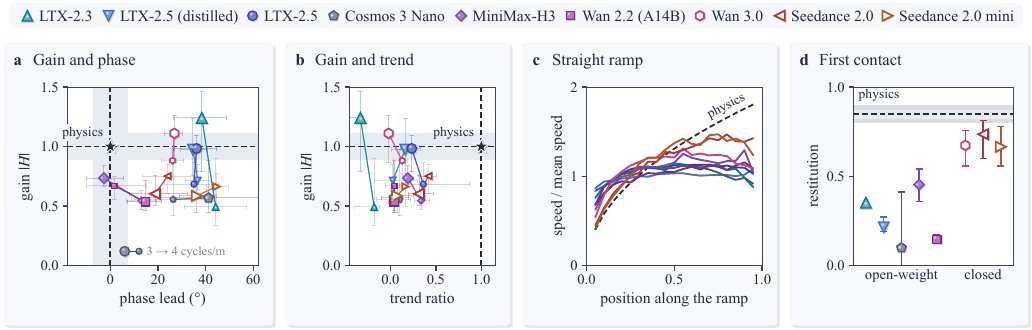}
  \caption{\textbf{Geometry modulates the motion.}
  (a) Gain and phase of the speed's response to the drawn undulation; physics sits at gain 1 and phase 0.
  (b) The same gain against the trend ratio: the undulation is followed while the mean slope adds little.
  (c) Speed along a straight ramp, normalized: physics accelerates, the models hold a preferred speed.
  (d) Restitution at the first contact; the prescribed value is 0.85.
  Filled markers are open-weight models and hollow markers closed ones; grey bands are measurement floors.}
  \label{fig:response}
\end{figure*}

\paragraph{The drawn geometry is preserved.}
In the track family, 83--100\% of each model's videos are intact.
In them the drawn undulation keeps its amplitude to within 1\% and the ball stays on the track.
In the deflector family, seven models keep 67--88\% of their videos intact, with the deflectors within $0.2^\circ$ of their drawn angle.
LTX-2.3 and LTX-2.5 move parts of the deflector scenes or the camera, and enter the analysis through the track family.
Whatever separates generated from physical motion below arises with the geometry in place.

\paragraph{The speed follows the drawn shape.}
On undulating tracks the speed of the generated ball rises and falls with the drawn profile (\cref{fig:response}a).
Relative to each model's own tempo, the gain at 3 cycles per metre is 0.54--1.24 of the physical one.
Eight models lead the physical phase by $15^\circ$--$41^\circ$, at least twice the floor of $6^\circ$--$7.5^\circ$, and MiniMax-H3 responds in phase.
The response repeats across seeds.
When tracks share their first 0.4\,m and differ afterwards, the response in the later section follows the phase drawn there, with a slope of 0.80--1.07 in six models.
The models read the undulation from the first frame, and the motion carries it where it is drawn.

\paragraph{The mean slope contributes little.}
In the physical motion the mean slope supplies most of the ball's speed.
In the generated motion the acceleration along the straight track is at most 0.65 of what the ball's own transit time implies, and below 0.3 in seven models (\cref{tab:models}); on the undulating tracks it is at most 0.42 (\cref{fig:response}b).
On a straight ramp the distance covered grows with time with an exponent of 0.5--1.6, where uniform acceleration gives 2 and a constant speed gives 1 (\cref{fig:response}c).
A ball that relaxes toward a preferred speed reproduces this signature as a whole: a flat trend, a reduced gain and a phase lead.
Among ten candidate mechanisms passed through the same camera and estimator, four models are best described by a preferred speed that varies with the local slope and five by a constant preferred speed with temporal smoothing.
Time-rescaled Newtonian motion fits none of the nine (supplement).

\paragraph{Closed models bounce, open-weight models slide.}
The deflector family gives the discrete counterpart.
At the first contact the three closed models bounce, with a restitution of 0.67--0.74 against the prescribed 0.85 and an exit direction $16^\circ$--$35^\circ$ from the physical one.
MiniMax-H3 bounces in part, with a restitution of 0.45.
Cosmos~3 Nano, the distilled LTX-2.5 and Wan~2.2 slide along the deflector, with a restitution of 0.10--0.21 and exits $57^\circ$--$78^\circ$ away (\cref{fig:response}d).
In both groups the ball meets the drawn surface where it is drawn.

\subsection{Does the modulation form a local dynamics?}
\label{sec:state}

A state-mediated physical dynamics makes three testable predictions here: successive events compose through the state, only geometry the ball has reached can affect it, and its evolution obeys the energy bound.

\begin{figure*}[t]
  \centering
  \includegraphics{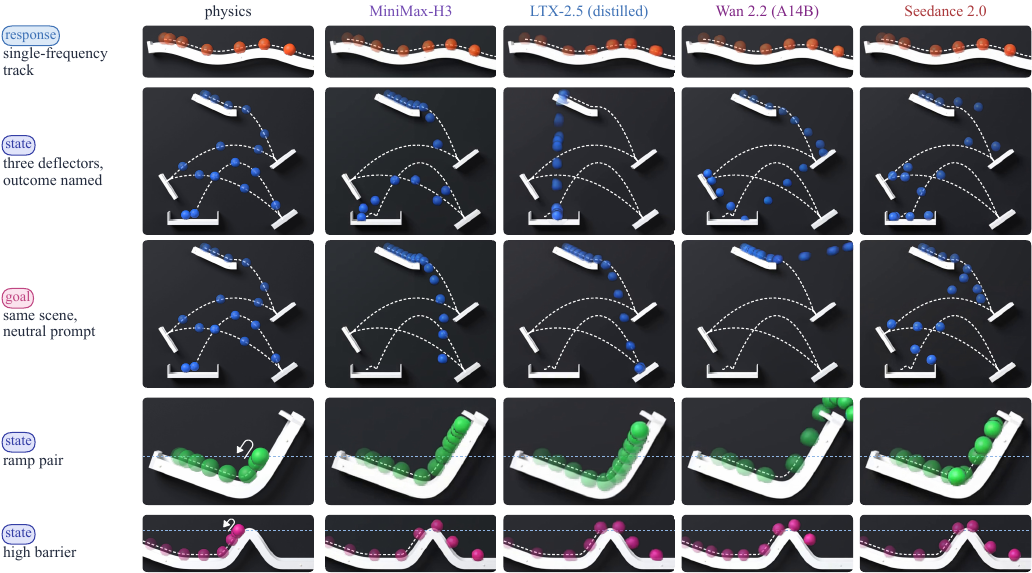}
  \caption{\textbf{Cases.}
  Five scenes (rows) in the simulated motion and in four models (columns).
  The ball is drawn at equal time steps, from light to full, so that its spacing shows the speed; the white dashed curve is the physical path and the blue dashed line the release height.
  The third row repeats the scene of the second with a prompt that leaves the outcome unnamed.
  Each model contributes the video closest to its median reading on that scene.}
  \label{fig:cases}
\end{figure*}

\begin{figure}[t]
  \centering
  \includegraphics{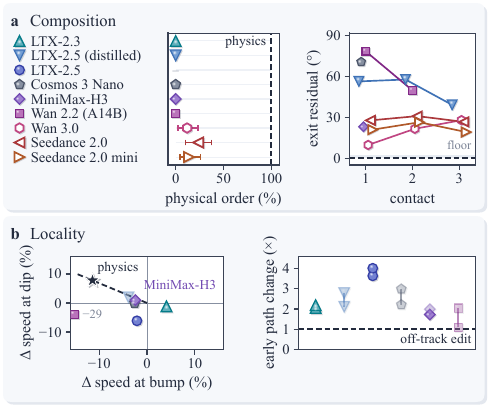}
  \caption{\textbf{Composition and locality.}
  (a) Share of three-deflector videos with the physical order of contacts, and exit residual by contact index; the rows name the models, and a dash marks a reading that was not taken.
  (b) Change of speed at a bump or dip against physics, and change of the path before the ball reaches the edit, relative to an edit of equal size off the track; faded marks do not separate from the off-track edit.}
  \label{fig:state}
\end{figure}

\paragraph{Successive contacts fail to compose.}
With three deflectors the simulator's ensemble keeps a single order of contacts.
The closed models reproduce it in 12--24\% of their videos and the open-weight models in none (\cref{fig:state}a).
In MiniMax-H3, Wan~2.2 and Cosmos~3 Nano the ball seldom reaches a second contact: it slides off the first deflector and drops.
In the two Seedance models the exit direction is off by $19^\circ$--$31^\circ$ at the first, the second and the third contact alike, so the chain is lost through independent local deviations, and in Wan~3.0 the deviation grows from $10^\circ$ to $28^\circ$.
A contact law calibrated on each model's single-deflector videos improves the prediction of the three-deflector scenes for one model only, so the deviations do not reduce to a consistent wrong material.

\paragraph{Edits act ahead of the ball.}
We add a 15\,mm bump or dip to one section of a track and generate with the same seed as for the unedited track.
Physics changes the speed at the edit by $-11\%$ and $+8\%$ and leaves the motion before it unchanged.
MiniMax-H3 slows at the bump by 2.4\%, a fifth of the physical change, and reaches the edit at the unchanged time.
In LTX-2.5 the ball reaches either edit 3--6\% sooner and slows at the dip, where physics speeds up.
In three open-weight models the path before the ball reaches the edit changes 1.7--4.0 times as much as under an edit of equal size placed off the track, beyond the seed-matched null (\cref{fig:state}b).
The shape ahead of the ball takes part in the motion from the start.

\begin{figure}[t]
  \centering
  \includegraphics{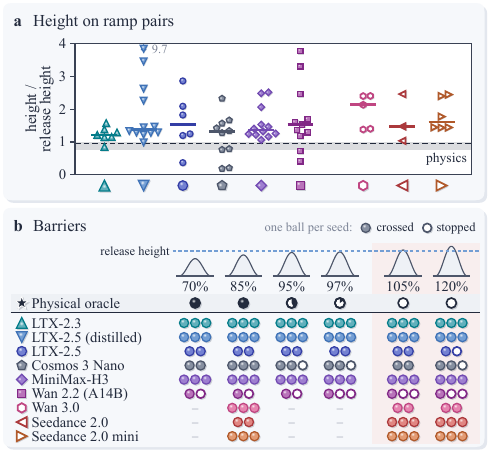}
  \caption{\textbf{Energy.}
  (a) Height reached over release height on ramp pairs, one mark per video with bars at the median; physics lies in the band below 1, and the glyphs under the swarms name the models in the order of the rows of (b).
  (b) Barriers, one ball per seed, solid where the ball crossed and hollow where it stopped; columns give the barrier height in percent of the release height, tinted above it.
  The physical oracle crosses below the release height and none above; its ball is filled by the share of the 64-member ensemble that crosses.}
  \label{fig:energy}
\end{figure}

\paragraph{The ball climbs past its release height.}
On a ramp pair the physical ball turns back at 0.93--0.98 of its release height.
In 63 of 72 readable videos the generated ball rises above the release height, and in 37 it runs to the top of the ramp (\cref{fig:cases,fig:energy}a).
The mean height ratio per model is 1.15--2.36.
Barriers show the same result as a threshold.
Physics passes the barriers below the release height and none above it.
Six models cross every barrier they were given, and overall 43 of 49 physically impassable crossings are made (\cref{fig:teaser,fig:energy}b).
In physics the speed kept over the top falls from 0.53 of the speed at the foot to zero as the barrier rises.
In the models it stays at the same level at every height, for example 0.42--0.46 in Seedance~2.0 mini.
The ball slows on a rise, as the shape suggests, by an amount unrelated to the height gained.
The preferred-speed description of \cref{sec:response} reaches its limit here: no single rule of that kind reproduces one model's behaviour on the undulating track, at the local edit and on the ramp pair together (supplement).

\subsection{What organizes the global trajectory?}
\label{sec:goal}

The local response does not propagate as a consistent physical state.
What, then, organizes where the ball ends and when?
Two paired interventions identify strong global controls.

\begin{figure*}[t]
  \centering
  \includegraphics{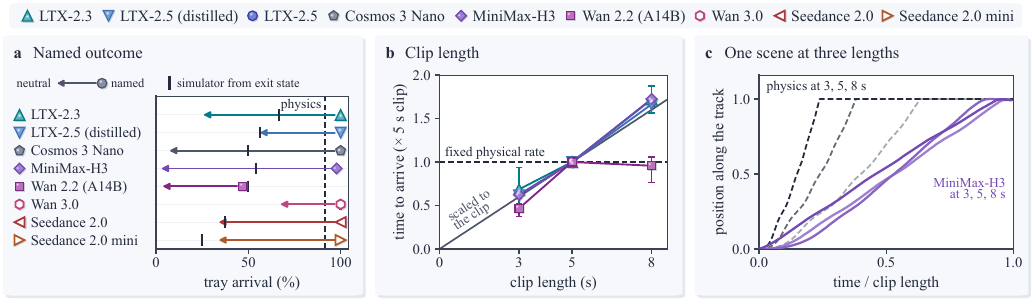}
  \caption{\textbf{Text controls the destination, clip length the timing.}
  (a) Tray arrival under the prompt that names the outcome (marker) and under the neutral prompt (arrow head), for the same first frames and seeds; ticks mark the arrival the simulator reaches from the model's own exit state, and the dashed line the simulator from the initial state.
  (b) Arrival time against clip length; the dashed line shows a fixed physical rate, the solid line scaling with the clip.
  (c) One scene and seed at 3, 5 and 8\,s: position against normalized time.}
  \label{fig:goal}
\end{figure*}

\paragraph{The named outcome controls the destination.}
The prompt of the deflector family says that the ball bounces on the deflectors until it comes to rest in the tray.
Under this prompt the ball reaches the tray in 98--100\% of the trackable videos of seven models, whether or not the contacts were right.
The generated motion itself leads there less often: continued by the simulator from the state in which the ball leaves its last correct contact, it reaches the tray in 11--84\% of the videos of six of these models, which exceed the simulator by 16--87 percentage points.
We then generate the same scenes, first frames and seeds with a prompt that describes the scene and leaves the outcome unnamed.
Tray arrival falls in all eight models, by 33 to 96 points (\cref{fig:goal}a).
The excess over the simulator is then at most 27 points in every model, and in MiniMax-H3 and the two Seedance models it is gone.
Wan~3.0 retains the most, with 67\% arrival under the neutral prompt.
The drawn tray by itself barely attracts the ball: when the tray is shifted, the landing point follows it by at most 0.42 of the shift, and with the tray removed no model brings the ball to rest where it was.

\paragraph{Clip length controls the timing.}
For four open-weight models we change only the length of the clip, from 5\,s to 3 or 8\,s, with the scene, prompt and seed fixed.
A physical event keeps its duration, so its tempo is independent of the clip length.
The generated tempo grows with the clip length with an exponent $\gamma$ of 0.85--1.04 in three models, where 1 means that the event is scaled to the clip (\cref{fig:goal}b).
MiniMax-H3 reaches the end of the track at 90--93\% of the clip at every length, LTX-2.3 at 55--59\% and the distilled LTX-2.5 at about half.
Plotted against normalized time, the three motions of one seed fall on one curve (\cref{fig:goal}c).
The response to the undulation keeps its gain and phase through this stretching, and a longer clip brings neither more arrivals at the tray nor a better order of contacts.
The time left after the event is filled by a ball at rest or by a second run of the event.
Wan~2.2 doubles its speed at 3\,s and keeps its 5\,s pace at 8\,s.
The named outcome and the clip length act together: under the neutral prompt the same scenes run 1.2--2.4 times slower in seven of eight models, and many are still under way when the clip ends.

\subsection{Photographed first frames and regeneration}
\label{sec:robust}

\begin{table}[t]
  \centering
  \caption{\textbf{Photographed first frames and regeneration.}
  (a) Change of each reading from rendered to photographed first frames of the same geometry: a check where the reading stays within the seed-to-seed spread and the change where it leaves it, bold when it is attributed to the photograph and in parentheses when it lies within the range of the weaker control; a dash where there is no reading.
  (b) Displacement $\lambda$ of the regenerated motion from the source (0) toward physics (1), by source and noise level.}
  \label{tab:robust}
  \small
  \definecolor{tbtBand}{HTML}{EEF1F6}
\definecolor{tbtRose}{HTML}{FBE3EE}
\definecolor{tbtCheck}{HTML}{8790A1}
\definecolor{tbtFaint}{HTML}{8790A1}
\definecolor{tbtNote}{HTML}{7D8597}
\definecolor{tbtSw0}{HTML}{FFFFFF}
\definecolor{tbtSw25}{HTML}{EBEDF9}
\definecolor{tbtSw50}{HTML}{D6DAF2}
\definecolor{tbtSw75}{HTML}{C2C8EC}
\definecolor{tbtSw100}{HTML}{ADB6E5}
\providecommand*{\tabglyph}[1]{\raisebox{-0.14ex}{\includegraphics[height=0.72em]{figs/glyphs/#1}}\hspace{0.34em}}
\setlength{\aboverulesep}{0pt}\setlength{\belowrulesep}{0pt}\setlength{\extrarowheight}{0.2pt}
\setlength{\tabcolsep}{2.4pt}\setlength{\fboxsep}{0pt}\setlength{\fboxrule}{0.3pt}
\noindent\makebox[\linewidth][l]{\textbf{(a)}~Photographed first frames}\par\vspace{2pt}
\noindent{\footnotesize
\begin{tabular}{>{\raggedright\arraybackslash}p{\dimexpr\linewidth-12\tabcolsep-136.8pt\relax}>{\centering\arraybackslash}p{28.6pt}>{\centering\arraybackslash}p{28.0pt}>{\centering\arraybackslash}p{27.6pt}>{\centering\arraybackslash}p{39.7pt}>{\centering\arraybackslash}p{12.9pt}}
\toprule
 & \multicolumn{3}{>{\columncolor{tbtBand}[\tabcolsep][0pt]}c}{Track} & \multicolumn{2}{>{\columncolor{tbtBand}[0pt][\tabcolsep]}c}{Deflector} \\
Model & \multicolumn{1}{>{\columncolor{tbtBand}[\tabcolsep][\tabcolsep]}c}{gain $|H|$} & \multicolumn{1}{>{\columncolor{tbtBand}[\tabcolsep][\tabcolsep]}c}{tempo $\alpha$} & \multicolumn{1}{>{\columncolor{tbtBand}[\tabcolsep][0pt]}c}{readable} & \multicolumn{1}{>{\columncolor{tbtBand}[0pt][\tabcolsep]}c}{exit residual} & \multicolumn{1}{>{\columncolor{tbtBand}[\tabcolsep][\tabcolsep]}c}{tray} \\
\midrule
\tabglyph{ltx23_full}LTX-2.3 & \textcolor{tbtCheck}{$\checkmark$} & \textcolor{tbtCheck}{$\checkmark$} & \textcolor{tbtCheck}{$\checkmark$} & -- & -- \\
\tabglyph{ltx25_distilled}LTX-2.5~(distilled) & \textcolor{tbtCheck}{$\checkmark$} & \textcolor{tbtCheck}{$\checkmark$} & \textcolor{tbtCheck}{$\checkmark$} & \textcolor{tbtCheck}{$\checkmark$} & \textcolor{tbtCheck}{$\checkmark$} \\
\tabglyph{ltx25_full}LTX-2.5 & \textcolor{tbtCheck}{$\checkmark$} & \textcolor{tbtCheck}{$\checkmark$} & \cellcolor{tbtRose}\textbf{$-$42\,\%} & -- & -- \\
\tabglyph{cosmos3_nano}Cosmos 3 Nano & \textcolor{tbtCheck}{$\checkmark$} & \textcolor{tbtCheck}{$\checkmark$} & \textcolor{tbtCheck}{$\checkmark$} & \textcolor{tbtCheck}{$\checkmark$} & \textcolor{tbtCheck}{$\checkmark$} \\
\tabglyph{minimax_h3}MiniMax-H3 & \textcolor{tbtCheck}{$\checkmark$} & \textcolor{tbtCheck}{$\checkmark$} & \textcolor{tbtCheck}{$\checkmark$} & \cellcolor{tbtRose}\textbf{$+$18$^\circ$} & \textcolor{tbtCheck}{$\checkmark$} \\
\tabglyph{wan22_i2v_a14b}Wan 2.2~(A14B) & \textcolor{tbtCheck}{$\checkmark$} & \textcolor{tbtCheck}{$\checkmark$} & \textcolor{tbtCheck}{$\checkmark$} & \textcolor{tbtCheck}{$\checkmark$} & \textcolor{tbtCheck}{$\checkmark$} \\
\arrayrulecolor{tbtFaint}\specialrule{0.35pt}{0pt}{0pt}\arrayrulecolor{black}
\tabglyph{wan30}Wan 3.0 & ($+$0.35) & ($+$0.26) & \textcolor{tbtCheck}{$\checkmark$} & \textcolor{tbtCheck}{$\checkmark$} & \textcolor{tbtCheck}{$\checkmark$} \\
\tabglyph{seedance20}Seedance 2.0 & \cellcolor{tbtRose}\textbf{$+$0.66} & \textcolor{tbtCheck}{$\checkmark$} & \textcolor{tbtCheck}{$\checkmark$} & \textcolor{tbtCheck}{$\checkmark$} & \textcolor{tbtCheck}{$\checkmark$} \\
\tabglyph{seedance20_mini}Seedance 2.0 mini & \textcolor{tbtCheck}{$\checkmark$} & \textcolor{tbtCheck}{$\checkmark$} & \textcolor{tbtCheck}{$\checkmark$} & \textcolor{tbtCheck}{$\checkmark$} & \textcolor{tbtCheck}{$\checkmark$} \\
\bottomrule
\end{tabular}}\par\vspace{6pt}
\noindent\makebox[\linewidth]{\textbf{(b)}~Regeneration\hfill{\footnotesize$\lambda$:\ 0 source\ \fcolorbox{tbtFaint}{tbtSw0}{\rule{0pt}{0.95ex}\rule{1.5ex}{0pt}}\hspace{0.1em}\fcolorbox{tbtFaint}{tbtSw25}{\rule{0pt}{0.95ex}\rule{1.5ex}{0pt}}\hspace{0.1em}\fcolorbox{tbtFaint}{tbtSw50}{\rule{0pt}{0.95ex}\rule{1.5ex}{0pt}}\hspace{0.1em}\fcolorbox{tbtFaint}{tbtSw75}{\rule{0pt}{0.95ex}\rule{1.5ex}{0pt}}\hspace{0.1em}\fcolorbox{tbtFaint}{tbtSw100}{\rule{0pt}{0.95ex}\rule{1.5ex}{0pt}}\ 1 physics}}\par\vspace{2pt}
\noindent{\footnotesize
\begin{tabular}{>{\raggedright\arraybackslash}p{\dimexpr\linewidth-16\tabcolsep-126.0pt\relax}>{\centering\arraybackslash}p{21.0pt}>{\centering\arraybackslash}p{21.0pt}>{\centering\arraybackslash}p{21.0pt}p{0pt}>{\centering\arraybackslash}p{21.0pt}>{\centering\arraybackslash}p{21.0pt}>{\centering\arraybackslash}p{21.0pt}}
\toprule
 & \multicolumn{3}{>{\columncolor{tbtBand}[\tabcolsep][\tabcolsep]}c}{\tabglyph{ltx25_distilled}LTX-2.5~(distilled)} &  & \multicolumn{3}{>{\columncolor{tbtBand}[\tabcolsep][\tabcolsep]}c}{\tabglyph{wan22_i2v_a14b}Wan 2.2~(A14B)} \\
\makebox[\linewidth]{Source\hfill{\scriptsize\color{tbtNote}noise level}} & \multicolumn{1}{>{\columncolor{tbtBand}[\tabcolsep][\tabcolsep]}c}{0.70} & \multicolumn{1}{>{\columncolor{tbtBand}[\tabcolsep][\tabcolsep]}c}{0.80} & \multicolumn{1}{>{\columncolor{tbtBand}[\tabcolsep][\tabcolsep]}c}{0.90} &  & \multicolumn{1}{>{\columncolor{tbtBand}[\tabcolsep][\tabcolsep]}c}{0.90} & \multicolumn{1}{>{\columncolor{tbtBand}[\tabcolsep][\tabcolsep]}c}{0.94} & \multicolumn{1}{>{\columncolor{tbtBand}[\tabcolsep][\tabcolsep]}c}{0.95} \\
\midrule
constant speed & \cellcolor[HTML]{FEFEFF}0.01 & \cellcolor[HTML]{FFFFFF}0.00 & \cellcolor[HTML]{FFFFFF}0.00 &  & \cellcolor[HTML]{FEFEFF}0.01 & \cellcolor[HTML]{FEFEFF}0.02 & \cellcolor[HTML]{FEFEFF}0.02 \\
phase $+90^\circ$ & \cellcolor[HTML]{FFFFFF}0.00 & \cellcolor[HTML]{FDFDFE}0.02 & \cellcolor[HTML]{F7F8FC}0.10 &  & \cellcolor[HTML]{FEFEFF}0.01 & \cellcolor[HTML]{FAFAFD}0.07 & \cellcolor[HTML]{F9F9FD}0.08 \\
slowed $2\times$ & \cellcolor[HTML]{FFFFFF}0.00 & \cellcolor[HTML]{FAFBFE}0.06 & \cellcolor[HTML]{ECEEF9}0.24 &  & \cellcolor[HTML]{FFFFFF}0.00 & \cellcolor[HTML]{FFFFFF}0.00 & \cellcolor[HTML]{FEFEFF}0.01 \\
\arrayrulecolor{tbtFaint}\specialrule{0.35pt}{0pt}{0pt}\arrayrulecolor{black}
restitution 0.3 & \cellcolor[HTML]{FEFEFF}0.01 & \cellcolor[HTML]{FDFDFE}0.03 & \cellcolor[HTML]{E5E8F7}0.32 &  & \cellcolor[HTML]{FEFEFF}0.01 & \cellcolor[HTML]{AEB6E5}0.99 & \cellcolor[HTML]{9DA7E0}1.40 \\
transparent deflector & \cellcolor[HTML]{F9FAFD}0.07 & \cellcolor[HTML]{ADB5E5}1.01 & \cellcolor[HTML]{ADB5E5}1.01 &  & \cellcolor[HTML]{E8EBF8}0.28 & \cellcolor[HTML]{A6AFE3}1.09 & \cellcolor[HTML]{A4ADE2}1.12 \\
\bottomrule
\end{tabular}}\par

\end{table}

\paragraph{The pattern holds on photographs.}
We built eight of the scenes, photographed them, measured their geometry from the photographs and ran the simulator on the measured geometry.
With the photographs as first frames, the readings of six of the nine models stay within their seed-to-seed spread of the readings on renders of the same geometry (\cref{tab:robust}a).
The three changes are specific: LTX-2.5 moves the camera on photographs, MiniMax-H3 shows larger exit residuals, and Seedance~2.0 a higher gain.
The pattern of \cref{sec:response,sec:state,sec:goal} is reproduced on photographs.
The speed follows the drawn undulation, with a phase lead in eight models.
The trend ratio stays below 0.5.
The order of three contacts appears in at most one of six videos per model.
The ball reaches the named tray in every trackable video of every model except Wan~2.2, which misses it on renders as well.

\paragraph{Regeneration returns the model's own motion.}
Finally we give two models a source video whose structure is intact and whose dynamics are wrong, add noise and let the model regenerate it.
We read $\lambda$, the displacement of the regenerated motion from the source toward physics: 0 keeps the source, 1 reaches physics.
A motion slowed twofold keeps its tempo.
A constant-speed motion keeps its timing, and the response to the undulation is added back.
A phase shifted by $90^\circ$ returns to the model's own lead.
For these sources $\lambda$ stays below 0.1, except at the highest noise level of the distilled LTX-2.5, where the output drifts toward the model's own video (\cref{tab:robust}b).
A violated contact is changed: a ball that passes through a drawn deflector is made to hit it, with $\lambda\approx1$.
What replaces the source is the model's own trajectory for that scene, with the order of contacts still wrong and a restitution of 0.16--0.45 in place of the prescribed 0.85.
For a source with a wrong restitution, the regenerated motion of Wan~2.2 moves 0.66 of the way toward the model's own image-to-video output of the same scene and seed, and 0.10 toward physics.

\section{Discussion}
\label{sec:discussion}

\paragraph{Three signals, three roles.}
The measurements assign a role to each input of an image-to-video model.
The geometry drawn in the first frame is kept, and it modulates the motion where it is drawn.
The text names an outcome, and the motion is brought to it, whereas text that names a physical parameter leaves the dynamics in place~\cite{khanbayov2026physweep}.
The clip length fixes a span, and the event is fitted into it.
A similar division appears in video-to-audio generation, where the caption decides what is generated and the video decides when~\cite{li2026flatsounds}.
A fixed law organizes motion differently: through a state that is updated where the ball is, composed across events and bounded by energy.
The generated motion reproduces the local imprint of geometry on speed and direction, and the destination and the pace come from the conditions on the whole clip.
This is consistent with generation that shapes the clip as a whole: a destination and a duration are available as global conditions, while a state has to be computed event by event~\cite{diazchao2026seriality,li2026violate}.

\paragraph{Evaluating physics in generated video.}
Two common readings are affected.
A score based on the outcome of an event can be satisfied through the prompt: with the outcome named, the ball arrives more often than its own motion allows.
A score normalized for tempo absorbs the clip length: the same event is slower in a longer clip.
The paired interventions used here address both, because the quantity read is the change of the motion under a controlled change of one input, with the law as the reference for what that change should be.
This matters when video generators are used to predict the consequence of an action: a description that names the intended outcome can bias an outcome-based evaluation even when the intervening dynamics are wrong.

\paragraph{Scope.}
The study covers one rigid ball in two planar families of geometry, nine image-to-video models and clips of 3 to 8\,s.
The reference is a simulator that is exact for the drawn geometry and the stated law, and photographed first frames test the visual domain.
The design carries over to other laws and scenes: it needs a family of geometries with matched statistics, a simulator for the reference, and pairs of generations that differ in one thing.

{
  \small
  \bibliographystyle{ieeenat_fullname}
  \bibliography{refs}
}

\clearpage
\raggedbottom
\setcounter{section}{0}
\renewcommand{\thesection}{\Alph{section}}
\renewcommand{\theHsection}{supp.\Alph{section}}
\counterwithin{figure}{section}
\counterwithin{table}{section}
\counterwithin{equation}{section}
\twocolumn[{\centering
  \pdfbookmark[0]{Supplementary Material}{supplementary}%
  \Large\textbf{\papertitle}\\
  \vspace{0.5em}Supplementary Material\\
  \vspace{1.0em}}]
\raggedbottom
\edef\suppdbltop{\the\value{dbltopnumber}}\setcounter{dbltopnumber}{1}

\begin{tcolorbox}[supp card,title={\textbf{Key}}]
\scriptsize
\renewcommand{\arraystretch}{1.25}%
\begin{tabular}{@{}l@{\hspace{1.2mm}}l@{\hspace{3mm}}l@{\hspace{1.2mm}}l@{}}
\multicolumn{2}{@{}l}{\gnote{open-weight models (filled marks)}} & \multicolumn{2}{l}{\gnote{closed models (hollow marks)}}\\
\mname{ltx23} & & \mname{wan30} & \\
\mname{ltx25d} & & \mname{sd20} & \\
\mname{ltx25} & & \mname{sd20mini} & \\
\mname{cosmos} & & \gnote{integrity only:} & \\
\mname{h3} & & \mname{tiv5b} & \\
\mname{wan22} & & & \\
\end{tabular}

\cardsep
\begin{tabular}{@{}p{24mm}@{\hspace{1.5mm}}p{50mm}@{}}
\colorbox{sOracle}{\oraclelabel} & the simulator's reading through the same measurement chain\\
1.24\,\ci{0.79}{1.46}\,\nn{24} & median, 95\% interval over scenes, denominator\\
\ivbar[mGlacierD]{0.79}{1.24}{1.46}{1}{0}{1.6} & a value and its interval on a common scale, with a tick at the physical value\\
\sharebar{104}{107}{120} & intact (dark) and readable (light) share of the videos\\
\dumbbell{100}{25}{91} & tray arrival under the named (filled) and the neutral prompt (hollow)\\
\tile[qRespL]{1.6} & a value singled out in the caption\\
\notrun & the model did not run this arm (\nd: no reading)\\
\end{tabular}

\cardsep
Header bands follow the four questions: \cardtag{hResp}{response} \cardtag{hState}{state} \cardtag{hGoal}{goal} \cardtag{hTime}{time} \cardtag{hInt}{integrity, setup}
\end{tcolorbox}

\section{Scene families}
\label{sec:supp-families}

Every scene is a world $W=(\Omega,L,x_0)$ drawn on a vertical board: a ball released from rest at $x_0$, a fixed law $L$, and white parts whose shape is the geometry $\Omega$.
The two main families vary $\Omega$ over matched levels while the prompt stays fixed within a family, so a level reaches the model through the first frame alone.
Each intervention family changes one element of a base scene and keeps the rest.
For each we state what is held fixed, what changes, what the stated law predicts and what the family isolates.

\newsavebox{\supplvlbox}
\newcommand{\lvl}[3][21.5mm]{%
  \sbox{\supplvlbox}{\includegraphics[width=#1]{supp/thumbs/#2.jpg}}%
  \raisebox{\dimexpr0.75ex-2.8pt-0.5\ht\supplvlbox\relax}{\vbox{\offinterlineskip
    \vspace{2.4pt}\hbox{\color{sInkB}\tiny #3}\vspace{1.2pt}\hbox{\usebox{\supplvlbox}}\vspace{2.8pt}}}}

\suppneed{15}
\subsection{Common visual world}
\label{sec:supp-common}

\begin{scard}[hInt]{Common visual world}{setup}
\begin{minipage}[t]{0.40\linewidth}\vspace{0pt}
\thumb{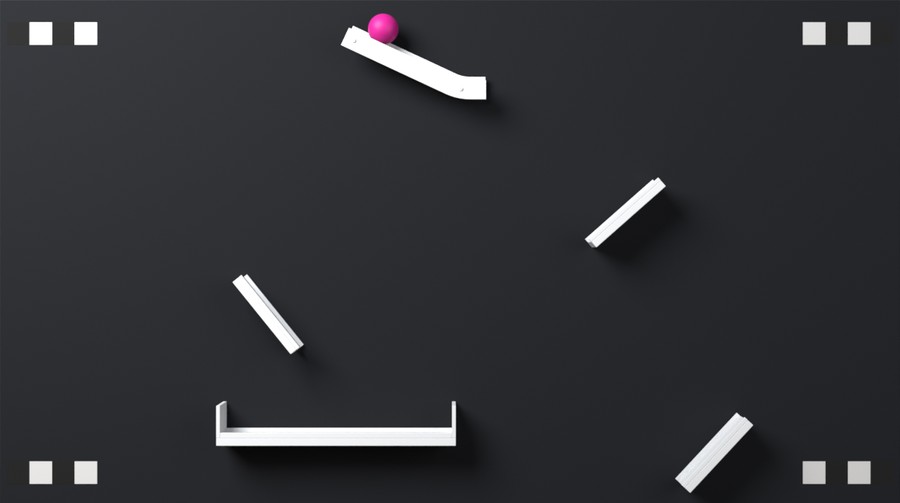}
\end{minipage}\hfill
\begin{minipage}[t]{0.57\linewidth}\vspace{0pt}\raggedright
\renewcommand{\arraystretch}{1.08}%
\begin{tabular}{@{}>{\color{sInkB}}l@{\hspace{1.2mm}}>{\raggedright\arraybackslash}p{31mm}@{}}
board & matte black, $1.20\times0.675$\,m (16:9), on a wooden table before a wall\\
markers & four 30\,mm squares per corner\\
parts & white acrylic\\
ball & 40\,mm, orange, green, blue or magenta, named in the prompt\\
camera & locked, 50\,mm lens, 1.85\,m from the board centre\\
\end{tabular}
\end{minipage}

\vspace{2pt}
A checker strip of four 30\,mm squares in each corner gives the scale and a per-frame reference for camera motion.
Frames are path-traced at $1280\times720$ and downsampled to $832\times480$.
Exact renders of the simulated motion exist at 24 and 30\,fps and, for the model that generates at 16\,fps, at exactly 16\,fps.
The first frame given to every model is the frame at $t=0$ of the exact render.
Board coordinates have their origin at the lower-left corner, $x$ to the right and $y$ up.
\end{scard}

\suppneed{8}
\subsection{Track family}
\label{sec:supp-track}

\begin{scard}[hResp]{Track family}{response}
The track is two laser-cut white acrylic plates 5\,mm thick, and the ball rides on their sharp inner top edges, 24\,mm apart, with the far plate 8\,mm higher.
Over $x\in[0,1.0]$\,m the upper edge follows
\begin{equation}
h(x)=H_0-\bar a\,x+\textstyle\sum_k A_k\sin(2\pi f_k x+\varphi_k)
\label{eq:supp-profile}
\end{equation}
and ends at a stop block (levels in Table~\ref{tab:supp-track-levels}, with random phases on the broadband level).
The track starts at a local crest of $h$.
With rolling resistance $c_{\mathrm{rr}}$ the energy balance gives
\begin{equation}
v^2(s)=v_0^2-2g_{\mathrm{eff}}\bigl[h(x(s))-h(x_0)+c_{\mathrm{rr}}\,(x(s)-x_0)\bigr]
\label{eq:supp-energy}
\end{equation}
($g_{\mathrm{eff}}$ in Sec.~\ref{sec:supp-reference}).

\vspace{2pt}
\noindent\begin{minipage}{\linewidth}
\captionof{table}{Track family levels, ten scenes each. $f$, $A$: frequency and amplitude of each component. $N_{\min}$: lowest normal force on the nominal trajectories, in units of $mg$. Last column: scenes in which all 64 ensemble members reach the stop (62--63 in the others).}
\label{tab:supp-track-levels}
\centering
\setlength{\tabcolsep}{2.0pt}
\begin{stab}{hResp}{lccccc}
\toprule
\shead level \hu{(first frame)} & $f$ \hu{(m$^{-1}$)} & $A$ \hu{(mm)} & $\bar a$ & $N_{\min}$ & 64/64 \\
\bandrule
\lvl{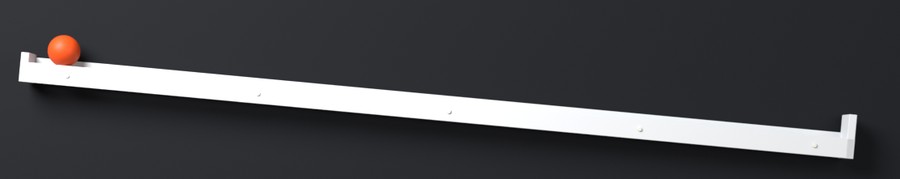}{straight} & \nd & \nd & 0.076--0.132 & 0.99 & 10 \\
\lvl{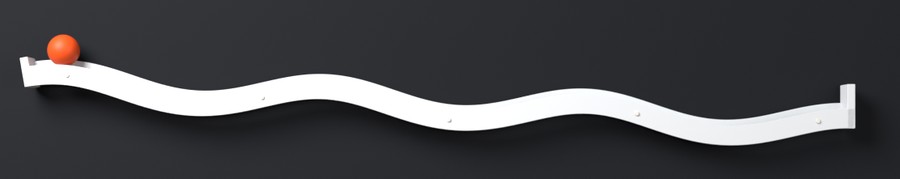}{single-frequency} & 3 & 15 & 0.033--0.115 & 0.52 & 7 \\
\lvl{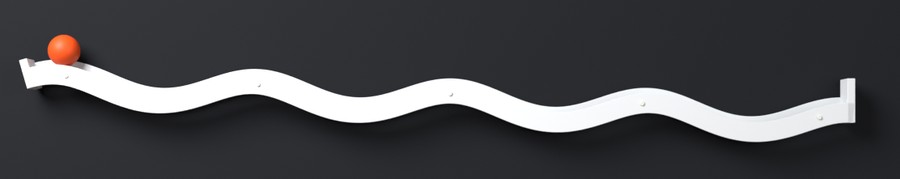}{single-frequency} & 4 & 15 & 0.031--0.085 & 0.40 & 10 \\
\lvl{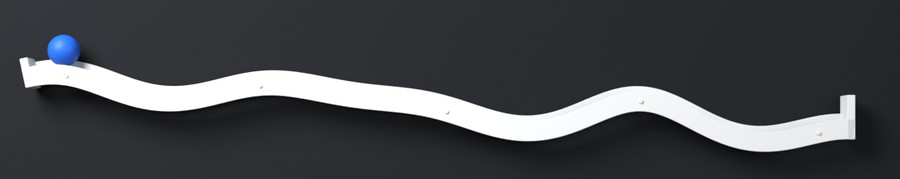}{broadband} & 2, 3, 4 & 10 each & 0.030--0.100 & 0.30 & 7 \\
\bottomrule
\end{stab}
\end{minipage}\par
\fields{The length, the release rule (the ball is released where the downhill slope first reaches 0.025, at least 10\,mm past the crest), the prompt and the physical transit time (1.62--2.15\,s). Every nominal trajectory keeps a normal force of at least $0.3\,mg$, which bounds $Af^2$ and sets the physical bandwidth of the family, and a speed of at least 0.25\,m/s beyond $x=0.16$\,m.}
{The drawn profile. The mean slope $\bar a$ is drawn per scene and enters as a covariate, and the straight level needs a steeper $\bar a$ because it lacks the first downhill of a crest start.}
{By Eq.~\eqref{eq:supp-energy} the speed follows the drawn height with unit gain and zero phase at every frequency of the family, and the mean slope accelerates the ball ($s\propto t^2$ on the straight level).}
{The transfer from drawn shape to speed, resolved by spatial frequency, at matched transit time.}
\end{scard}

\suppneed{8}
\subsection{Deflector family}
\label{sec:supp-deflector}

\begin{scard}[hResp]{Deflector family}{response, state}
A launch ramp (a 25\degree{} descent of 0.15\,m ending in a horizontal lip at $x=0.647$\,m) releases the ball into free flight toward $K$ flat deflectors ($120\times10$\,mm, with 5\,mm lips on the front and back that keep the ball in the board plane) and a white tray with 40\,mm walls and a felt floor.
Each deflector faces the launch ramp, so every contact throws the ball upward and flights between contacts last 0.2--0.4\,s (levels in Table~\ref{tab:supp-deflector-levels}).

\vspace{2pt}
\noindent\begin{minipage}{\linewidth}
\captionof{table}{Deflector family levels, eight scenes each. Order and tray: shares of the 64 ensemble members that keep the nominal sequence and that rest in the tray. Last: time of the last contact.}
\label{tab:supp-deflector-levels}
\centering
\setlength{\tabcolsep}{2.2pt}
\begin{stab}{hResp}{lccccc}
\toprule
\shead level & $K$ & angles & order & tray & last \hu{(s)} \\
\bandrule
\lvl[14mm]{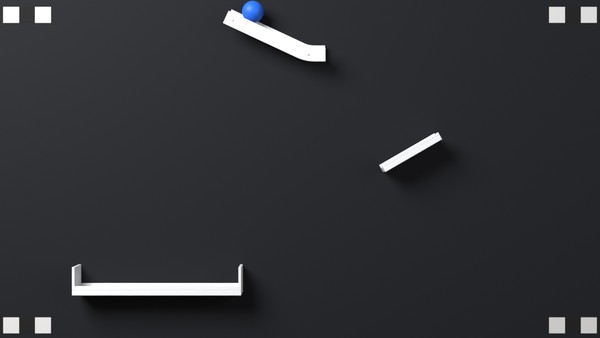}{single} & 1 & 20, 30, 45, 60\degree & 1.00 & 0.86--1.00 & 0.54--0.57 \\
\lvl[14mm]{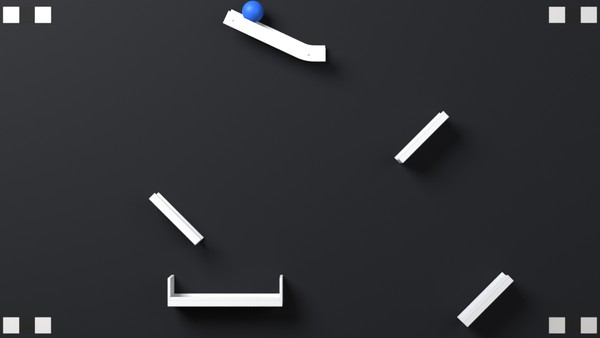}{regular} & 3 & \snum{+45}/\snum{-45}/\snum{+45}\degree & 0.94--1.00 & 0.81--0.98 & 1.12--1.16 \\
\lvl[14mm]{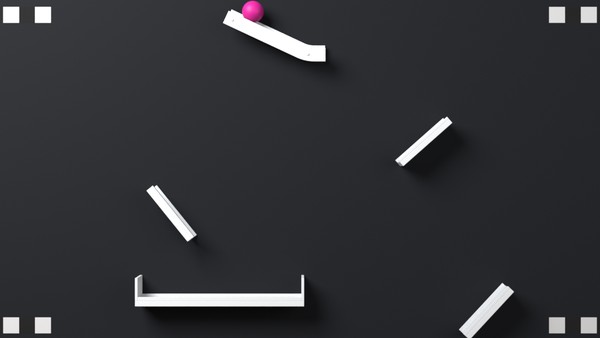}{irregular} & 3 & $\pm$(20--60)\degree & 0.91--1.00 & 0.83--0.98 & 1.05--1.15 \\
\bottomrule
\end{stab}
\end{minipage}\par
\fields{Each irregular layout shares launch ramp, tray position and total drop with the regular layout of the same index, with mean $|\theta|$ of 43.1--49.2\degree. Every nominal path hits each deflector at least 1.5 ball radii from its ends, flies at least 0.19\,s between contacts (0.22\,s between irregular deflectors, at least four frames on each side of a contact at 24\,fps), makes its last contact before 4.5\,s and rests in the tray before 5\,s. The tray widens from 200 to 300\,mm when fewer than 80\% of the ensemble rest in it.}
{The deflectors. The two single-deflector scenes of an angle slide the deflector by up to 20\,mm along its axis about a fixed contact point (190\,mm below the lip, 250\,mm at 20\degree), so they are physically identical. The regular layout is $+45/-45/+45\degree$ at equal spacing ($140$\,mm $\pm15\%$). The irregular layout alternates in sign with magnitudes in $[20\degree,60\degree]$ at random spacing.}
{The exit of each contact is set by the drawn normal and the prescribed restitution. On the single deflector the physical exit angles at 20, 30, 45 and 60\degree{} are 111.2, 130.0, 162.1 and 193.1\degree, a slope of 2.06 (a mirror gives 2). On three deflectors the ball visits $p_1\to p_2\to p_3$ and rests in the tray.}
{A discrete geometric event and the composition of three, and the paired layouts separate the physical path from a zigzag template with the same drop, start, tray and mean angle.}
\end{scard}

\suppneed{8}
\subsection{Local edits and shared-front tracks}
\label{sec:supp-local}

\begin{scard}[hState]{Local edits}{state}
\raggedright
Twelve tracks share their first 0.25\,m and release point and then blend into a straight ramp of slope $\bar a\in\{0.03,0.04,0.05,0.06\}$.
An edit $\pm\varepsilon(1-u^2)^3$ ($\varepsilon=15$\,mm, width 0.15\,m) is centred at 0.42, 0.50 or 0.58\,m.

\vspace{3pt}
{\centering
\setlength{\tabcolsep}{0.8pt}
\begin{tabular}{cccc}
\thumb[0.232\linewidth]{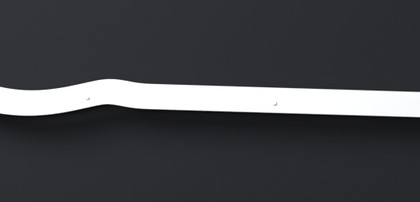} & \thumb[0.232\linewidth]{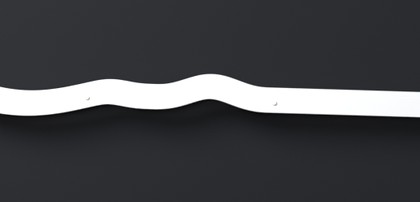} &
\thumb[0.232\linewidth]{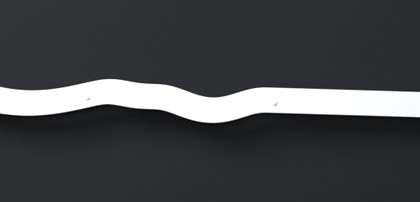} & \thumb[0.232\linewidth]{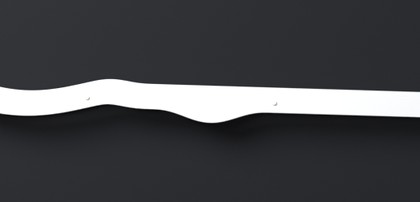}\\
\thumbcap{base} & \thumbcap{bump} & \thumbcap{dip} & \thumbcap{null edit (lower edge)}\\
\end{tabular}\par}

\vspace{2pt}
\noindent\begin{minipage}{\linewidth}
\captionof{table}{Local edits: physical arrival shift at the stop, bump / dip (ms). The ball enters the edit at 0.99, 1.11--1.12 and 1.23--1.25\,s for the three centres and reaches the base stop at 1.94, 1.89, 1.85 and 1.82\,s for the four slopes. The speed at the edit centre is 0.62--0.74 (base), 0.45--0.61 (bump) and 0.75--0.85\,m/s (dip), and 62--64 members reach the stop. $^\ast$: also run on the closed models.}
\label{tab:supp-local}
\centering
\setlength{\tabcolsep}{5pt}
\begin{stab}{hState}{cccc}
\toprule
\shead $\bar a$ & centre 0.42\,m & centre 0.50\,m & centre 0.58\,m \\
\bandrule
0.03 & \snum{+45} / \snum{-14}$^\ast$ & \snum{+41} / \snum{-13} & \snum{+37} / \snum{-11} \\
0.04 & \snum{+41} / \snum{-13} & \snum{+36} / \snum{-11} & \snum{+32} / \snum{-10}$^\ast$ \\
0.05 & \snum{+38} / \snum{-12} & \snum{+33} / \snum{-10} & \snum{+29} / \snum{-8} \\
0.06 & \snum{+35} / \snum{-11} & \snum{+30} / \snum{-9}$^\ast$ & \snum{+25} / \snum{-7} \\
\bottomrule
\end{stab}
\end{minipage}\par
\fields{The first 0.25\,m and the release point of each track (an $f=3$\,m$^{-1}$ crest start, $A=15$\,mm, $\bar a=0.06$), the seed, and one ensemble per track whose members coincide until the edit.}
{One element per arm. Each track has five arms: base, bump ($+\varepsilon$), dip ($-\varepsilon$), a null edit (the same shape added to the lower edges of the plates, area 10.3\,cm$^2$, leaving the rolling edge and the simulated motion unchanged) and a rerun of the base with the same seed.}
{Identical motion up to the edit, $\Delta v^2=\mp2g_{\mathrm{eff}}\varepsilon=\mp0.177$\,m$^2$/s$^2$ across it, and recovered speed after it with a persisting arrival shift (Table~\ref{tab:supp-local}).}
{Whether a model changes speed at the edit in the physical direction, whether an edit ahead of the ball changes the motion before arrival, and, through the null edit and the rerun, how much an unrelated change of equal size and the sampler itself move the generation.}
\end{scard}

\begin{scardp}[hState]{Shared-front tracks}{response, state}
\centering
\setlength{\tabcolsep}{0.8pt}
\begin{tabular}{cc}
\thumb[0.485\linewidth]{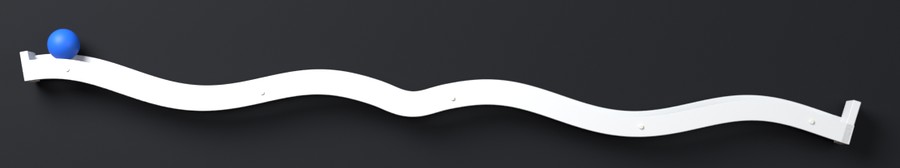} & \thumb[0.485\linewidth]{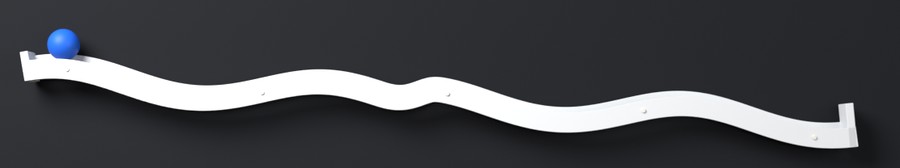}\\
\thumbcap{track 0} & \thumbcap{track 4}\\
\end{tabular}
\fields{The first 0.40\,m of twelve $f=3$\,m$^{-1}$ tracks ($A=15$\,mm, $\bar a=0.06$), all within the rules of the family.}
{The back section, blended in over 0.40--0.50\,m, with a random phase (3.61, 0.58, 0.32, 2.84, 5.29, 4.24, 5.47, 0.35, 1.39, 6.24, 3.09 and 1.51\,rad for tracks 0--11).}
{The motion is identical until the front section ends at 1.12\,s, and the back-section undulation follows the drawn phase.}
{Whether the back-section response follows the drawn phase or continues the front, and whether the motion over the shared front changes with geometry the ball has not reached.}
\end{scardp}

\subsection{Ramp pairs, barriers and chutes}
\label{sec:supp-energyspec}

These families test energy and state, and each prediction holds at any tempo (Table~\ref{tab:supp-energy-scenes}).

\begin{scardp}[hState]{Ramp pair}{state}
\begin{minipage}[t]{0.36\linewidth}\vspace{0pt}\thumb{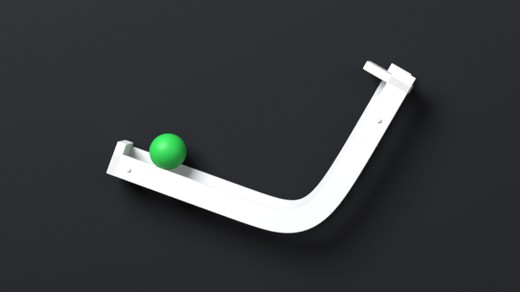}\end{minipage}\hfill
\begin{minipage}[t]{0.61\linewidth}\vspace{0pt}\raggedright
A 20\degree{} descent into a rounded valley and an ascent of 60, 30, 15, 5 or 0\degree, with release 45\,mm above the valley and a stop block on each ascent at 2.52, 1.61, 1.46 and 1.32 times the release height.
\end{minipage}
\fields{The descent, the valley and the release height.}
{The ascent angle.}
{The ball turns around just below its release height, lower on longer ascents through rolling resistance.}
{Whether the climb is bounded by the release height.}
\end{scardp}

\begin{scardp}[hState]{Barrier}{state}
\begin{minipage}[t]{0.36\linewidth}\vspace{0pt}\thumb{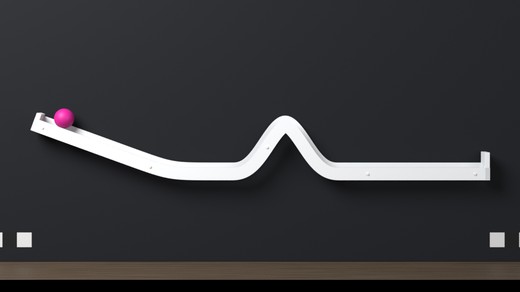}\end{minipage}\hfill
\begin{minipage}[t]{0.61\linewidth}\vspace{0pt}\raggedright
The same descent into a flat valley and a hump $(1-u^2)^3$ 220\,mm wide whose height is 0.70--1.20 of the 80\,mm release drop.
\end{minipage}
\fields{The descent, the valley, the hump width and the 80\,mm release drop.}
{The height of the hump, 0.70--1.20 of the release drop.}
{The ball passes below the rolling-resistance threshold 0.971 ($1-c_{\mathrm{rr}}L/h_{\mathrm{rel}}$ for the 0.46\,m path to the crest) with a crest speed that falls to zero at the threshold, and the ensemble splits at 0.95 and 0.971 by design.}
{Whether the outcome has a threshold at the release height.}
\end{scardp}

\begin{scardp}[hState]{Chute}{state}
\begin{minipage}[t]{0.36\linewidth}\vspace{0pt}\thumb{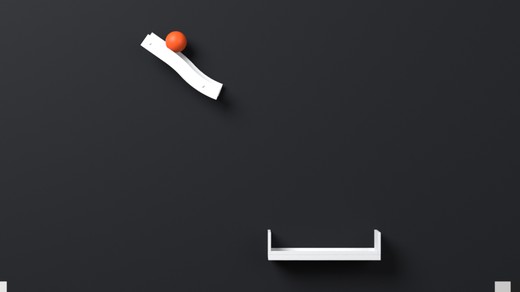}\end{minipage}\hfill
\begin{minipage}[t]{0.61\linewidth}\vspace{0pt}\raggedright
A chute set holds the exit point $(0.55,0.45)$\,m, the exit tangent ($-20\degree$) and the exit speed (0.900\,m/s) fixed and changes only the curvature $\kappa$ of the last 60\,mm.
\end{minipage}
\fields{The exit point, the exit tangent and the exit speed.}
{The curvature $\kappa$ of the last 60\,mm.}
{The four flights are one parabola, and their nominal landing points agree to $10^{-10}$\,m.}
{Whether a drawn shape that the stated law ignores moves the outcome.}
\end{scardp}

\begin{table*}[t]
\centering
\scriptsize
\caption{Ramp pairs, barriers and chutes. Times are physical, counts are ensemble members out of 64, and $^\ast$ marks scenes also run on the closed models.}
\label{tab:supp-energy-scenes}
\setlength{\tabcolsep}{3pt}
\begin{stab}{hState}{lccclcccclccc}
\toprule
\shead \multicolumn{4}{c}{ramp pairs} & \multicolumn{5}{c}{barriers} & \multicolumn{4}{c}{chutes} \\
\arrayrulecolor{sRuleB}\cmidrule(lr){1-4}\cmidrule(lr){5-9}\cmidrule(lr){10-13}\arrayrulecolor{sInk}
\shead ascent & turn \hu{(s)} & ratio & travel \hu{(mm)} & ratio & outcome & time \hu{(s)} & speed \hu{(m/s)} & pass & $\kappa$ \hu{(m$^{-1}$)} & exit, land \hu{(s)} & $N_{\min}$ & landing \hu{(m)} \\
\bandrule
60\degree$^\ast$ & 0.54 & 0.980 & 35 \gnote{(33--37)} & 0.70 & pass & 0.78 & 0.51 \gnote{(0.53)} & 64 & 0$^\ast$ & 0.31, 0.53 & 0.82 & 0.714--0.744 \\
30\degree & 0.64 & 0.975 & 92 \gnote{(84--96)} & 0.85$^\ast$ & pass & 0.83 & 0.34 \gnote{(0.35)} & 64 & \snum{+4} & 0.29, 0.51 & 0.82 & 0.714--0.742 \\
15\degree$^\ast$ & 0.87 & 0.965 & 183 \gnote{(162--193)} & 0.95 & pass & 0.90 & 0.14 \gnote{(0.15)} & 27 & \snum{-4}$^\ast$ & 0.34, 0.56 & 0.61 & 0.714--0.741 \\
5\degree & 1.74 & 0.929 & 509 \gnote{(411--572)} & 0.971 & pass & 1.03 & 0.03 \gnote{(0.03)} & 10 & \snum{+8}$^\ast$ & 0.26, 0.48 & 0.49 & 0.715--0.741 \\
0\degree$^\ast$ & stop 1.36 & \nd & \nd & 1.05$^\ast$ & turn back & 0.86 & \nd & 0 & & & & \\
 & & & & 1.20$^\ast$ & turn back & 0.83 & \nd & 0 & & & & \\
\bottomrule
\end{stab}\\[3pt]
{\scriptsize\color{sInkB} Ramps: turn-around time, height ratio and horizontal travel up the ascent (ensemble range in brackets). Barriers: time at the crest (or turn-around), crest speed with its ratio to the speed at the foot (reached at 0.61\,s). Chutes: $N_{\min}$ in units of $mg$, nominal landing 0.7377\,m for all four.}
\end{table*}

\suppneed{8}
\subsection{Goal interventions}
\label{sec:supp-trayspec}

\begin{scard}[hGoal]{Goal interventions: trays and prompt pairs}{goal}
\fields{The neutral prompt of the tray interventions (Sec.~\ref{sec:supp-prompts}), which names neither deflectors nor tray. Scenes built on a main-set scene keep its colour and ensemble members, so the two first frames differ only in the changed part. The wooden table enters the simulator because the ball can miss the tray (Table~\ref{tab:supp-tray}).}
{The tray, a second tray or the prompt. \emph{Two trays} place a second, identical tray under the launch lip. The three-deflector layouts leave no room for it, so all four scenes use single deflectors. \emph{No tray} removes the tray. \emph{Prompt pairs} regenerate the 16 regular and irregular main-set scenes with the neutral prompt, with first frame, ensemble and seed unchanged.}
{In the \emph{shifted-tray} scenes the stated law gives a capture gain of zero: the landing point does not move with the tray, and in the beyond group the ball flies over the moved tray to the table. The stated law with its nominal parameters ($\mu=0$) lands at the left edge of a one-sided ensemble spread ($+9$ to $+145$\,mm at 20\degree{} with median $+107$, $-11$ to $+128$ at 30\degree, $-23$ to $+70$ at 45\degree{} and $-42$ to $+16$ at 60\degree), so a model that lands at the ensemble median sits up to 107\,mm off the nominal point, and the primary reading compares two tray positions of the same base and seed (Sec.~\ref{sec:supp-readings}).}
{Whether a drawn tray attracts the ball, which of two trays receives it, whether a model keeps a stopping place that is no longer drawn, and whether the destination follows the text.}
\end{scard}

\suppneed{8}
\subsection{State controls}
\label{sec:supp-statecontrols}

\begin{scard}[hState]{State controls: near misses and permuted layouts}{state}
\fields{The neutral prompt (Sec.~\ref{sec:supp-prompts}), the free-flight path in the near misses, and the three deflector magnitudes in a permuted pair.}
{A deflector near the path or the order of the deflectors. \emph{Near misses} place a deflector end 10\,mm or 40\,mm from the free-flight path, with a hit control. \emph{Permuted layouts} use the same three magnitudes in the orders $(+a,-b,+c)$ and $(+a,-c,+b)$, each with its tray at its own landing point.}
{No contact at the near misses and a contact at the hit control, and on each permuted layout the order in which its deflectors are drawn (Table~\ref{tab:supp-tray}).}
{The radius within which a deflector attracts a contact (zero under the stated law), and whether the contact order follows the drawn order.}
\end{scard}

\newcommand{\traythumb}[1]{\raisebox{-0.45\height}{\vbox{\vspace{1.6pt}\hbox{\includegraphics[width=15mm]{supp/thumbs/#1.jpg}}\vspace{1.6pt}}}}
\begin{table*}[t]
\centering
\scriptsize
\caption{Tray interventions, near misses, permuted layouts and prompt pairs, all with the neutral prompt. Shifts are relative to the nominal landing point, positive toward the incoming ball. Counts are ensemble members out of 64. $^\ast$: also run on the closed models.}
\label{tab:supp-tray}
\setlength{\tabcolsep}{3pt}
\begin{stab}{hGoal}{c>{\raggedright\arraybackslash}p{3.7cm}>{\raggedright\arraybackslash}p{1.8cm}>{\raggedright\arraybackslash}p{4.0cm}>{\raggedright\arraybackslash}p{5.3cm}}
\toprule
\shead first frame & scene & base & change & physical outcome \hu{(ensemble)} \\
\bandrule
\traythumb{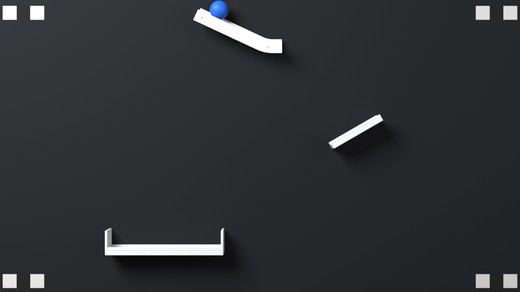} & shifted tray, in: 20, 30$^\ast$, 45, 60$^\ast$\degree & single deflector & 240\,mm tray at $+60$\,mm & lands in the tray off-centre (64 each) \\[2.6pt]
\traythumb{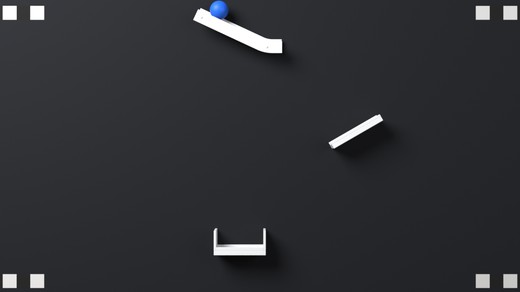} & shifted tray, beyond: 20, 30$^\ast$, 45, 60$^\ast$\degree & single deflector & 100\,mm tray at $+240$, $+220$, $+160$, $+120$\,mm & flies over it to the table (64), touches it (0), leaves the view \\[2.6pt]
\traythumb{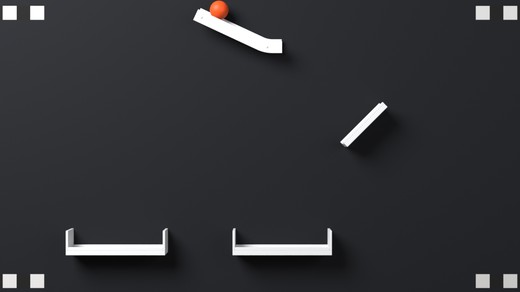} & two trays: 45\degree{} (2 scenes, 1$^\ast$) & single deflector & second 200\,mm tray under the lip & physical tray (63; 64), second tray (0) \\[2.6pt]
\traythumb{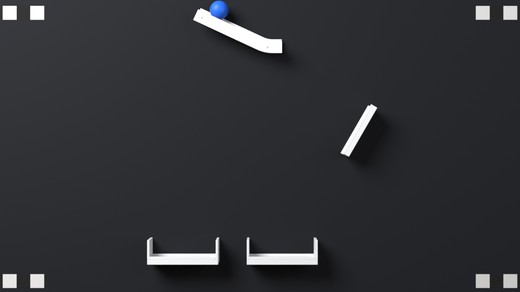} & two trays: 60\degree{} (2 scenes, 1$^\ast$) & single deflector & second tray, both 140\,mm & physical tray (64; 64), second tray (0) \\[2.6pt]
\traythumb{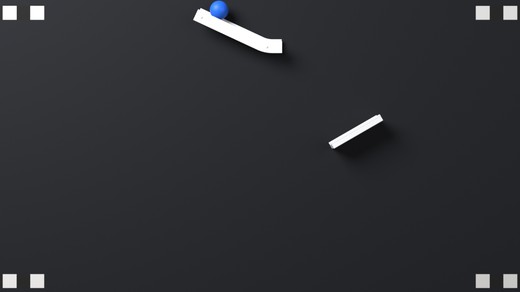} & no tray: 30\degree{} and irregular & main-set scenes & tray removed & lands on the table and leaves the view (64; 63 within 5\,s) \\[2.6pt]
\traythumb{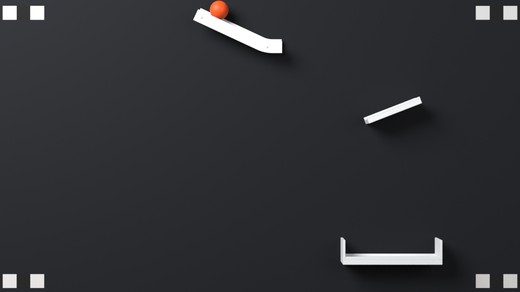} & near miss: 10\,mm above$^\ast$, 40\,mm above, 10\,mm below & free flight & deflector end at the stated clearance & no contact (0); smallest clearance 4.3, 35.7, 6.9\,mm \\[2.6pt]
\traythumb{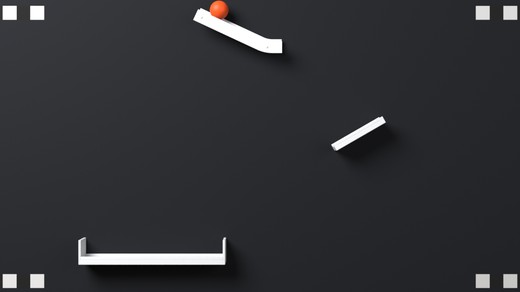} & near miss: hit control$^\ast$ & free flight & 30\degree{} deflector on the path & contact, then tray (64) \\[2.6pt]
\traythumb{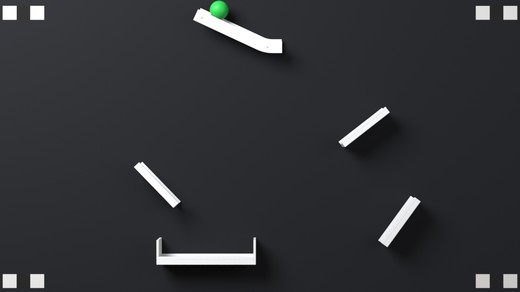} & permuted pairs (3) & new irregular & magnitudes 38/46/54, 40/47/55, 37/47/52\degree, two orders & order kept 58--63, tray 53--63 \\[2.6pt]
\traythumb{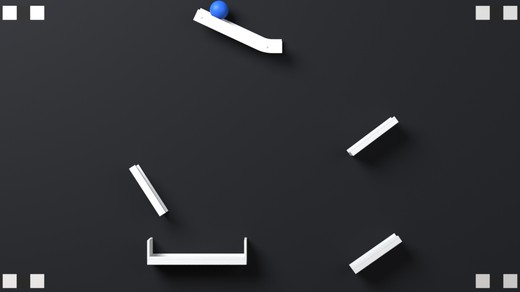} & prompt pairs (16; 4$^\ast$) & regular, irregular & neutral prompt & as in the main set \\
\bottomrule
\end{stab}
\end{table*}

\suppneed{8}
\subsection{Clip length}
\label{sec:supp-lengthspec}

\begin{scard}[hTime]{Clip length}{time}
\fields{The first frame, the prompt and the seed of eight scenes (one per track level, the 30\degree{} and 60\degree{} single deflectors, and a regular layout with its paired irregular layout). The 5\,s clips are the main-set videos.}
{The length of the clip: 3, 5 and 8\,s.}
{The stated law keeps its events at fixed times, so its tempo is independent of the clip length.}
{Whether timing is set by the motion or by the clip.}
\end{scard}

\suppneed{9}
\subsection{Photographed first frames}
\label{sec:supp-photospec}

\begin{scard}[hInt]{Photographed first frames}{integrity, visual domain}
\raggedright
The same eight scenes were built from laser-cut parts and photographed from a tripod with the camera parallel to the board within 0.7\degree.

\vspace{3pt}
{\centering
\setlength{\tabcolsep}{0.8pt}
\begin{tabular}{cc}
\thumb[0.48\linewidth]{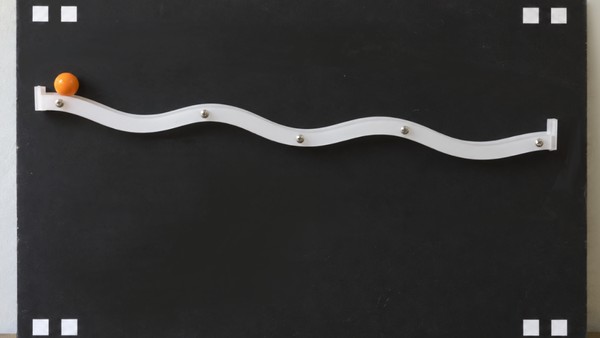} & \thumb[0.48\linewidth]{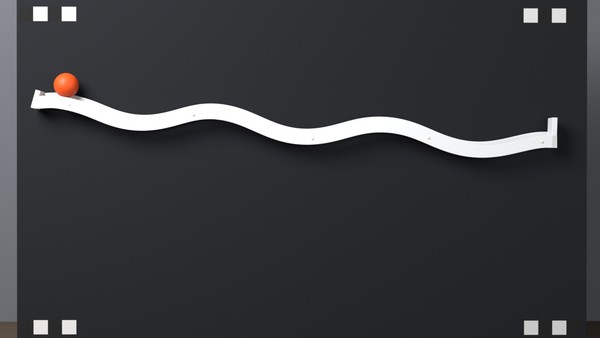}\\
\end{tabular}\par
\begin{tabular}{cc}
\thumb[0.48\linewidth]{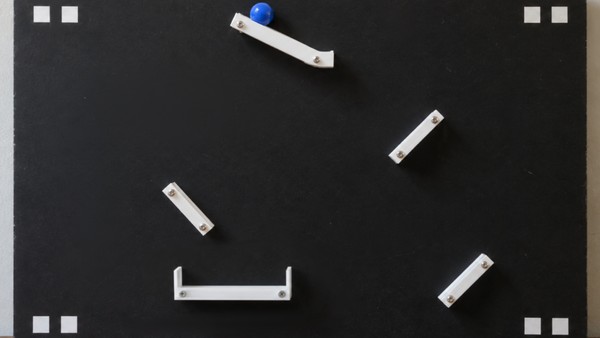} & \thumb[0.48\linewidth]{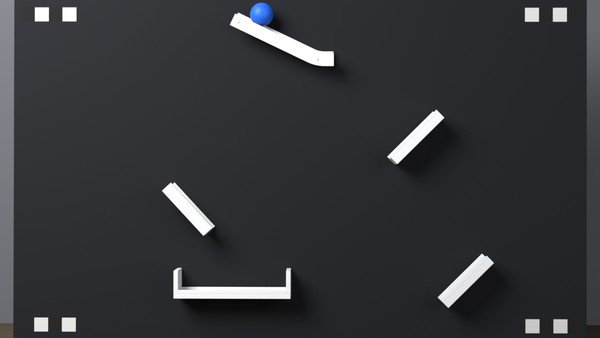}\\
\thumbcap{photograph} & \thumbcap{rendered twin}\\
\end{tabular}\par}
\fields{The geometry. It is measured from the photograph (the laser-cut end blocks set the scale, the corner strips the board homography, and the white parts are fitted in board coordinates, Table~\ref{tab:supp-asbuilt}), and the simulator runs its ensemble on it. The prompt is the main-set prompt. The control of an open-weight model is a rendered twin (measured geometry, the photograph's camera, our appearance) wherever the as-built motion leaves the nominal band, and the main-set render where it stays inside. Closed models run only the photograph rows and are compared with their own nominal-scene readings, a weaker control.}
{The visual domain of the first frame. Each photograph is cropped to 16:9 and scaled to $1280\times720$ as the first frame. The build departs from the drawing: the board is $1166\times723$\,mm, the checker squares are 27.6--29.2\,mm and the balls 46--50\,mm.}
{The as-built ensemble of Table~\ref{tab:supp-asbuilt}.}
{The visual domain of the first frame with the geometry held fixed.}
\end{scard}

\newcommand{\photothumb}[1]{\raisebox{-0.45\height}{\vbox{\vspace{1.6pt}\hbox{\includegraphics[width=16mm]{supp/thumbs/photo_#1.jpg}}\vspace{1.6pt}}}}
\begin{table*}[t]
\centering
\scriptsize
\caption{Photographed first frames: as-built geometry against the design (track offset in $x/y$ and rotation, launch exit speed with design 0.760\,m/s), the as-built ensemble (on tracks all 64 members reach the stop, on deflectors the shares keeping the design order and resting in the tray), the nominal band, and the rendered control.}
\label{tab:supp-asbuilt}
\setlength{\tabcolsep}{4pt}
\begin{stab}{hInt}{clllll}
\toprule
\shead photograph & scene & as-built against design & ensemble & nominal band & control \\
\bandrule
\photothumb{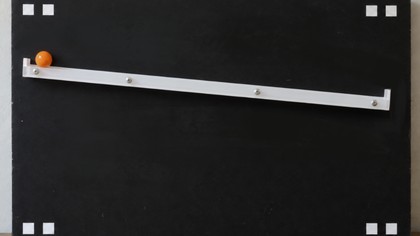} & straight & \snum{-16.3}/\snum{+52.3}\,mm, \snum{+0.07}\degree; $\bar a$ 0.093$\to$0.092 & stable & outside, 4\% faster & twin \\[2.6pt]
\photothumb{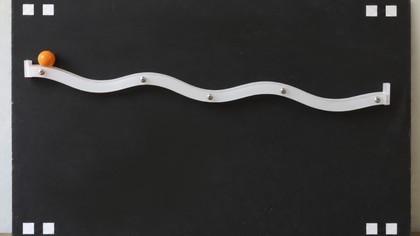} & $f=3$ & \snum{-14.9}/\snum{+42.5}\,mm, \snum{-1.36}\degree; $\bar a$ 0.055$\to$0.076; $A$ 15$\to$17.5\,mm & stable & outside, 19\% faster & twin \\[2.6pt]
\photothumb{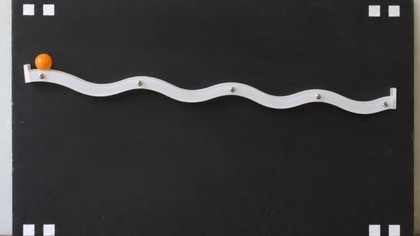} & $f=4$ & \snum{-10.8}/\snum{+28.6}\,mm, \snum{-1.03}\degree; $\bar a$ 0.048$\to$0.077; $A$ 15$\to$14.7\,mm & stable & outside 20\% of the time & twin \\[2.6pt]
\photothumb{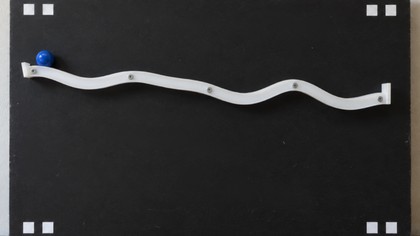} & broadband & \snum{-13.8}/\snum{+47.7}\,mm, \snum{-0.62}\degree; $\bar a$ 0.069$\to$0.078; $A$ 10.6/11.5/10.2\,mm & stable & outside, 15\% faster & twin \\[2.6pt]
\photothumb{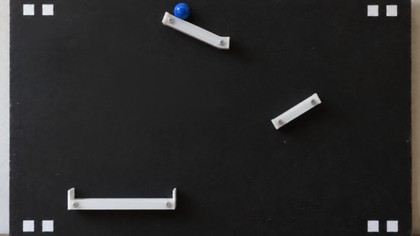} & single, 30\degree & deflector 31.7\degree, $149\times22$\,mm; exit 0.815\,m/s & 1.00 / 1.00 & inside & main-set render \\[2.6pt]
\photothumb{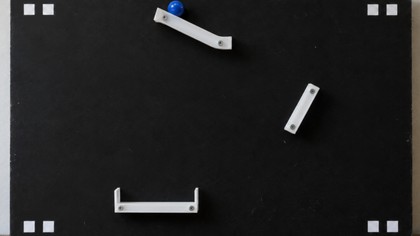} & single, 60\degree & deflector 61.1\degree, $149\times23$\,mm; exit 0.872\,m/s & 1.00 / 1.00 & contact 17\,ms late & twin \\[2.6pt]
\photothumb{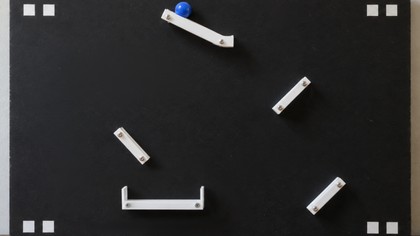} & regular & deflectors \snum{+44.8}/\snum{-47.1}/\snum{+44.8}\degree; exit 0.839\,m/s & 0.95 / 0.95 & contacts 1, 3 early & twin \\[2.6pt]
\photothumb{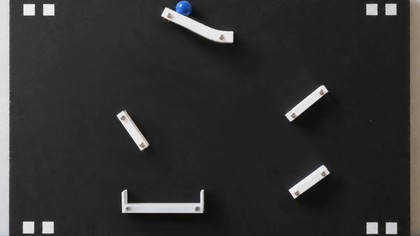} & irregular & deflectors \snum{+39.1}/\snum{-54.7}/\snum{+36.4}\degree{} (design \snum{+38.8}/\snum{-55.5}/\snum{+36.2}); exit 0.764\,m/s & 0.95 / 1.00 & contact 1 5\,ms late & twin \\
\bottomrule
\end{stab}
\end{table*}

\suppneed{9}
\subsection{Regeneration sources}
\label{sec:supp-regenspec}

\begin{scard}[hInt]{Regeneration sources}{probe}
\raggedright
Six scenes of a twelve-scene pilot set serve as sources: three tracks with $A=20$\,mm ($f=1$ and twice $f=3$\,m$^{-1}$) and three deflector layouts (one regular, two irregular), each with its exact render.

\vspace{3pt}
\noindent\begin{minipage}[t]{0.49\linewidth}\vspace{0pt}\thumb{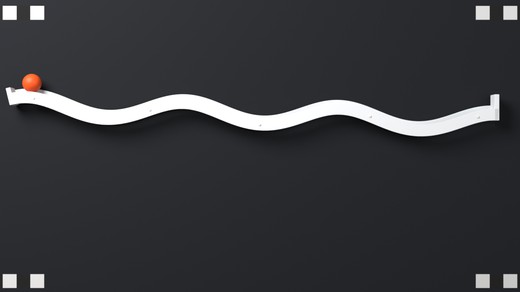}\end{minipage}\hfill
\begin{minipage}[t]{0.49\linewidth}\vspace{0pt}\thumb{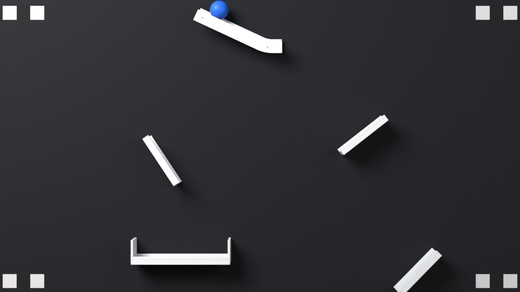}\end{minipage}

\vspace{2pt}
\noindent\begin{minipage}{\linewidth}
\captionof{table}{Regeneration sources: how each version is built and the reading that identifies it, on the source and for the physical oracle.}
\label{tab:supp-regen-sources}
\centering\scriptsize
\setlength{\tabcolsep}{2.6pt}
\begin{stab}{hInt}{l>{\raggedright\arraybackslash}p{30mm}l}
\toprule
\shead version & construction & source / oracle \\
\bandrule
constant speed & tracks: uniform speed over the physical transit time & $|H|$ 0.02 / 1.0 \\
phase $+90\degree$ & tracks: undulation of $v^2(s)$ advanced by a quarter period & phase \snum{+86} / 0\degree \\
slowed $2\times$ & both: the simulated motion at half speed & $\alpha$ 2.0 / 1.0 \\
restitution 0.3 & deflectors: every contact with $e=0.3$ & $e_n$ 0.30 / 0.85 \\
transparent & deflectors: the ball passes through deflector 2 & order 0 / 1 \\
\bottomrule
\end{stab}
\end{minipage}\par
\fields{The drawn structure. Sources are rendered at 24\,fps and at exactly 16\,fps, with a first frame equal to the exact render's within one grey level.}
{The dynamics. Each scene has three versions that keep the drawn structure and break the dynamics (Table~\ref{tab:supp-regen-sources}). A deflector ball that misses the tray lands on the table and rolls to a stop, so no version passes through a drawn surface other than the transparent deflector.}
{The physical readings of Table~\ref{tab:supp-regen-sources}.}
{Which parts of a motion a model keeps when it re-renders it and whether a replacement moves toward the stated law or toward the model's own version of the scene.}
\end{scard}

\setcounter{dbltopnumber}{\suppdbltop}

\section{The reference}
\label{sec:supp-reference}

The physical reference is a simulator that integrates the stated law on the geometry drawn in the first frame, and a 64-member ensemble that perturbs what the drawing leaves open gives each scene a tolerance band.
The reference is exact for the drawn geometry and the stated law.
All physical comparisons in the paper are defined relative to this oracle.

\paragraph{The stated law.}
On tracks, ramps and chutes the ball (radius $r=20$\,mm) rolls without slipping on the two inner plate edges, whose chord lies $r_{\mathrm{eff}}=15.5$\,mm from the ball centre.
A solid ball ($k=2/5$) then carries the inertia factor $1+\chi=1+kr^2/r_{\mathrm{eff}}^2=5/3$ and accelerates with $g_{\mathrm{eff}}=g/(1+\chi)$.
Rolling resistance is a force $c_{\mathrm{rr}}\,mg\cos\theta$ with $c_{\mathrm{rr}}=0.005$, the ball stops at the end block, and every nominal trajectory keeps a positive normal force.
Between contacts the ball flies freely, and a deflector contact reverses and scales the normal velocity by $e=0.85$ and keeps the tangential one ($\mu=0$).
The tray has a felt floor ($e=0.25$) and 40\,mm walls, and in the tray interventions the table below the board has restitution 0.60, friction 0.30 and rolling resistance 0.010.

\paragraph{The ensemble.}
Each scene draws 64 members from a key of its own (Table~\ref{tab:supp-ensemble}), perturbing what a first frame cannot show: the material of the ball, restitution and friction (with spin), rolling resistance and a few millimetres of the release point.
The arms of an intervention share their key, so the arms of a local edit coincide until the edit and a tray intervention reuses the members of its base: differences between arms come from the changed geometry alone.

\begin{table}[t]
\centering
\scriptsize
\caption{The stated law (nominal) and the ensemble (64 members). $\mathcal U$: uniform, $\mathcal N$: normal. The inertia factor is that of a solid (2/5) or a hollow (2/3) ball. The bars show each uniform range (band) with the nominal value (tick) on a scale from 0 to the range's upper end.}
\label{tab:supp-ensemble}
\setlength{\tabcolsep}{3pt}
\begin{stab}{hInt}{llllc}
\toprule
\shead family & quantity & nominal & ensemble & \\
\bandrule
rolling & inertia factor $k$ & 2/5 & $\{2/5,\ 2/3\}$ & \\
 & rolling resistance $c_{\mathrm{rr}}$ & 0.005 & $\mathcal U[0,\ 0.02]$ & \rangebar{0}{0.02}{0.005}{0}{0.02} \\
 & release point & release rule & $+\,\mathcal N(0,\ 3\,\mathrm{mm})$ & \\
\rowrule
deflectors & restitution $e$ & 0.85 & $\mathcal U[0.80,\ 0.90]$ & \rangebar{0.80}{0.90}{0.85}{0}{0.9} \\
 & friction $\mu$ & 0 & $\mathcal U[0,\ 0.3]$ & \rangebar{0}{0.3}{0}{0}{0.3} \\
 & release point & launch ramp & $+\,\mathcal N(0,\ 3\,\mathrm{mm})$ & \\
\rowrule
table & restitution & 0.60 & $\mathcal U[0.5,\ 0.7]$ & \rangebar{0.5}{0.7}{0.60}{0}{0.7} \\
 & friction & 0.30 & $\mathcal U[0.2,\ 0.4]$ & \rangebar{0.2}{0.4}{0.30}{0}{0.4} \\
 & rolling resistance & 0.010 & $\mathcal U[0.005,\ 0.02]$ & \rangebar{0.005}{0.02}{0.010}{0}{0.02} \\
\bottomrule
\end{stab}
\end{table}

\paragraph{How the band is used.}
Smooth tracks and flat contacts make each exit depend on the drawn normal and the incoming state, so the families are irregular in layout and regular in dynamics, and a scene enters the study only when its ensemble stays together: on tracks every member reaches the stop (up to two stop at the release point), on deflectors at least 90\% keep the nominal sequence and 80\% rest in the tray.
The physical value of every reading is its reading on the exact render of the nominal trajectory, measured through the same chain as the videos, and the tables give the ensemble spread where it matters (ramp heights, landing ranges, barrier passes).
The barrier heights 0.95 and 0.971 split the ensemble (27 and 10 of 64 pass), so there a model is read against a share.
The simulator also runs from states measured in a video: the destination counterfactual rolls the stated law forward from a model's own exit state, and geometric forgiveness runs the ensemble with one deflector removed.
For photographed first frames the ensemble runs on the measured geometry and decides whether the as-built scene is stable and whether it stays in the nominal band (Table~\ref{tab:supp-asbuilt}).

\section{Measurement}
\label{sec:supp-measurement}

Every video is reduced to the trajectory of the ball in board coordinates, and every reading is a function of that trajectory and the drawn geometry.
This part defines the readings, shows that the chain returns the physical values on exact renders and names each error injected into motions built to be wrong, and gives the floors against which model readings are compared.

\subsection{From pixels to the board}
\label{sec:supp-tracking}

The eight squares of the corner strips give a per-frame homography from the board to the image, which absorbs small camera motion.
Letterboxing is undone by the frame map of each model's canvas.
The ball is segmented inside the board with a hue window for its colour (orange 355--40\degree, green 105--157\degree, blue 200--244\degree, magenta 300--345\degree) and located by its chroma-weighted centroid.
White parts are bright pixels with saturation below 0.25, so the ball or its blur never counts as a part.
The ball centre lies 27.9\,mm in front of the board, and its plane is mapped by scaling the board homography about the principal point.
On a track the centre is projected onto the offset centreline to give the arc length $s(t)$, differentiated by a Savitzky--Golay filter (seven frames at 24\,fps), and on deflectors the planar path $(x,y)$ is used.
The drawn geometry is read from the white mask of every frame (the upper track edge refitted at the input frequencies, the axis and centroid of each deflector), minus the same extraction on the exact render.

\subsection{Readings}
\label{sec:supp-readings}

Table~\ref{tab:supp-readings} summarizes the readings of the main paper.

\paragraph{Track readings.}
The tempo $\alpha$ is the video time from first motion to a common arc length over the physical time.
On deflectors the launch tempo times the ball from first motion to leaving the launch ramp.
For the gain, $v^2$ is regressed jointly on $[1,\ s-\bar s,\ (s-\bar s)^2]$ and, per drawn component, on $A_k\sin\psi_k$ and $A_k\cos\psi_k$ with $\psi_k=2\pi f_k x(s)+\varphi_k$, and the coefficient pair $z_k=c_k+\mathrm{i}\,d_k$ gives
\begin{equation}
H_k=z_k/z_k^{\mathrm{ref}},
\label{eq:supp-gain}
\end{equation}
where the reference is the simulated motion rescaled to the video's tempo, aligned at first motion and read through the same filter.
$|H|$ is the gain and $\arg H$ the phase, positive when the speed undulation leads the drawn height.
The stated law and any uniformly slowed physical motion give $|H|=1$ and phase 0.
A component is read when the video traverses at least 1.5 of its cycles with 60\% of the frames in motion, and a $2f$ term enters when its frequency is not another component (intervals in Sec.~\ref{sec:supp-statistics}).
The trend ratio compares two gravities, $g_{\mathrm{trend}}=g_{\mathrm{eff}}T/T_{\mathrm{ref}}$ from the mean slope $T$ of the fitted $v^2$ trend and $g_{\mathrm{time}}=g_{\mathrm{eff}}/\alpha^2$ from the tempo.
It is 1 for a time-rescaled physical motion and near 0 for a ball held at a preferred speed.
On the straight level $s=s_0+c(t-t_0)^p$ gives the exponent $p$ (2 under the stated law, 1 at constant speed).
The relaxation rate comes from an output-error fit of $s(t)$ itself, with $s=s_0$ before $t_0$ and
\begin{equation}
\ddot s=a_c-g_w\,W(s)-b\,\dot s
\label{eq:supp-oe}
\end{equation}
afterwards ($W$ the undulating part of $\sin\theta$, fourth-order Runge--Kutta at 1/240\,s).
The stated law has $b=0$ and $g_w=g_{\mathrm{eff}}$.
The balanced accuracy is the share of moving frames on which the tangential acceleration has the sign of the local downhill direction, averaged over downhill and uphill frames ($|a|<0.06$\,m/s$^2$ counts one half).
The kinematic gain regresses the measured centre's vertical offset about its trend on the exact one, and it is 1 when the ball stays on the drawn rails.

\paragraph{Deflector readings.}
A contact is a minimum of the ball-to-deflector distance within $r+8$\,mm where a bounce model (two parabolas sharing contact point and vertical acceleration) fits at least twice as well as one parabola, with windows bounded by the neighbouring events.
The contact order merges adjacent repeats and is compared with the physical sequence.
The residual is the angle between the measured exit and the stated law's exit for the measured incoming direction, and $e_n\in[0,1.2]$ and $e_t\in[-1,1]$ are fitted along the deflector normal and tangent (0.85 and 1 under the stated law).
The calibrated law fits $(e_n,\mu)$ on a model's single-deflector contacts, freezes it and predicts its three-deflector contacts, which helps only a model with a wrong but consistent contact law.
A ball has \emph{arrived} when it is visible in half of the last 0.3\,s and its median centre lies in the tray (between the inner walls, from 10\,mm below the floor to 20\,mm above the walls).
For a wrong route, the destination counterfactual rolls the stated law forward for 5\,s from the exit state after the last correct contact (or the launch exit).
The difference between the model's arrival and that of the rollouts is the steering beyond the ball's own state, and it is tested with the state two frames later and with the velocity rescaled by the apparent gravity or by the launch tempo.

\paragraph{Intervention readings.}
\emph{Local edits}: with the anchor $t_a$ at 0.10\,m of travel, $T=t_{\mathrm{end}}-t_a$, $T_{\mathrm{pre}}=t_{\mathrm{in}}-t_a$, $v_{\mathrm{pulse}}$ the mean speed in the edit and $v_{\mathrm{pre}}$ that from 50\,mm after release to 30\,mm before the edit, each arm is paired with the base of the same track and seed,
\begin{equation}
dv=\frac{(v_{\mathrm{pulse}}/v_{\mathrm{pre}})_{\mathrm{arm}}}{(v_{\mathrm{pulse}}/v_{\mathrm{pre}})_{\mathrm{base}}}-1,\quad
dt=\frac{(T/T_{\mathrm{pre}})_{\mathrm{arm}}}{(T/T_{\mathrm{pre}})_{\mathrm{base}}}-1,
\label{eq:supp-dvdt}
\end{equation}
so that the tempo of the two videos cancels.
Locality is read from the change of $T_{\mathrm{pre}}$ and the RMS path difference before the edit.
\emph{Shared fronts}: the gain is read on the back section ($x\ge0.52$\,m) with its own phase $\varphi_b$, and the circular-linear slope of $\Phi=\delta+\varphi_b$ on $\varphi_b$ is 1 when the response follows the drawing and 0 when it continues the front.
An exploratory fit $|H|e^{\mathrm{i}\Phi}=G\,e^{\mathrm{i}\varphi_b}+C$ across tracks splits a locked part $|G|$ with lead $\arg G$ from a part $|C|$ with the same phase on every track.
\emph{Ramps}: the turn-around is the highest centre after the valley (quadratic fit within $\pm0.12$\,s), and the height ratio divides its height above the valley by that of the release.
Climbs are classed as turning back, reaching the top stop, still climbing or stopped.
\emph{Barriers}: a pass moves the centre 20\,mm beyond the crest, and the crest-speed ratio divides the crest speed by the speed at the foot.
\emph{Chutes and trays}: the landing is the $x$ where the centre first descends through 50\,mm above the tray rest height (720\,mm on chutes), and the chute exit direction comes from a parabola fitted to at most 12 flight frames.
The capture contrast pairs the in-tray and beyond scenes of one base and seed, $(x_{\mathrm{beyond}}-x_{\mathrm{in}})/(\mathrm{shift}_{\mathrm{beyond}}-\mathrm{shift}_{\mathrm{in}})$, 0 when the landing ignores the tray and 1 when it follows it, so that the model's own flight bias cancels.
\emph{Clip length}: $\gamma$ is the slope of $\log\alpha$ on $\log T_{\mathrm{clip}}$ within each scene and seed (0 for a fixed physical rate, 1 when the event is stretched to the clip).
\emph{Regeneration}: with $r$ the regenerated motion, $s$ the source and $p$ the physical motion, summed over visible frames,
\begin{equation}
\lambda=\frac{\sum\langle r-s,\,p-s\rangle}{\sum\lVert p-s\rVert^2},
\label{eq:supp-lambda}
\end{equation}
0 when the output stays on the source and 1 when it is pulled onto physics.
The copy ratio is $\mathrm{RMS}\lVert r-s\rVert/\lVert s-p\rVert$, $\kappa$ is the same projection onto the model's own image-to-video video $q$, and $r-s=\lambda_2(p-s)+\kappa_2(q-s)+\epsilon$ separates the two directions.

\begin{table*}[t]
\centering
\caption{Readings used in the main paper, as definition cards: symbol and name, definition, the physical oracle and the floor at the three native frame rates (16 / 24 / 30\,fps).}
\label{tab:supp-readings}
\begin{tcbraster}[raster columns=4,raster equal height=rows,raster column skip=1.8mm,raster row skip=1.8mm,
  raster left skip=0mm,raster right skip=0mm]
\readcard{hResp}{$|H|$}{gain}{undulation of $v^2(s)$ against the simulated motion at the same tempo}{1}{0.09 / 0.11 / 0.06}
\readcard{hResp}{$\arg H$}{phase}{lead of the speed undulation over the drawn height}{0\degree}{6.8 / 7.5 / 6.0\degree}
\readcard{hResp}{$g_{\mathrm{trend}}/g_{\mathrm{time}}$}{trend ratio}{mean-slope acceleration over that implied by the tempo}{1}{\nd}
\readcard{hResp}{$b$}{relaxation rate}{rate of return to a preferred speed, Eq.~\eqref{eq:supp-oe}}{0}{0.09 / 0.06 / 0.05\,s$^{-1}$}
\readcard{hResp}{$\alpha$}{tempo}{generated over physical time of the same motion}{1}{0.08 (clip-length pairs)}
\readcard{hResp}{$p$}{straight-ramp exponent}{$s\propto(t-t_0)^p$ on the straight level}{2}{1.96--2.02}
\readcard{hResp}{$e_n$}{restitution}{apparent normal restitution of a contact}{0.85 (band 0.80--0.90)}{0.81--0.89 / 0.68--0.91 / 0.79--0.90}
\readcard{hState}{}{contact order}{merged contact sequence equals the physical one}{100\%}{100\%}
\readcard{hResp}{}{residual}{exit against the stated law's exit for the measured incoming state}{0\degree}{p90 4.2 / 3.2 / 2.3\degree}
\readcard{hState}{$h/h_0$}{height ratio}{highest point after the valley over the release height}{0.93--0.98}{0.006, single video 0.055--0.091 (tempo 2.5)}
\readcard{hGoal}{}{tray arrival}{ball at rest inside the tray over the last 0.3\,s}{100\%}{exact on renders}
\readcard{hTime}{$\gamma$}{duration exponent}{slope of $\log\alpha$ on $\log$ clip length}{0}{from the $\alpha$ floor}
\readcard{hInt}{$\lambda$}{regeneration pull}{projection of $r-s$ on $p-s$, Eq.~\eqref{eq:supp-lambda}}{1 (0 on the source)}{distance floor 1.6--2.3\,mm}
\begin{tcolorbox}[raster multicolumn=3,enhanced,colback=qIntL!45!white,colframe=qIntL!45!white,boxrule=0pt,arc=1mm,
  left=1.6mm,right=1.6mm,top=1mm,bottom=1mm,fontupper=\scriptsize\color{sInkB}]
The floor of a reading is its spread on the simulated motion with 1\,px detection noise at $832\times480$ for the three native frame rates (95th percentile of the absolute error, the 90th percentile for the residual, or the stated range). Track floors are those of the single-frequency level with $f=3$\,m$^{-1}$ (all levels in Table~\ref{tab:supp-floors-track}). The prescribed restitution 0.85 has the ensemble band 0.80--0.90. The height-ratio floor is that of the mean over videos, followed by the single-video floor at a model tempo of 2.5.
\end{tcolorbox}
\end{tcbraster}
\end{table*}

\subsection{Validation}
\label{sec:supp-validation}

\paragraph{Exact renders.}
Exact renders go through the full chain as if they were generated videos.
On the 120 track renders the fiducials agree with the scene camera to 0.04--0.07\,px, every ball stays on the track and reaches the stop, the tempo reads 1.000--1.004 and the gain 1.02--1.10 (a pixel-level bias of at most $+9\%$).
On the 72 deflector renders every sequence is correct with residuals below 0.32\degree, and pilot renders read $e_n=0.848$--0.852 on all 42 contacts.
The drawn-geometry extraction recovers the track amplitude to $1.2\times10^{-4}$, its phase to 0.017\degree{} and deflector angles to 0.004\degree.
Table~\ref{tab:supp-intervention-validation} gives the intervention readings on exact renders.

\paragraph{Synthetic mechanisms.}
Seven candidate dynamics, simulated on tracks and read through the same estimator, leave distinct fingerprints (Table~\ref{tab:supp-mechanisms}).
Spatial smoothing lowers the gain at zero phase, temporal smoothing lowers it with a phase lead, and a response lag produces a phase lag that grows with frequency.
A constant-speed cruise flattens the trend and its $b$ is recovered exactly, a speed that follows the local slope reads $+90\degree$ at chance accuracy, and a curvature heuristic produces a large $2f$ term.
A single real gain would misread three of them (lag as bandwidth, first order as no response, cruise as an unbounded ratio), which is why gain, phase, trend ratio and $b$ are read together.

\newcommand{\gspark}[1]{\gpad{\makecell[c]{\spark{#1}{0}{2.1}{1}\\[-1.8pt]\gnote{\tiny\supplist{#1}}}}}
\newcommand{\pspark}[2]{\gpad{\makecell[c]{\spark{#1}{-80}{100}{0}\\[-1.8pt]\gnote{\tiny #2}}}}
\begin{table*}[t]
\centering
\scriptsize
\caption{Fingerprints of synthetic mechanisms on tracks with $A=20$\,mm and $\bar a=0.06$ at $f=2,3,4,5$\,m$^{-1}$ (values listed in that order and drawn as a line from $f=2$ to 5, with the physical value dashed: gain on 0 to 2.1, phase on $-80$ to $+100\degree$), 24\,fps, $1280\times720$, 1\,px noise, every motion slowed $2\times$ and started 1\,s late. Accuracy at $f=3$, medians of three seeds.}
\label{tab:supp-mechanisms}
\setlength{\tabcolsep}{4pt}
\begin{stab}{hResp}{lcclll}
\toprule
\shead mechanism & gain $|H|$ & phase \hu{(\degree)} & $2f/1f$ & trend ratio & $b$ \hu{(s$^{-1}$)} / accuracy \\
\bandrule
\oraclerow \oraclelabel & \gspark{0.98,0.99,0.98,0.95} & \pspark{-1,0,2,-1}{\snum{-1}, 0, \snum{+2}, \snum{-1}} & 0.03, 0.06, 0.11, 0.09 & 0.99, 0.99, 0.99, 0.98 & 0.0 / 0.92 \\
\oraclerule
spatial smoothing, $\sigma=40$\,mm & \gspark{0.90,0.83,0.72,0.59} & \pspark{-1,1,-2,4}{\snum{-1}, \snum{+1}, \snum{-2}, \snum{+4}} & 0.03, 0.05, 0.08, 0.10 & 1.01, 1.14, 1.23, 1.32 & 0.0 / 0.93 \\
temporal smoothing, $\sigma=0.2$\,s & \gspark{0.91,0.70,0.47,0.28} & \pspark{7,19,30,34}{\snum{+7}, \snum{+19}, \snum{+30}, \snum{+34}} & 0.18, 0.18, 0.20, 0.25 & 0.68, 0.69, 0.73, 0.74 & 0.3--0.5 / 0.84 \\
response lag 0.1\,s & \gspark{0.98,0.87,0.74,0.59} & \pspark{-26,-40,-56,-72}{\snum{-26}, \snum{-40}, \snum{-56}, \snum{-72}} & 0.09, 0.11, 0.17, 0.24 & 1.11, 1.16, 1.15, 1.08 & \snum{-0.1}--0.6 / 0.80 \\
cruise, $b=5$\,s$^{-1}$ & \gspark{0.71,0.88,0.97,0.97} & \pspark{53,39,30,26}{\snum{+53}, \snum{+39}, \snum{+30}, \snum{+26}} & 0.05, 0.06, 0.04, 0.10 & 0.10, 0.13, 0.15, 0.15 & 5.0--5.1 / 0.78 \\
first order (speed from slope) & \gspark{0.90,1.25,1.65,2.00} & \pspark{90,90,91,92}{\snum{+90}, \snum{+90}, \snum{+91}, \snum{+92}} & 0.05, 0.06, 0.10, 0.10 & 0.01, 0.01, 0.00, 0.00 & $\ge80$ / 0.52 \\
curvature slowdown, $v/(1+c|\kappa|)$ & \gspark{0.87,0.72,0.66,0.56} & \pspark{-2,-5,4,9}{\snum{-2}, \snum{-5}, \snum{+4}, \snum{+9}} & 0.31, 0.88, 1.34, 2.16 & 0.99, 1.09, 1.01, 1.20 & 0.0 / 0.66 \\
\bottomrule
\end{stab}
\end{table*}

\paragraph{Motions with a known error.}
Motions that keep the drawn structure and break one piece of dynamics show that each reading names the error it was built with (Table~\ref{tab:supp-known-track}, where the reading that names each error is tinted).
Constant speed reads as no response, a phase advanced by 90\degree{} as a phase error of $+85\degree$ at full amplitude, and a motion slowed $2\times$ keeps unit gain and zero phase with $\alpha=1.98$.
Rendered and measured, they read gains of 0.024, 1.52 and 1.03 and phases of $+4$, $+86$ and $+1\degree$.
On deflectors the calibrated law recovers a fixed restitution of 0.5 ($e_n=0.49$) and then fits the three-deflector contacts to 1--2\degree{} against 15--16\degree{} for the stated law, and a transparent second deflector breaks every sequence.
Rendered and measured, a restitution of 0.3 reads 0.300--0.304, and a motion slowed $2\times$ keeps the order with $e_n=0.848$--0.853 and a contact gravity of 2.45--2.48\,m/s$^2$, a quarter of $g$.

\newcommand{\named}[1]{\tile[qGoalL]{#1}}
\begin{table}[t]
\centering
\scriptsize
\caption{Track readings on motions with a known error ($f=3$\,m$^{-1}$ and straight tracks, 24\,fps, $832\times480$, 1\,px noise, median of three seeds). Tinted: the reading that names the error.}
\label{tab:supp-known-track}
\setlength{\tabcolsep}{3.2pt}
\begin{stab}{hResp}{lcccccc}
\toprule
\shead motion & $|H|$ & phase & $\alpha$ & acc. & $b$ & $p$ \\
\bandrule
\oraclerow \oraclelabel & 1.01 & 0\degree & 1.00 & 0.99 & 0.03 & 2.00 \\
\oraclerule
constant speed & \named{0.02} & \snum{-27}\degree & 1.08 & 0.50 & 23.2 & 1.00 \\
phase $+90\degree$ & 1.51 & \named{\snum{+85}\degree} & 1.26 & 0.54 & \snum{-0.62} & \nd \\
slowed $2\times$ & 0.96 & \snum{+1}\degree & \named{1.98} & 0.90 & 0.00 & 1.99 \\
\bottomrule
\end{stab}
\end{table}

\paragraph{Integrity rule.}
On 208 exact renders mapped onto every model's canvas and frame rate the integrity rule (Sec.~\ref{sec:supp-integrity}) raises no flag, with part displacements up to 0.15\,mm and rotations up to 0.20\degree.
Injected perturbations are read as intended (Table~\ref{tab:supp-injected}).

\begin{table}[t]
\centering
\scriptsize
\caption{Injected perturbations and how the integrity rule reads them: \okmark\ passes, \flagmark\ flagged.}
\label{tab:supp-injected}
\setlength{\tabcolsep}{3pt}
\begin{stab}{hInt}{c>{\raggedright\arraybackslash}p{30mm}c>{\raggedright\arraybackslash}p{30mm}}
\toprule
\shead & passes & & flagged \\
\bandrule
\okmark & a deflector shifted by 3\,mm & \flagmark & a deflector shifted by 8\,mm \\
\okmark & a deflector turned by 1.5\degree & \flagmark & a deflector turned by 3\degree \\
\okmark & a board drifting by 2--6\,px & \flagmark & a deflector area grown by 60\% \\
\okmark & a darkened and blurred deflector & \flagmark & a removed launch ramp \\
\okmark & a ball enlarged by 30\% in area & \flagmark & a rail plate moved by 8\,mm \\
 & & \flagmark & half an outline moved by 6\,mm \\
 & & \flagmark & a white blob of 0.6 ball areas \\
 & & \flagmark & a second ball \\
\bottomrule
\end{stab}
\end{table}

\paragraph{Photographed first frames.}
The rendered twins measured with the photograph's camera set the floor of the photograph rows: with the correction for the ball's height above the board, tracking errors are 1.3--1.4\,mm on tracks (2.4\,mm on the broadband scene, whose ball is the largest) and 2.1--2.2\,mm on deflectors, the gain is biased by $+3$ to $+5\%$ ($+10\%$ at most on the broadband scene), the phase by at most 2.5\degree, and every order is correct, as on the nominal scenes.
Pose-level floors match the nominal ones except on two track scenes at 16\,fps (gain floor up to 0.31, phase floor up to 25\degree).

\subsection{Floors}
\label{sec:supp-floors}

A floor is the spread of a reading on the simulated motion with 1\,px Gaussian detection noise (three noise seeds) at each native frame rate and output size, and a model reading is interpreted only when its interval clears the floor at the model's rate and size (Tables~\ref{tab:supp-floors-track} and~\ref{tab:supp-floors-deflector}).
For the interventions the simulated motion is first slowed $2.5\times$, a typical model tempo (Table~\ref{tab:supp-intervention-validation}).
For clip-length pairs the floor of the 3\,s minus 5\,s difference includes the truncation of the short clip (0.08 for $\alpha$, 0.15 for $|H|$, 5\degree{} for phase, 0.07\,s$^{-1}$ for $b$, 0.08 for $p$ and for accuracy), and the 8\,s minus 5\,s differences on the simulated motion are zero.

\begin{table}[t]
\centering
\scriptsize
\caption{Track floors at $832\times480$: 95th percentiles of $\bigl||H|-1\bigr|$, $|$phase$|$, $2f/1f$, $|b|$ and the kinematic-gain error, and the lowest balanced accuracy. On the straight level $p$ reads 1.96--2.02 and $|b|\le0.06$\,s$^{-1}$.}
\label{tab:supp-floors-track}
\setlength{\tabcolsep}{3.2pt}
\begin{stab}{hInt}{llcccccc}
\toprule
\shead level & fps & $|H|$ & phase & $2f/1f$ & $|b|$ & kin. & acc. \\
\bandrule
$f=3$ & 16 & 0.09 & 6.8\degree & 0.21 & 0.09 & 0.049 & 0.81 \\
 & 24 & 0.11 & 7.5\degree & 0.15 & 0.06 & 0.027 & 0.86 \\
 & 30 & 0.06 & 6.0\degree & 0.13 & 0.05 & 0.059 & 0.86 \\
\rowrule
$f=4$ & 16 & 0.08 & 4.5\degree & 0.16 & 0.09 & 0.038 & 0.93 \\
 & 24 & 0.07 & 5.6\degree & 0.15 & 0.08 & 0.026 & 0.90 \\
 & 30 & 0.04 & 4.1\degree & 0.10 & 0.05 & 0.032 & 0.94 \\
\rowrule
broadband & 16 & 0.23 & 14.1\degree & 0.23 & 0.18 & 0.028 & 0.84 \\
 & 24 & 0.17 & 8.8\degree & 0.15 & 0.13 & 0.040 & 0.88 \\
 & 30 & 0.15 & 9.8\degree & 0.19 & 0.07 & 0.036 & 0.79 \\
\bottomrule
\end{stab}
\end{table}

\begin{table}[t]
\centering
\scriptsize
\caption{Deflector floors on the 24 main-set layouts (20 noise seeds per scene): exact order, largest residual, largest per-scene 90th percentile of the residual, and the range of $e_n$ (stated law 0.85).}
\label{tab:supp-floors-deflector}
\setlength{\tabcolsep}{3.2pt}
\begin{stab}{hInt}{llcccc}
\toprule
\shead fps & size & order & residual max & p90 max & $e_n$ \\
\bandrule
16 & $1280\times720$ & 100\% & 4.8\degree & 2.9\degree & 0.83--0.88 \\
 & $832\times480$ & 100\% & 8.6\degree & 4.2\degree & 0.81--0.89 \\
\rowrule
24 & $1280\times720$ & 100\% & 2.8\degree & 2.1\degree & 0.82--0.89 \\
 & $832\times480$ & 100\% & 4.3\degree & 3.2\degree & 0.68--0.91 \\
\rowrule
30 & $1280\times720$ & 100\% & 3.0\degree & 1.5\degree & 0.81--0.88 \\
 & $832\times480$ & 100\% & 4.4\degree & 2.3\degree & 0.79--0.90 \\
\bottomrule
\end{stab}
\end{table}

\begin{table*}[t]
\centering
\scriptsize
\caption{Intervention readings on exact renders (four frame-rate and size combinations) and their floors at a model-like tempo (simulated motion slowed $2.5\times$, 1\,px noise, 16 / 24 / 30\,fps): median over 36 pairs, and the 95th percentile (p95) of the error of a single pair or video.}
\label{tab:supp-intervention-validation}
\setlength{\tabcolsep}{3pt}
\begin{stab}{hInt}{>{\raggedright\arraybackslash}p{4.3cm}>{\raggedright\arraybackslash}p{1.8cm}>{\raggedright\arraybackslash}p{4.3cm}>{\raggedright\arraybackslash}p{5.6cm}}
\toprule
\shead reading & simulator & exact renders & at tempo 2.5 \\
\bandrule
local edit, bump: $dv$ / $dt$ & \snum{-11.4} / \snum{+2.8}\% & \snum{-10.9} to \snum{-11.8} / \snum{+2.7} to \snum{+3.0}\% & \snum{-11.1} to \snum{-11.4} / \snum{+2.6} to \snum{+3.0}\%; \gnote{p95 2.4--3.9 / 1.1--1.5\%} \\
local edit, dip: $dv$ / $dt$ & \snum{+7.8} / \snum{-0.8}\% & \snum{+7.3} to \snum{+8.1} / \snum{-0.6} to \snum{-1.3}\% & \snum{+7.5} to \snum{+8.2} / \snum{-0.5} to \snum{-1.1}\%; \gnote{p95 2.8--5.6 / 1.3--1.6\%} \\
local edit: $dt$ of the bump over its simulated value & 1 & 1.07--1.10 & 0.94--1.06 \\
null edit minus base: $dv$ / $dt$ & 0 & $\le0.08$ / $\le0.03\%$ & \gnote{p95 3.0--4.6 / 1.0--1.6\%}; flagged in 0 of 120 runs \\
path RMS before the edit & 0 & \gnote{p95} 0.07--0.09\,mm & \gnote{p95} 4.2--4.8\,mm \\
\rowrule
shared front: phase slope; front dispersion & 1; 0 & 1.00; 0.03--0.05\,mm & 0.92--1.02 \gnote{(single-video phase p95 19--32\degree)}; 1.6--1.7\,mm \\
ramp pair: mean height ratio & 0.962 & 0.959--0.961 & \gnote{single-video p95} 0.055--0.091 \\
barrier: correct outcomes; 50\% crossing & 6 of 6 & 6 of 6; 1.010 & 18 of 18 \\
chute: landing offset; slope per unit $\kappa$ & 0; 0 & \snum{-0.4} to \snum{-0.5}\,mm; \snum{-0.004} to \snum{+0.008}\,mm & \gnote{landing p95} 1.9--2.9\,mm; \gnote{exit angle p95} 3.6--4.0\degree \\
regeneration: $\lambda$ of the physical motion; of the source & 1; 0 & 0.998--1.004; \snum{-0.002} to \snum{+0.004} & \gnote{distance floor} 1.6--2.3\,mm \\
\rowrule
shifted tray: capture contrast & 0 & \snum{-0.001} to \snum{+0.001} & \nd \\
two trays: physical; second tray & 100\%; 0 & 100\%; 0 & \nd \\
near miss: contacts; deflection & 0; 0\degree & 0 of 3; 0.00--0.15\degree & \nd \\
prompt pairs, permuted layouts & 0 & paired differences 0 & \nd \\
\bottomrule
\end{stab}
\end{table*}

\section{Integrity}
\label{sec:supp-integrity}

A generated video is read only while the scene stays the one that the first frame shows.
This part states the rule, reports the intact and readable shares of every model in every experiment, and describes the failure forms that recur, which are findings in their own right about how each model treats the drawn apparatus.

\suppneed{8}
\subsection{The rule}
\label{sec:supp-integrity-rule}

\begin{scard}[hInt]{Intact and readable}{integrity}
A video is \emph{intact} when five flags stay clear.
\fentry{f}{count}{a second blob of the ball's colour with a quarter of its area for more than three frames, or a resting blob of 1.6 times its area and aspect ratio 1.4 (two touching balls)}
\fentry{m}{missing}{the ball not found for more than 12 frames, a ball leaving the view being an outcome}
\fentry{m}{apparatus}{a part (each deflector, the tray, the launch ramp, each rail plate), tracked as a rigid body in board coordinates, that for six or more frames shifts by more than a quarter of the ball radius (5\,mm), turns by more than 2\degree, changes its white area outside $[0.6,1.4]$, moves its outline by more than 5\,mm or disappears}
\fentry{m}{extra}{a new white region of half the ball's area for more than six frames}
\fentry{l}{fiducial drift}{a board moving by more than 8\,px at $832\times480$, or found in fewer than 90\% of the frames}
The apparatus thresholds sit at the edge of what a reading can see.
A 2\degree{} rotation turns the exit by 4\degree, inside the 4--9\degree{} residual floor, a 5\,mm shift keeps the normal and the 8\,mm contact window, and 40\% of area is the light on lips and bevels.

\cardsep
A video is \emph{readable} when every flag but apparatus is clear and every apparatus change starts after the ball's last use of that part (the launch ramp after the ball leaves it, a struck deflector after its last contact, the tray and unstruck deflectors after the landing, the rails after the last measured point).
The intact share is the integrity statistic, and dynamics are read on readable videos, with denominators in the order all, intact, readable, trackable (ball visible in half of the frames), reached (the ball gets to where the reading is taken) and estimable.
\end{scard}

\subsection{Shares per model and experiment}
\label{sec:supp-integrity-shares}

Six models keep 96--100\% of their track videos intact (Table~\ref{tab:supp-integrity}), and the deflector scenes, whose parts are struck and whose tray receives the ball, cost every model more.
LTX-2.3 keeps the tracks and rarely the deflector scenes, and LTX-2.5 moves its camera in 64 of 72 deflector videos, which are therefore reported here only.
A tenth model, Wan~2.2 (TI2V-5B), generated on the two main families, keeps too few videos intact for any reading and appears only in this table.

\newcommand{\sh}[3]{\gpad{\makecell[c]{\sharebar{#1}{#2}{#3}\\[-1.6pt]{#1/#2\,\nn{#3}}}}}
\newcommand{\shx}[3]{\gpad{\makecell[c]{\sharebar{#1}{#2}{#3}\\[-1.6pt]{#1/\nd\,\nn{#3}}}}}
\newcommand{\shthree}[7]{\gpad{\makecell[c]{\setlength{\shw}{4.2mm}\sharebar{#1}{#2}{#7}\,\sharebar{#3}{#4}{#7}\,\sharebar{#5}{#6}{#7}\\[-1.6pt]{#1/#2, #3/#4, #5/#6\,\nn{#7}}}}}
\newcommand{\nrcell}{\notrun[11mm]}
\begin{table*}[t]
\centering
\scriptsize
\caption{Intact / readable videos per model and experiment, with the number of videos in brackets. The bars draw the intact (dark) and readable (light) shares of that number. Clip length: the new 3\,s and 8\,s clips (the 5\,s clips are main-set videos). Photographed: photograph and twin rows together. Regeneration: the three noise levels of each model. \nd: not read (LTX-2.5 on deflectors).}
\label{tab:supp-integrity}
\setlength{\tabcolsep}{2.6pt}
\begin{stab}{hInt}{lccccccccc}
\toprule
\shead & track & deflector & local edit, & ramps, barriers, & trays, & \multicolumn{2}{c}{clip length} & & \\
\shead model & main set & main set & shared front & chutes & prompt pairs & 3\,s & 8\,s & photographed & regeneration \\
\bandrule
\mname{ltx23} & \sh{104}{107}{120} & \sh{1}{14}{72} & \sh{155}{167}{216} & \sh{28}{32}{45} & \sh{20}{33}{120} & \sh{17}{19}{24} & \sh{8}{9}{24} & \sh{18}{19}{45} & \nrcell \\
\mname{ltx25d} & \sh{120}{120}{120} & \sh{54}{57}{72} & \sh{216}{216}{216} & \sh{36}{39}{45} & \sh{56}{67}{120} & \sh{22}{22}{24} & \sh{18}{20}{24} & \sh{35}{39}{45} & \shthree{63}{68}{68}{68}{58}{60}{72} \\
\mname{ltx25} & \sh{99}{100}{120} & \shx{4}{0}{72} & \sh{209}{209}{216} & \sh{20}{20}{33} & \nrcell & \nrcell & \nrcell & \sh{13}{13}{24} & \nrcell \\
\mname{cosmos} & \sh{106}{106}{120} & \sh{48}{48}{72} & \sh{174}{174}{216} & \sh{33}{33}{45} & \sh{49}{57}{120} & \nrcell & \nrcell & \sh{40}{40}{45} & \nrcell \\
\mname{h3} & \sh{120}{120}{120} & \sh{63}{68}{72} & \sh{216}{216}{216} & \sh{45}{45}{45} & \sh{119}{119}{120} & \sh{21}{21}{24} & \sh{23}{23}{24} & \sh{42}{42}{45} & \nrcell \\
\mname{wan22} & \sh{115}{115}{120} & \sh{58}{58}{72} & \sh{199}{199}{216} & \sh{43}{43}{45} & \sh{101}{105}{120} & \sh{22}{22}{24} & \sh{19}{19}{24} & \sh{41}{41}{45} & \shthree{62}{66}{60}{60}{62}{63}{72} \\
\grouprule
\mname{wan30} & \sh{120}{120}{120} & \sh{63}{63}{72} & \sh{45}{45}{45} & \sh{23}{23}{27} & \sh{35}{36}{36} & \nrcell & \nrcell & \sh{24}{24}{24} & \nrcell \\
\mname{sd20} & \sh{120}{120}{120} & \sh{52}{52}{72} & \sh{45}{45}{45} & \sh{22}{22}{27} & \sh{35}{35}{36} & \nrcell & \nrcell & \sh{24}{24}{24} & \nrcell \\
\mname{sd20mini} & \sh{120}{120}{120} & \sh{62}{62}{72} & \sh{45}{45}{45} & \sh{27}{27}{27} & \sh{31}{35}{36} & \nrcell & \nrcell & \sh{23}{23}{24} & \nrcell \\
\grouprule
\mname{tiv5b} & \sh{36}{36}{120} & \sh{1}{1}{72} & \nrcell & \nrcell & \nrcell & \nrcell & \nrcell & \nrcell & \nrcell \\
\bottomrule
\end{stab}
\end{table*}

\subsection{Recurring failure forms}
\label{sec:supp-failures}

\begin{figure*}[tp]
\centering
\includegraphics[width=\linewidth,height=7.9in,keepaspectratio]{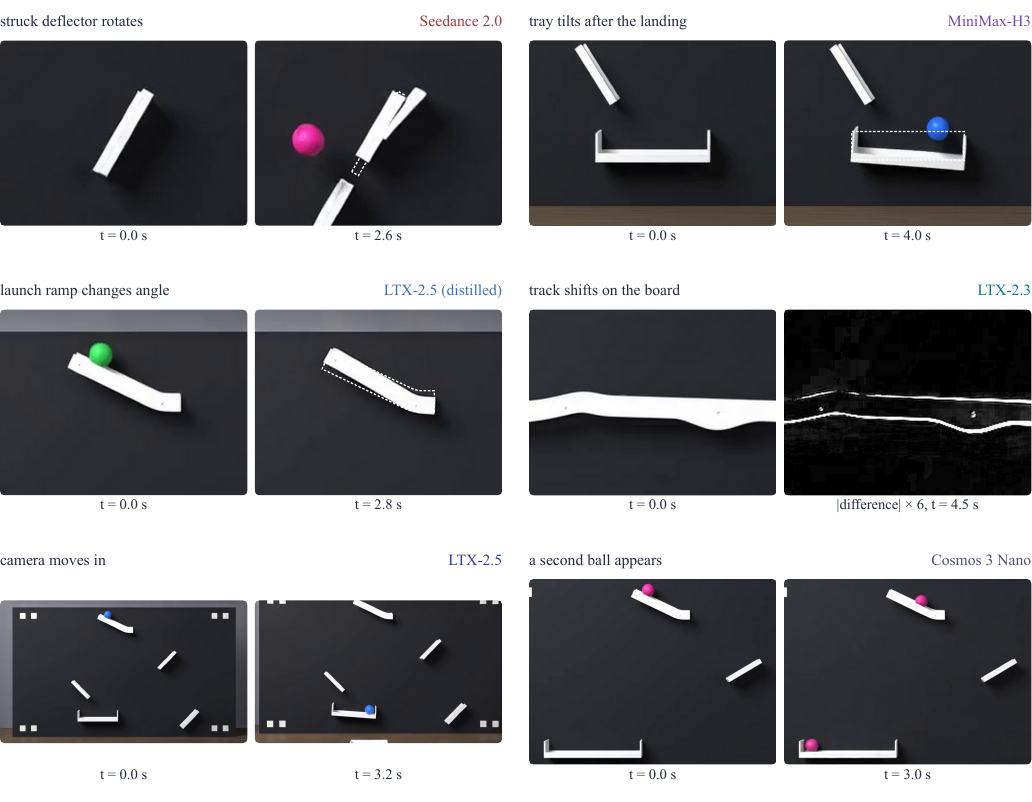}
\caption{\textbf{Recurring integrity failures.} Each pair shows the first frame and a later frame of a representative flagged video, the one nearest the median change among the flagged videos of the model shown. The later frame is the frame of largest change; for the shifted track it is shown as the amplified difference from the first frame, and for the duplicated ball it is a frame with both balls. Dashed outlines mark where the part was in the first frame. The examples show a rotating deflector, tilting tray, rotating launch ramp, shifted track, camera dolly, and ball duplication; counts and, where measured, magnitudes are reported in Table~\ref{tab:supp-failures}.}
\label{fig:supp-failures}
\end{figure*}

Table~\ref{tab:supp-failures} lists the forms that recur, and Fig.~\ref{fig:supp-failures} shows examples.
Struck parts separate the models: the Seedance models let a struck deflector turn, LTX-2.3 and MiniMax-H3 let the tray tilt after the landing, LTX-2.5 (distilled) rotates the launch ramp, mostly after the ball has left it, and LTX-2.3 treats deflectors as loose objects in every visual domain.
Other forms belong to the camera or the object count (LTX-2.5 dollies in, Cosmos 3 Nano duplicates the ball), and some grow with the clip (LTX-2.3 changes more parts in longer clips, and Wan~2.2 (A14B) lets the ball vanish after it has arrived).

\newcommand{\failthumb}[1]{\raisebox{-0.47\height}{\vbox{\vspace{1.3pt}\hbox{\includegraphics[width=15.5mm,trim=#1,clip]{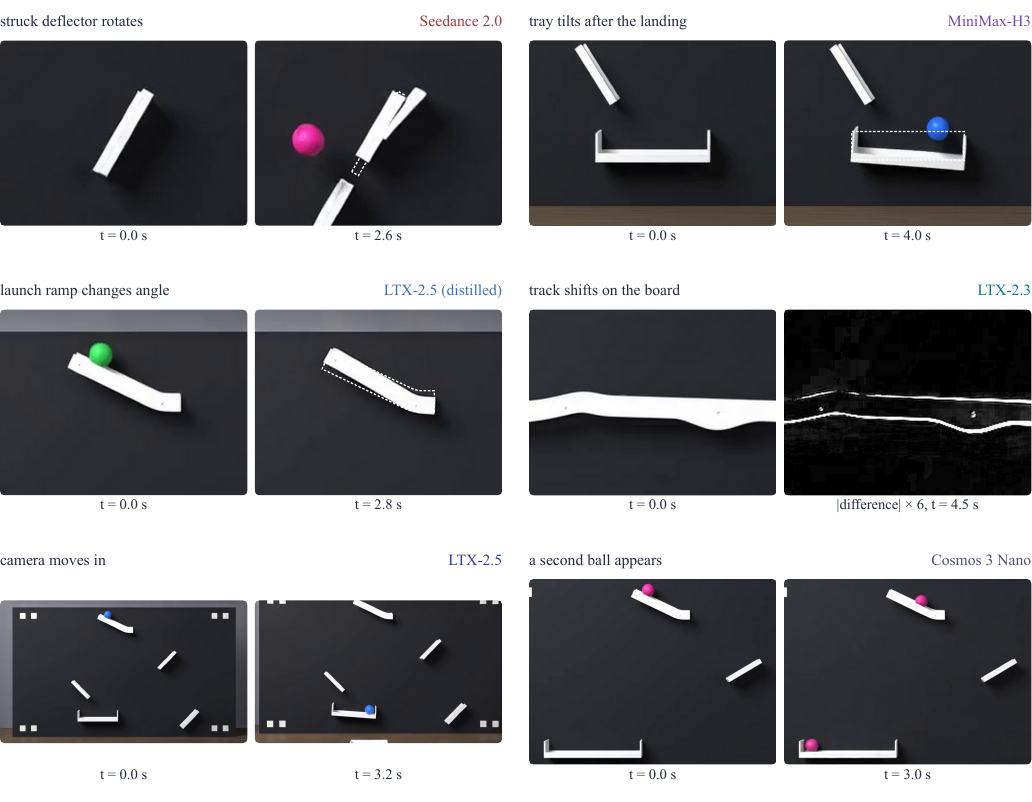}}\vspace{1.3pt}}}}
\begin{table*}[tp]
\centering
\scriptsize
\caption{Recurring failure forms, with the later frame of the example in Fig.~\ref{fig:supp-failures} where one is shown (a moved part's first-frame position dashed). Counts are flagged videos over all videos of the experiment.}
\label{tab:supp-failures}
\setlength{\tabcolsep}{3.5pt}
\begin{stab}{hInt}{c>{\raggedright\arraybackslash}p{3.1cm}>{\raggedright\arraybackslash}p{3.4cm}>{\raggedright\arraybackslash}p{8.2cm}}
\toprule
\shead example & form & model & where and how often \\
\bandrule
\failthumb{122.3 276.6 253.9 19.5} & struck deflector rotates at or after impact & \mshort{sd20}, \mshort{sd20mini} & deflector main set, apparatus 14/72 and 7/72; one deflector turned 26\degree{} and moved 17\,mm for 92 frames \\[3pt]
\failthumb{376.3 276.5 0 19.4} & tray tilts or shifts after the landing & \mshort{ltx23}, \mshort{h3} & deflector main set (LTX-2.3: 13--24\,mm and 6.6--8.5\degree); MiniMax-H3 apparatus 9/72, mostly after the landing \\[3pt]
\failthumb{122.3 147.2 253.9 148.6} & launch ramp rotates to another angle & \mshort{ltx25d} & deflector main set, 3--9\degree, 13 of its 18 flagged videos, 9 of them after the ball has left the ramp; trays and prompt pairs 62/120 \\[3pt]
 & deflectors move as loose objects & \mshort{ltx23} & apparatus on 71/72 deflector main-set and 100/120 tray and prompt videos; on photographs a deflector is knocked 125\,mm and turned 124\degree, on twins it ends in the tray; chute trays move 7--27\,mm in 12/12 \\
\failthumb{376.3 147.1 0 148.6} & rails shift against the board & \mshort{ltx23} & tracks: 16/120 in the main set, 58/180 local-edit videos (4.5--5.2\,mm, about 1\degree), in every arm including the null edit \\[3pt]
\failthumb{122.3 28.2 253.9 288.2} & camera dollies in & \mshort{ltx25} & fiducial drift on 64/72 deflector videos, 12/120 track videos, 9/12 photographed track videos against 0/12 rendered twins \\[3pt]
\failthumb{376.3 18.0 0 278.0} & second ball & \mshort{cosmos} & count flags 23/72 deflector main set, 51/120 trays and prompt pairs, 41/216 local edits and shared fronts (a second ball at the left edge of the frame), 10/45 ramps, barriers and chutes \\[3pt]
 & new objects appear & \mshort{ltx23} & photographed deflector scenes, extra 10/12 against 3/9 on twins (a large white tray enters from below) \\
 & ball vanishes after the event & \mshort{wan22} & 8\,s clips: 4/7 readable track videos lose the ball after it reaches the stop; missing 13/72 on the deflector main set \\
 & apparatus changes grow with clip length & \mshort{ltx23} & deflector clips: apparatus 6, 12 and 12 of 12 at 3, 5 and 8\,s \\
 & scene breaks down & \mshort{tiv5b} & count 33/120 and fiducial drift 48/120 on tracks; apparatus 70/72 on deflectors \\
\bottomrule
\end{stab}
\end{table*}

\section{Models and generation settings}
\label{sec:supp-models}

Nine image-to-video models generate the videos: six open-weight models run locally and three closed models are called through APIs.
This part gives what a reader needs to reproduce a generation: weights and precision, output size, frame rate and length, sampler, steps and guidance, the prompts, the seeds, and how the closed models were called.

\subsection{Models}
\label{sec:supp-models-settings}

Table~\ref{tab:supp-models} lists the settings of LTX-2.3 and LTX-2.5~\cite{hacohen2026ltx2,lightricks2026ltx23,lightricks2026ltx25}, Wan~2.2~\cite{wan2025wan,wan2025wan22}, MiniMax-H3~\cite{minimax2026h3}, Cosmos~3 Nano~\cite{nvidia2026cosmos3,nvidia2026cosmos3nano}, the Seedance models~\cite{seedance2026seedance2,volcengine2026seedancemini} and Wan~3.0~\cite{alibaba2026wan3}.
The open-weight models run through their packaged ComfyUI workflows at the 480p class of each model (Cosmos 3 Nano through its Diffusers pipeline), prompt rewriting is off for every model, and the sound tracks that some models produce are not used.
The first frame is the $1280\times720$ render, letterboxed without cropping to each open-weight model's canvas (bars of at most 18\,px) and sent unchanged to the closed models.
Models that sample with classifier-free guidance above 1 receive a structural negative prompt.
LTX-2.5 (distilled) samples without guidance and MiniMax-H3 has no negative input.

\newcommand{\yesno}[1]{\ifnum#1=1 \cbox{1}\else\cbox{0}\fi}
\begin{table*}[tp]
\centering
\scriptsize
\caption{Generation settings. Frames: frame rate $\times$ number of frames (duration). Neg.: structural negative prompt (Sec.~\ref{sec:supp-prompts}). Wan~2.2 (TI2V-5B) enters only the integrity table.}
\label{tab:supp-models}
\setlength{\tabcolsep}{3pt}
\begin{stab}{hInt}{l>{\raggedright\arraybackslash}p{3.3cm}ll>{\raggedright\arraybackslash}p{0.8cm}>{\raggedright\arraybackslash}p{2.1cm}>{\raggedright\arraybackslash}p{2.5cm}c}
\toprule
\shead model & weights, precision & output & frames & steps & guidance & sampler & neg. \\
\bandrule
\mname{ltx23} & 22B dev, fp8; distilled LoRA 384 v1.1 at 0.2 & $832\times480$ & $24\times121$ \gnote{(5.04\,s)} & 20 & video CFG 3, STG 1, rescale 0.9 & res\_2s, LTX scheduler & \yesno{1} \\
\mname{ltx25d} & 22B distilled, int8 & $832\times448$ & $24\times121$ \gnote{(5.04\,s)} & $8+3$, fixed & none (CFG 1) & Euler ancestral, distilled sigmas & \yesno{0} \\
\mname{ltx25} & 22B dev, bf16; first frame at strength 0.7 & $832\times480$ & $24\times121$ \gnote{(5.04\,s)} & 20 & video CFG 3, STG 1, rescale 0.9 & res\_2s, LTX scheduler & \yesno{1} \\
\mname{cosmos} & 16B, bf16 & $832\times480$ & $24\times121$ \gnote{(5.04\,s)} & 35 & 6.0 & UniPC, flow shift 10 & \yesno{1} \\
\mname{h3} & FL2VA, pruned int8 & $864\times480$ & $24\times124$ \gnote{(5.17\,s)} & 20 & none (guidance-distilled) & res\_multistep, shift 12 & \gnote{no input} \\
\mname{wan22} & I2V-A14B, two experts, fp8 & $832\times480$ & $16\times81$ \gnote{(5.06\,s)} & $8+8$ & CFG 3.5 & Euler, shift 5 & \yesno{1} \\
\grouprule
\mname{wan30} & API {\prompttt wan3.0-video} & $854\times480$ & $30\times150$ \gnote{(5.00\,s)} & \nd & \nd & 480P, 16:9, 5\,s, no prompt extension & \gnote{n/a} \\
\mname{sd20} & API {\prompttt doubao-seedance-\allowbreak 2-0-260128} & $864\times496$ & $24\times121$ \gnote{(5.04\,s)} & \nd & \nd & 480p, 16:9, 5\,s, no audio & \gnote{n/a} \\
\mname{sd20mini} & API {\prompttt doubao-seedance-\allowbreak 2-0-mini-260615} & $864\times496$ & $24\times121$ \gnote{(5.04\,s)} & \nd & \nd & 480p, 16:9, 5\,s, no audio & \gnote{n/a} \\
\grouprule
\mname{tiv5b} & TI2V-5B, fp16 & $832\times480$ & $24\times121$ \gnote{(5.04\,s)} & 16 & CFG 5 & UniPC, shift 8 & \yesno{1} \\
\bottomrule
\end{stab}
\end{table*}

\paragraph{Clip length.}
The 3\,s and 8\,s clips use 73 and 193 frames for LTX-2.3 and LTX-2.5 (distilled), 73 and 192 for MiniMax-H3, and 49 and 129 (16\,fps) for Wan~2.2 (A14B).
All other settings equal those of the 5\,s clips.

\paragraph{Regeneration.}
The source video is encoded by the model's autoencoder, noised to the level $\sigma$ of flow matching and denoised from there~\cite{meng2022sdedit}.
LTX-2.5 (distilled) uses the 24\,fps source on its $832\times448$ canvas with $\sigma\in\{0.7,0.8,0.9\}$, and Wan~2.2 (A14B) the exact 16\,fps source at $832\times480$ with $\sigma\in\{0.90,0.94,0.95\}$.
The levels are set per model, as the share of the model's own image-to-video deviation that each level reaches on a calibration scene: 16, 50 and 67\% for LTX-2.5 (distilled) and 14, 40 and 98\% for Wan~2.2 (A14B), whose deviation stays at 36--40\% from $\sigma=0.92$ to 0.94 and jumps to 98\% at 0.95.
The prompts are those of the pilot set, identical to the main-set prompts, and no negative prompt is used.

\paragraph{Closed models.}
The closed models were called through their providers' APIs one task at a time, with the request parameters in Table~\ref{tab:supp-models} and the watermark off.
Wan~3.0 receives the first frame inline and the Seedance models receive a link to the same file.
They run the two main sets, a subset of each intervention that needs no seed control, and the photograph rows of the photographed first frames.
The number of videos of every model in every experiment is given in Table~\ref{tab:supp-integrity}.
The subsets are three local-edit tracks (Table~\ref{tab:supp-local}) with their base, bump and dip, and shared-front tracks 0, 3, 4, 5, 7 and 11.
They include the 60, 15 and 0\degree{} ramp pairs, the 0.85, 1.05 and 1.20 barriers and the $\kappa=0,+8,-4$ chutes.
Of the tray interventions they run the 30\degree{} and 60\degree{} shifted trays of both groups, one two-tray scene per angle, and the 10\,mm-above near miss with its hit control, and of the prompt pairs four (two regular and two irregular layouts).

\subsection{Prompts}
\label{sec:supp-prompts}

Within a family the prompt is fixed, and {\prompttt\slot{colour}} stands for the ball colour of the scene.
\begin{figure*}[t]
\begin{tcbraster}[raster columns=3,raster equal height=rows,raster column skip=2.2mm,raster row skip=2.2mm,raster left skip=0mm,raster right skip=0mm]
\begin{promptcard}{Tracks}{track family}
\pfields{\cbl{1}{track main set}\cbl{1}{local edits}\cbl{1}{shared fronts}\cbl{1}{clip length}\cbl{1}{photographed tracks}\cbl{1}{regenerated tracks}}{\cbl{1}{yes, the stop at the end}\cbl{0}{no}}
\ptext{A realistic tabletop physics demonstration, filmed in real time as one continuous shot with a locked-off camera. A small \slot{colour} ball rests at the top of a narrow white track mounted on a matte black board with small black-and-white checker markers in its corners. The ball starts rolling from rest and follows the track under gravity until it reaches the stop at the end. The track, the board and the camera remain completely rigid and stationary. The scene only contains the described setup and action.}
\end{promptcard}%
\begin{promptcard}{Deflectors, named outcome}{deflector family}
\pfields{\cbl{1}{deflector main set}\cbl{1}{clip length}\cbl{1}{photographed deflectors}\cbl{1}{regenerated deflectors}}{\cbl{1}{yes, at rest in the tray}\cbl{0}{no}}
\ptext{A realistic tabletop physics demonstration, filmed in real time as one continuous shot with a locked-off camera. A small \slot{colour} ball rests at the top of a short white launch track mounted on a matte black board with small black-and-white checker markers in its corners; white flat paddles are fixed to the board below it and a white tray sits at the bottom. The ball starts rolling from rest, leaves the track, falls under gravity and bounces off the paddles until it comes to rest in the tray. The paddles, the track, the tray, the board and the camera remain completely rigid and stationary. The scene only contains the described setup and action.}
\end{promptcard}%
\begin{promptcard}{Deflectors, neutral}{interventions}
\pfields{\cbl{1}{prompt pairs}\cbl{1}{shifted trays}\cbl{1}{two trays}\cbl{1}{no tray}\cbl{1}{near misses}\cbl{1}{permuted layouts}}{\cbl{0}{yes}\cbl{1}{no}}
\ptext{A realistic tabletop physics demonstration, filmed in real time as one continuous shot with a locked-off camera. A small \slot{colour} ball rests at the top of a short white launch track mounted on a matte black board with small black-and-white checker markers in its corners; white parts are fixed to the board below it. The ball starts rolling from rest, leaves the track and moves under gravity. The white parts, the track, the board and the camera remain completely rigid and stationary. The scene only contains the described setup and action.}
\end{promptcard}%
\begin{promptcard}{Ramp pairs and barriers}{one phrase changed}
\pfields{\cbl{1}{ramp pairs}\cbl{1}{barriers}}{\cbl{0}{yes}\cbl{1}{no}}
\ptext{{\fontfamily{\familydefault}\selectfont\color{sInkC}in the track prompt:}\\[1pt]\textcolor{sInkC}{\sout{follows the track under gravity until it reaches the stop at the end}}\\[1pt]
$\rightarrow$ {\setlength{\fboxsep}{0.5pt}\colorbox{qRespL}{\strut moves along the track under gravity}}}
\end{promptcard}%
\begin{promptcard}{Chutes}{chutes}
\pfields{\cbl{1}{chutes}}{\cbl{0}{yes}\cbl{1}{no}}
\ptext{A realistic tabletop physics demonstration, filmed in real time as one continuous shot with a locked-off camera. A small \slot{colour} ball rests at the top of a short white track mounted on a matte black board with small black-and-white checker markers in its corners; a white tray is fixed to the board below the end of the track. The ball starts rolling from rest, leaves the end of the track and falls under gravity. The track, the tray, the board and the camera remain completely rigid and stationary. The scene only contains the described setup and action.}
\end{promptcard}%
\begin{promptcard}{Structural negative prompt}{guidance above 1}
\ptext{text, watermark, subtitles, people, hands, extra objects, camera motion, zoom, cuts, cartoon, painting, blurry}
\end{promptcard}
\end{tcbraster}
\caption{\textbf{Prompts.} One card per prompt: where it is used, whether it names the outcome, and its text, with the colour slot highlighted.}
\label{fig:supp-prompts}
\end{figure*}
The prompts call the deflectors paddles and the launch ramp a launch track (Fig.~\ref{fig:supp-prompts}).

\subsection{Seeds and repeatability}
\label{sec:supp-seeds}

Every open-weight model generates each scene with seeds 0, 1 and 2, and the seed drives the first noise source of the workflow.
Secondary noise sources keep the fixed values of the packaged workflows.
A rerun with the same seed reproduces the decoded frames of LTX-2.5 (distilled), LTX-2.3, LTX-2.5 and MiniMax-H3 exactly, and 33 of 36 Cosmos 3 Nano reruns are identical files.
Wan~2.2 (A14B) reruns differ in 0.2\% of the pixels by at most 7 grey levels, which leaves the measured trajectory unchanged, so the sampler floor of the local edits is zero for every open-weight model except LTX-2.3 (0.15\% in $dv$ and 0.03\% in $dt$).
The closed models either ignore the seed (Seedance) or do not reproduce it (Wan~3.0), so their three generations per scene are replicates, and paired designs compare arm means across independent samples.
A twelve-scene pilot set (seeds 0 and 1) supplies the image-to-video versions that the regeneration probe compares with.

\section{Results by question}
\label{sec:supp-results}

This part gives the per-model results behind each statement of the main paper, in the order of its questions.
How the drawn geometry modulates motion (main paper, Sec.~\ref{sec:response}) is covered by Sec.~\ref{sec:supp-res-response} and the contacts of Sec.~\ref{sec:supp-res-composition}.
Whether the modulation forms a local dynamics (Sec.~\ref{sec:state}) is covered by Secs.~\ref{sec:supp-res-composition} to~\ref{sec:supp-res-energy}, what organizes the global trajectory (Sec.~\ref{sec:goal}) by Secs.~\ref{sec:supp-res-outcome} and~\ref{sec:supp-res-timing}, and photographed first frames and regeneration (Sec.~\ref{sec:robust}) by Secs.~\ref{sec:supp-res-photo} and~\ref{sec:supp-res-regeneration}.
Table~\ref{tab:models} of the main paper holds one number per reading, and the tables below give each reading with its interval, floor and denominator.
Unless stated otherwise, values are medians over readable, estimable videos with 95\% intervals from a bootstrap over scenes (Sec.~\ref{sec:supp-statistics}), and $n$ is the number of videos.

\subsection{Response on the track family}
\label{sec:supp-res-response}

Every model keeps the drawn track.
The ball's speed rises and falls with the drawn undulation, at 0.54 to 1.24 times the physical gain at 3 cycles per metre and ahead of the physical phase in eight models, while the mean slope adds little acceleration.

\paragraph{Geometry.}
On the readable videos the drawn undulation keeps its amplitude (per-model median ratio 0.998--1.004, per model and level 0.993--1.006) and the ball stays on the track (median on-track fraction 0.98--1.00, Table~\ref{tab:supp-geometry}).
In the deflector family the deflectors keep their drawn angle (median change 0.04--0.16\degree, 95th percentile at most 0.37\degree) and position (median shift 0.15--0.29\,mm).
The kinematic gain of 0.99--1.12 shows the ball riding the drawn rails, so what follows concerns how the speed depends on the geometry.

\begin{table}[t]
\centering
\scriptsize
\caption{Drawn geometry on readable videos. Track: intact videos, median amplitude ratio of the drawn undulation, median on-track fraction and kinematic gain. Deflectors: intact videos, median change of the deflector angle with its 95th percentile in brackets, and median shift.}
\label{tab:supp-geometry}
\setlength{\tabcolsep}{1.9pt}
\begin{stab}{hResp}{lccccccc}
\toprule
\shead & \multicolumn{4}{c}{track} & \multicolumn{3}{c}{deflectors} \\
\shead model & intact & ampl. & on track & kin. & intact & angle \hu{(\degree)} & shift \hu{(mm)} \\
\bandrule
\oraclerow \oraclelabel & & 1 & 1 & 1 & & 0 & 0 \\
\oraclerule
\mshort{ltx23} & 104/120 & 0.998 & 0.975 & 1.08 & 1/72 & 0.11 \gnote{(0.28)} & 0.29 \\
\mshort{ltx25d} & 120/120 & 1.004 & 1.000 & 1.12 & 54/72 & 0.07 \gnote{(0.19)} & 0.24 \\
\mshort{ltx25} & 99/120 & 0.999 & 1.000 & 1.04 & 4/72 & \nd & \nd \\
\mshort{cosmos} & 106/120 & 0.999 & 1.000 & 1.00 & 48/72 & 0.04 \gnote{(0.16)} & 0.15 \\
\mshort{h3} & 120/120 & 1.001 & 1.000 & 1.02 & 63/72 & 0.13 \gnote{(0.28)} & 0.25 \\
\mshort{wan22} & 115/120 & 0.999 & 1.000 & 1.00 & 58/72 & 0.05 \gnote{(0.17)} & 0.20 \\
\grouprule
\mshort{wan30} & 120/120 & 1.001 & 1.000 & 1.03 & 63/72 & 0.06 \gnote{(0.16)} & 0.18 \\
\mshort{sd20} & 120/120 & 1.001 & 1.000 & 0.99 & 52/72 & 0.07 \gnote{(0.22)} & 0.27 \\
\mshort{sd20mini} & 120/120 & 1.002 & 1.000 & 1.05 & 62/72 & 0.16 \gnote{(0.37)} & 0.27 \\
\bottomrule
\end{stab}
\end{table}

\paragraph{Gain and phase.}
At $f=3$\,m$^{-1}$ the gain of LTX-2.3, both LTX-2.5 models and Wan~3.0 includes the physical value and that of the other five models lies below it (0.54--0.73).
At $f=4$ every gain lies below 1 (0.49--0.88, Table~\ref{tab:supp-gain}), and over all levels and broadband components the gain is 0.44--1.24.
At $f=3$ eight models lead the physical phase by 15--41\degree, at least twice the per-video floor of 6.0--7.5\degree{} (Wan~2.2 (A14B), with the smallest lead of $+15\degree$, sits at 2.1 times its floor), and MiniMax-H3 responds in phase ($-3\degree$, $[-10,+6]$).
At $f=4$ the lead is $+2$ to $+45\degree$, with Wan~2.2 (A14B) the in-phase model there.
The broadband components give the same picture with wider intervals.
The response repeats across seeds (phase coherence across the three seeds of a scene 0.67--0.96).
A second harmonic accompanies it: $2f/1f$ is 0.18--0.71 at $f=3$ and $f=4$, against 0.02--0.05 for the physical oracle.

\begin{table*}[t]
\centering
\perarmhead\\[-0.5pt]
\perarm{31}{387.3}
\vspace{3pt}
\caption{Response on the track family. Top: the first three rows of Fig.~\ref{fig:supp-perarm}, gain and phase by spatial frequency (large marks single-frequency tracks, small marks broadband components, physical value dashed, floor grey) and the speed along the straight ramp. Bottom: the complex gain, gain $|H|$ and phase (degrees, positive = speed leads the drawn height), for the single-frequency levels and the three broadband components, with 95\% intervals below each median. $n$: estimable videos at $f=3$, $f=4$ and broadband.}
\label{tab:supp-gain}
\scriptsize
\setlength{\tabcolsep}{2.9pt}
\begin{stab}{hResp}{llcccccccccc}
\toprule
\shead & & \multicolumn{5}{c}{gain $|H|$} & \multicolumn{5}{c}{phase \hu{(\degree)}} \\
\shead model & $n$ & $f=3$ & $f=4$ & bb 2 & bb 3 & bb 4 & $f=3$ & $f=4$ & bb 2 & bb 3 & bb 4 \\
\bandrule
\oraclerow \oraclelabel & & 1 & 1 & 1 & 1 & 1 & 0 & 0 & 0 & 0 & 0 \\
\oraclerule
\mname{ltx23} & \gnote{24, 20, 24} & \vci{1.24}{0.79}{1.46} & \vci{0.49}{0.34}{0.90} & \vci{1.10}{0.58}{1.33} & \vci{1.02}{0.84}{1.58} & \vci{0.78}{0.47}{1.12} & \vci{+38}{+27}{+49} & \vci{+44}{+31}{+57} & \vci{+32}{+6}{+67} & \vci{+41}{+26}{+56} & \vci{+54}{+28}{+79} \\
\mname{ltx25d} & \gnote{25, 23, 27} & \vci{0.97}{0.76}{1.24} & \vci{0.71}{0.63}{0.85} & \vci{0.64}{0.47}{0.89} & \vci{0.80}{0.55}{0.98} & \vci{0.95}{0.83}{1.09} & \vci{+35}{+30}{+41} & \vci{+37}{+32}{+42} & \vci{+27}{+16}{+38} & \vci{+34}{+26}{+47} & \vci{+31}{+22}{+39} \\
\mname{ltx25} & \gnote{24, 12, 15} & \vci{0.98}{0.79}{1.09} & \vci{0.68}{0.56}{0.86} & \vci{0.72}{0.59}{1.20} & \vci{0.92}{0.84}{1.31} & \vci{0.62}{0.46}{0.95} & \vci{+36}{+27}{+45} & \vci{+35}{+23}{+45} & \vci{+42}{+10}{+67} & \vci{+36}{+19}{+47} & \vci{+57}{+41}{+76} \\
\mname{cosmos} & \gnote{21, 19, 16} & \vci{0.57}{0.44}{0.88} & \vci{0.56}{0.42}{0.98} & \vci{0.82}{0.58}{0.97} & \vci{0.88}{0.36}{1.19} & \vci{0.91}{0.59}{1.09} & \vci{+41}{+13}{+66} & \vci{+26}{+9}{+41} & \vci{+11}{-29}{+52} & \vci{+12}{-7}{+28} & \vci{+34}{+18}{+50} \\
\mname{h3} & \gnote{30, 30, 30} & \vci{0.73}{0.66}{0.84} & \vci{0.54}{0.47}{0.63} & \vci{0.44}{0.33}{0.77} & \vci{0.68}{0.53}{0.85} & \vci{0.72}{0.58}{0.81} & \vci{-3}{-10}{+6} & \vci{+13}{+7}{+19} & \vci{+28}{-11}{+67} & \vci{+7}{-15}{+30} & \vci{+30}{+14}{+46} \\
\mname{wan22} & \gnote{28, 30, 26} & \vci{0.54}{0.46}{0.71} & \vci{0.67}{0.56}{0.75} & \vci{0.69}{0.28}{0.95} & \vci{0.78}{0.53}{0.91} & \vci{0.90}{0.61}{1.07} & \vci{+15}{+7}{+21} & \vci{+2}{-10}{+15} & \vci{+28}{-1}{+68} & \vci{+23}{+8}{+50} & \vci{+9}{-5}{+27} \\
\grouprule
\mname{wan30} & \gnote{30, 30, 30} & \vci{1.11}{0.86}{1.26} & \vci{0.88}{0.73}{0.99} & \vci{0.84}{0.60}{0.98} & \vci{1.07}{0.87}{1.24} & \vci{1.14}{1.00}{1.37} & \vci{+27}{+23}{+30} & \vci{+26}{+21}{+31} & \vci{+38}{+17}{+57} & \vci{+33}{+23}{+42} & \vci{+26}{+21}{+30} \\
\mname{sd20} & \gnote{30, 30, 30} & \vci{0.60}{0.51}{0.75} & \vci{0.75}{0.61}{0.85} & \vci{0.64}{0.53}{0.76} & \vci{0.71}{0.61}{0.90} & \vci{0.79}{0.62}{1.20} & \vci{+19}{+8}{+32} & \vci{+24}{+14}{+35} & \vci{+20}{-6}{+46} & \vci{+19}{+9}{+30} & \vci{+28}{+18}{+37} \\
\mname{sd20mini} & \gnote{30, 30, 30} & \vci{0.59}{0.45}{0.65} & \vci{0.66}{0.49}{0.78} & \vci{0.57}{0.47}{0.61} & \vci{0.54}{0.45}{0.69} & \vci{0.70}{0.59}{0.78} & \vci{+36}{+29}{+44} & \vci{+45}{+40}{+50} & \vci{+37}{+9}{+61} & \vci{+30}{+19}{+41} & \vci{+51}{+43}{+60} \\
\bottomrule
\end{stab}
\end{table*}

\paragraph{Mean slope, relaxation and tempo.}
On the straight track the trend ratio is $-0.10$ to 0.65 against 1 for the stated law, below 0.3 in seven models and 0.62 and 0.65 in the two Seedance models (Table~\ref{tab:supp-dynamics}).
On the undulating levels it is at most 0.42 ($-0.33$ to 0.31 at $f=3$).
On the straight level the exponent $p$ is 0.51--1.62 against 2: LTX-2.5 (distilled) moves at nearly constant speed and the two Seedance models come closest to a uniform acceleration.
The relaxation rate is positive everywhere (0.48--7.2\,s$^{-1}$ on the straight level, 1.3--14.6\,s$^{-1}$ on the undulating levels and 2.0--12.3\,s$^{-1}$ at $f=3$), where the stated law has none, so the speed returns toward a preferred value within 0.07--0.74\,s.
The balanced accuracy of 0.60--0.84 at $f=3$ lies above the monotone baseline (0.60--0.68) and below the physical oracle (0.90--1.00), and Wan~2.2 (A14B) sits at the baseline.
All models are slower than the simulated motion ($\alpha=1.5$--3.2 at $f=3$).

\begin{table*}[t]
\centering
\scriptsize
\caption{Dynamics on the track family: tempo, trend ratio, relaxation rate $b$, straight-ramp exponent $p$ and balanced accuracy (the physical oracle of the same scenes reads 0.90--1.00), with 95\% intervals below each median. $n$: estimable videos at $f=3$ and on the straight level.}
\label{tab:supp-dynamics}
\setlength{\tabcolsep}{4.2pt}
\begin{stab}{hResp}{llccccccc}
\toprule
\shead & & tempo & \multicolumn{2}{c}{trend ratio} & \multicolumn{2}{c}{relaxation rate $b$ \hu{(s$^{-1}$)}} & exponent & accuracy \\
\shead model & $n$ & $\alpha$, $f=3$ & straight & $f=3$ & straight & $f=3$ & $p$, straight & $f=3$ \\
\bandrule
\oraclerow \oraclelabel & & 1 & 1 & 1 & 0 & 0 & 2 & 0.90--1.00 \\
\oraclerule
\mname{ltx23} & \gnote{24, 27} & \vci{2.05}{1.70}{2.61} & \vci{0.15}{0.09}{0.17} & \vci{-0.33}{-0.59}{-0.30} & \vci{4.02}{2.79}{5.07} & \vci{12.33}{9.10}{16.15} & \vci{1.02}{0.95}{1.09} & \vci{0.67}{0.64}{0.72} \\
\mname{ltx25d} & \gnote{25, 30} & \vci{1.55}{1.36}{1.65} & \vci{-0.10}{-0.18}{-0.06} & \vci{0.16}{0.01}{0.27} & \vci{7.22}{4.52}{19.36} & \vci{5.29}{3.02}{10.05} & \vci{0.51}{0.42}{0.73} & \vci{0.84}{0.83}{0.87} \\
\mname{ltx25} & \gnote{24, 26} & \vci{1.63}{1.27}{1.73} & \vci{0.02}{0.02}{0.14} & \vci{0.23}{0.11}{0.33} & \vci{3.29}{1.93}{7.65} & \vci{4.93}{2.96}{6.09} & \vci{0.95}{0.76}{1.03} & \vci{0.75}{0.68}{0.81} \\
\mname{cosmos} & \gnote{21, 27} & \vci{1.53}{1.16}{1.75} & \vci{-0.02}{-0.07}{0.04} & \vci{0.04}{-0.10}{0.20} & \vci{6.73}{4.83}{22.26} & \vci{4.99}{2.47}{8.16} & \vci{0.91}{0.86}{1.01} & \vci{0.71}{0.62}{0.76} \\
\mname{h3} & \gnote{30, 30} & \vci{2.66}{2.44}{2.87} & \vci{0.22}{0.21}{0.25} & \vci{0.19}{0.16}{0.26} & \vci{1.39}{0.96}{1.61} & \vci{2.10}{1.73}{2.23} & \vci{1.19}{1.17}{1.21} & \vci{0.79}{0.74}{0.82} \\
\mname{wan22} & \gnote{28, 29} & \vci{3.17}{2.82}{3.32} & \vci{0.22}{0.06}{0.27} & \vci{0.04}{-0.02}{0.11} & \vci{1.79}{1.26}{3.54} & \vci{2.80}{2.57}{3.39} & \vci{1.14}{1.03}{1.19} & \vci{0.60}{0.57}{0.63} \\
\grouprule
\mname{wan30} & \gnote{30, 30} & \vci{2.32}{2.04}{2.45} & \vci{0.25}{0.22}{0.35} & \vci{-0.02}{-0.04}{0.04} & \vci{1.84}{1.27}{2.11} & \vci{5.67}{5.07}{6.96} & \vci{1.23}{1.19}{1.25} & \vci{0.77}{0.74}{0.79} \\
\mname{sd20} & \gnote{30, 30} & \vci{2.21}{2.06}{2.30} & \vci{0.62}{0.54}{0.80} & \vci{0.31}{0.20}{0.47} & \vci{0.62}{0.46}{0.77} & \vci{1.97}{1.38}{2.95} & \vci{1.54}{1.47}{1.72} & \vci{0.74}{0.70}{0.76} \\
\mname{sd20mini} & \gnote{30, 30} & \vci{1.98}{1.79}{2.28} & \vci{0.65}{0.63}{0.88} & \vci{0.07}{0.02}{0.25} & \vci{0.48}{0.26}{0.74} & \vci{4.62}{3.03}{6.29} & \vci{1.62}{1.54}{1.79} & \vci{0.67}{0.64}{0.70} \\
\bottomrule
\end{stab}
\end{table*}

\paragraph{Mechanism fits.}
Each model's fingerprint was compared with ten candidate dynamics: the stated law, spatial smoothing, temporal smoothing, a response lag, a constant-speed cruise (relaxation at rate $b$ toward a preferred speed with gravity kept), a first-order law in which speed follows the local slope, a curvature slowdown, the cruise with a Gaussian temporal smoothing of width $\tau$, a relaxation with time constant $\tau$ toward a slope-dependent target speed $v^\ast-k\,\beta_0W(s)$ without a physical drive, and a relaxation toward a saturated physical speed.
Each candidate ran on a parameter grid along the drawn tracks of four scenes per level at the model's tempo, first-motion time and frame rate, through the scene camera with 1\,px noise and the same estimator (14{,}400 simulated motions), and 30 components per model (gain, phase, $2f/1f$, gravities, $b$, $g_w$ and accuracy per level, and $p$, $b$ and trend ratio on the straight level) enter $\chi^2$ with $\sigma^2$ the squared floor plus the squared half-width of the observed interval.
The lowest AIC picks the best candidate, and a candidate suffices when $\chi^2/n\le2$ with no component beyond $3\sigma$.
Two families describe the nine models most economically (Table~\ref{tab:supp-mechfit}).
Relaxation to a slope-dependent target speed within 0.2--0.4\,s ($k=0.6$) fits Cosmos 3 Nano, LTX-2.3, LTX-2.5 and Wan~2.2 (A14B).
Cruise with temporal smoothing, a relaxation toward a constant speed at 0.5--3.5\,s$^{-1}$ (time constant 0.3--2\,s) followed by a smoothing of 0.1--0.2\,s, fits LTX-2.5 (distilled), MiniMax-H3, Seedance 2.0, Seedance 2.0 mini and Wan~3.0.
The time-rescaled stated law suffices for none of the nine ($\chi^2/n$ 6.5--27.1) because it keeps the acceleration along the mean slope, and the first-order law scores worst (46--8768).
One mechanism suffices for five models, and the largest deviations share one shape.
On undulating tracks the relaxation rate of MiniMax-H3 and the two Seedance models is 1.6--3.7 times the rate that the fitted mechanism carries over from the straight level, Wan~3.0 relaxes on the straight level at half the carried rate, and the near-constant speed of LTX-2.5 (distilled) on the straight level ($p=0.51$) is produced by no candidate.
The pull toward the preferred speed thus changes with the shape of the track, and no single preferred-speed rule reproduces these models across the levels of the family.
The two families lie within 0.1--0.2 in $\chi^2/n$ for Cosmos 3 Nano and LTX-2.5, and the shared-front tracks place both with the smoothing family (Sec.~\ref{sec:supp-res-locality}), so their label is tentative.
Sec.~\ref{sec:supp-res-energy} carries both families to local edits, ramp pairs and barriers.

\newcommand{\best}[1]{\tile[qRespL]{#1}}
\begin{table*}[t]
\centering
\scriptsize
\caption{Mechanism fits on the track family: best candidate by AIC with its parameters, largest deviation $|z|$, whether one mechanism suffices (\okmark\ yes, \flagmark\ no), and $\chi^2/n$ ($n=30$) of every candidate at its best grid point, with the best candidate tinted. Law: the time-rescaled stated law. Cruise+s.: cruise with temporal smoothing. Target: relaxation to a slope-dependent target speed. Sat.: saturating relaxation.}
\label{tab:supp-mechfit}
\setlength{\tabcolsep}{3pt}
\begin{stab}{hResp}{lllcccccccccccc}
\toprule
\shead & & & & & \multicolumn{10}{c}{$\chi^2/n$ of each candidate} \\
\shead model & best & parameters & $|z|_{\max}$ & suff. & law & spatial & temporal & lag & cruise & 1st order & curvature & cruise+s. & target & sat. \\
\bandrule
\mname{ltx23} & target & $\tau$ 0.2\,s, $k$ 0.6 & 2.4 & \okmark & 18.4 & 18.4 & 4.7 & 18.7 & 1.9 & 168.5 & 16.1 & 1.7 & \best{1.6} & 2.5 \\
\mname{ltx25d} & cruise+s. & $b$ 3.5\,s$^{-1}$, $\tau$ 0.1\,s & 3.3 & \flagmark & 14.8 & 14.4 & 7.1 & 18.2 & 2.1 & 191.9 & 13.6 & \best{1.3} & 1.4 & 4.6 \\
\mname{ltx25} & target & $\tau$ 0.2\,s, $k$ 0.6 & 2.5 & \okmark & 11.5 & 10.9 & 3.7 & 12.9 & 2.4 & 133.8 & 10.5 & 1.2 & \best{1.0} & 2.0 \\
\mname{cosmos} & target & $\tau$ 0.2\,s, $k$ 0.6 & 1.7 & \okmark & 14.5 & 14.2 & 7.1 & 14.6 & 1.0 & 45.9 & 13.8 & 0.8 & \best{0.7} & 1.5 \\
\mname{h3} & cruise+s. & $b$ 1\,s$^{-1}$, $\tau$ 0.2\,s & 3.2 & \flagmark & 25.2 & 23.8 & 7.6 & 24.3 & 6.4 & 6355.5 & 24.4 & \best{1.4} & 1.8 & 2.3 \\
\mname{wan22} & target & $\tau$ 0.4\,s, $k$ 0.6 & 1.6 & \okmark & 13.4 & 13.2 & 6.8 & 13.5 & 2.1 & 1493.5 & 12.2 & 1.2 & \best{0.7} & 1.6 \\
\grouprule
\mname{wan30} & cruise+s. & $b$ 3.5\,s$^{-1}$, $\tau$ 0.1\,s & 3.7 & \flagmark & 27.1 & 27.3 & 12.5 & 29.9 & 2.6 & 2375.6 & 26.6 & \best{1.9} & 3.0 & 7.5 \\
\mname{sd20} & cruise+s. & $b$ 0.5\,s$^{-1}$, $\tau$ 0.2\,s & 2.4 & \okmark & 6.5 & 5.9 & 1.6 & 6.8 & 3.5 & 8768.3 & 5.9 & \best{0.9} & 1.6 & 1.8 \\
\mname{sd20mini} & cruise+s. & $b$ 0.5\,s$^{-1}$, $\tau$ 0.2\,s & 3.1 & \flagmark & 12.1 & 11.8 & 3.0 & 14.1 & 8.1 & 3812.2 & 10.6 & \best{2.2} & 4.7 & 5.4 \\
\bottomrule
\end{stab}
\end{table*}

\subsection{Contacts and composition on the deflector family}
\label{sec:supp-res-composition}

At the first contact the closed models bounce, MiniMax-H3 bounces in part and three open-weight models slide along the deflector.
On three deflectors the physical order of contacts appears in at most a quarter of any model's videos, each contact errs on its own, and most models still bring the ball into the named tray.

\paragraph{First contact.}
Pooled over all 24 deflector scenes, the closed models reach an apparent normal restitution of 0.67--0.74 against the prescribed 0.85 (band 0.80--0.90) and leave 16--35\degree{} from the simulator's exit for the measured incoming direction (Table~\ref{tab:supp-firstcontact}).
MiniMax-H3 reads 0.45 and 23\degree, and Cosmos 3 Nano, LTX-2.5 (distilled) and Wan~2.2 (A14B) slide (0.10--0.21, exits 57--78\degree{} away).
On the single deflector MiniMax-H3 bounces as much as the closed models (0.66).
There the exit direction turns with the deflector angle at 0.44--2.07\degree{} per degree against the physical 2.06, and the interval excludes 0 only for MiniMax-H3 and Wan~2.2 (A14B).
LTX-2.3 has one estimable contact and is left uninterpreted.

\newcommand{\enbar}[4]{\ivband{0.80}{0.90}\ivbar[#1]{#2}{#3}{#4}{0.85}{0}{1}}
\begin{table*}[t]
\centering
\perarmhead\\[-0.5pt]
\perarm{224}{259.3}
\vspace{3pt}
\caption{Contact dynamics on the deflector family. Top: exit-angle residual by contact index and first-contact restitution by layout. Bottom: restitution, exit residual and angle-response slope by model, with 95\% intervals where estimated and denominators. Grey bands show the physical restitution range in the table and the measurement floors in the charts; floor values are reported in Table~\ref{tab:supp-floors-deflector}.}
\label{tab:supp-firstcontact}
\scriptsize
\setlength{\tabcolsep}{3.4pt}
\begin{stab}{hResp}{lcccccccc}
\toprule
\shead & \multicolumn{4}{c}{first-contact restitution $e_n$} & \multicolumn{2}{c}{residual \hu{(\degree)}} & exit slope \\
\shead model & all scenes & & one deflector & three deflectors & all scenes & one deflector & one deflector \\
\bandrule
\oraclerow \oraclelabel & 0.85 \gnote{(0.80--0.90)} & & 0.85 & 0.85 & 0 & 0 & 2.06 \\
\oraclerule
\mname{ltx23} & 0.35\,\nn{1} & \nd & \nd\,\nn{0} & 0.35\,\nn{1} & 17 & \nd & \nd\,\nn{0} \\
\mname{ltx25d} & \vcin{0.21}{0.19}{0.27}{46} & \enbar{mGlacierD}{0.19}{0.21}{0.27} & 0.21\,\nn{15} & 0.23\,\nn{31} & \vci{57}{45}{65} & 58 & \vcin{0.69}{-0.86}{3.01}{15} \\
\mname{cosmos} & \vcin{0.10}{0.00}{0.41}{15} & \enbar{mSlateD}{0.00}{0.10}{0.41} & 0.09\,\nn{4} & 0.10\,\nn{11} & \vci{68}{40}{83} & 61 & \vcin{4.60}{4.19}{5.02}{4} \\
\mname{h3} & \vcin{0.45}{0.36}{0.54}{61} & \enbar{mLavD}{0.36}{0.45}{0.54} & 0.66\,\nn{19} & 0.41\,\nn{42} & \vci{23}{18}{29} & 23 & \vcin{1.04}{0.44}{1.59}{19} \\
\mname{wan22} & \vcin{0.15}{0.12}{0.17}{42} & \enbar{mOrchidD}{0.12}{0.15}{0.17} & 0.14\,\nn{19} & 0.15\,\nn{23} & \vci{78}{70}{92} & 70 & \vcin{2.07}{0.58}{4.02}{19} \\
\grouprule
\mname{wan30} & \vcin{0.67}{0.56}{0.75}{62} & \enbar{mRoseD}{0.56}{0.67}{0.75} & 0.50\,\nn{21} & 0.75\,\nn{41} & \vci{16}{10}{22} & 29 & \vcin{1.23}{-0.01}{2.79}{21} \\
\mname{sd20} & \vcin{0.74}{0.60}{0.81}{52} & \enbar{mCoralD}{0.60}{0.74}{0.81} & 0.70\,\nn{18} & 0.75\,\nn{34} & \vci{35}{28}{46} & 47 & \vcin{0.60}{-0.26}{1.89}{18} \\
\mname{sd20mini} & \vcin{0.67}{0.56}{0.78}{59} & \enbar{mPeachD}{0.56}{0.67}{0.78} & 0.41\,\nn{17} & 0.78\,\nn{42} & \vci{21}{16}{28} & 21 & \vcin{0.44}{-0.11}{1.61}{17} \\
\bottomrule
\end{stab}
\end{table*}

\paragraph{Composition.}
With three deflectors (regular and irregular layouts pooled, 16 scenes) the physical order appears in 12--24\% of the trackable videos of the closed models and in none of the open-weight models' videos, while exact renders reproduce it every time (Table~\ref{tab:supp-contacts}).
In the two Seedance models the exit is off by 19--31\degree{} at the first, second and third contact alike, and in Wan~3.0 the residual grows from 10\degree{} at the first contact to 22 and 28\degree{} at the second and third.
MiniMax-H3, Wan~2.2 (A14B) and Cosmos 3 Nano seldom reach a second contact (2 of 42, 8 of 23 and 4 of 11 first contacts).
A contact law fitted on each model's single-deflector contacts and then frozen improves the three-deflector prediction for Seedance 2.0 only (26.4 to 22.1\degree{} on regular and 30.9 to 24.8\degree{} on irregular layouts).
A consistent material of the wrong kind would be absorbed by this calibration, so the contact errors of the other models depend on the layout.

\newcommand{\ordbar}[4]{\ivbar[#1]{#2}{#3}{#4}{100}{0}{100}}
\newcommand{\calib}[2]{#1\,$\to$\,#2}
\begin{table*}[t]
\centering
\scriptsize
\caption{Composition on the three-deflector layouts (regular and irregular pooled, prompt naming the tray). Exact order: share of trackable videos whose merged contact sequence equals the physical one, with 95\% interval and counts, drawn on 0 to 100\% with the physical 100\% as a tick. Residual at contact $k=1,2,3$ with the number of contacts. Calibrated law: median residual under the stated law and under the law fitted on the model's single-deflector contacts, on regular and irregular layouts. Tray: share of trackable videos that end in the tray. Tinted: the one model for which the calibrated law improves the prediction. Floor of the residual 2.0--4.2\degree{} (90th percentile).}
\label{tab:supp-contacts}
\setlength{\tabcolsep}{3.6pt}
\begin{stab}{hState}{lcccccccc}
\toprule
\shead & \multicolumn{2}{c}{exact order} & \multicolumn{3}{c}{residual at contact $k$ \hu{(\degree)}} & \multicolumn{2}{c}{stated $\to$ calibrated law \hu{(\degree)}} & tray \\
\shead model & & share & $k=1$ & $k=2$ & $k=3$ & regular & irregular & \\
\bandrule
\oraclerow \oraclelabel & & 100\% & 0 & 0 & 0 & 0 & 0 & 81--98\% \gnote{of members} \\
\oraclerule
\mname{ltx23} & \ordbar{mIcecyanD}{0}{0}{0} & 0\%\,\nn{0/13} & 17\,\nn{1} & \nd & \nd & \nd & \nd & 100\%\,\nn{13} \\
\mname{ltx25d} & \ordbar{mGlacierD}{0}{0}{0} & 0\%\,\nn{0/40} & 56\,\nn{31} & 58\,\nn{25} & 39\,\nn{18} & \calib{47.0}{47.4} & \calib{50.5}{51.9} & 100\%\,\nn{40} \\
\mname{cosmos} & \ordbar{mSlateD}{0}{0}{0} & 0\%\,\nn{0/34} & 71\,\nn{11} & 15\,\nn{4} & 20\,\nn{1} & \calib{37.3}{72.3} & \calib{70.8}{38.3} & 100\%\,\nn{34} \\
\mname{h3} & \ordbar{mLavD}{0}{0}{0} & 0\%\,\nn{0/46} & 23\,\nn{42} & 27\,\nn{2} & \nd & \calib{24.7}{24.7} & \calib{21.3}{21.0} & 98\%\,\nn{46} \\
\mname{wan22} & \ordbar{mOrchidD}{0}{0}{0} & 0\%\,\nn{0/36} & 78\,\nn{23} & 50\,\nn{8} & 4\,\nn{4} & \calib{77.6}{80.5} & \calib{68.2}{72.6} & 47\%\,\nn{36} \\
\grouprule
\mname{wan30} & \ordbar{mRoseD}{2}{12}{23} & \vcin{12\%}{2}{23}{5/42} & 10\,\nn{41} & 22\,\nn{39} & 28\,\nn{14} & \calib{17.4}{18.9} & \calib{16.8}{18.0} & 100\%\,\nn{42} \\
\mname{sd20} & \ordbar{mCoralD}{11}{24}{38} & \vcin{24\%}{11}{38}{8/34} & 28\,\nn{34} & 31\,\nn{27} & 27\,\nn{20} & \tile[qStateL]{\calib{26.4}{22.1}} & \tile[qStateL]{\calib{30.9}{24.8}} & 100\%\,\nn{34} \\
\mname{sd20mini} & \ordbar{mPeachD}{5}{14}{26} & \vcin{14\%}{5}{26}{6/43} & 21\,\nn{42} & 26\,\nn{35} & 19\,\nn{26} & \calib{23.4}{23.7} & \calib{19.2}{19.6} & 100\%\,\nn{43} \\
\bottomrule
\end{stab}
\end{table*}

\paragraph{Destination against route.}
Under the prompt that names the outcome the ball reaches the tray in 98--100\% of the trackable three-deflector videos of seven models whatever the route, and in 47\% of those of Wan~2.2 (A14B) (Table~\ref{tab:supp-contacts}).
The geometry leaves little room for a wrong route.
The ensemble rests in the tray in 81--98\% of its members on these layouts, and with any one deflector removed in 0--30\% (medians 2\% on regular and 13\% on irregular layouts).
Continued by the stated law from the state in which the generated ball leaves its last correct contact (or the launch ramp), the ball reaches the tray in 11--84\% of the wrong-route videos of six of these models, from 11\% for MiniMax-H3 to 84\% for Wan~3.0 (Table~\ref{tab:supp-destination}).
The models exceed it by $+16$ to $+87$ percentage points with intervals that exclude 0, LTX-2.3 by $+8$ on 13 videos, and rescaling the exit speed by the model's apparent gravity leaves $+29$ to $+43$ points.
Wan~2.2 (A14B) goes the other way: its ball slides along the first deflector and often stops short of the tray.
This table covers all named-prompt videos, and Sec.~\ref{sec:supp-res-outcome} pairs them with the neutral prompt.

\begin{table}[t]
\centering
\scriptsize
\caption{Destination counterfactual on the three-deflector layouts, all named-prompt videos. Simulator: tray arrival of the stated law rolled forward from the model's exit state after its last correct contact (videos used / wrong-route videos). Difference: model minus simulator in percentage points with 95\% interval, then with the exit speed rescaled by the launch tempo and by the apparent gravity (videos used).}
\label{tab:supp-destination}
\setlength{\tabcolsep}{2.4pt}
\begin{stab}{hGoal}{lcccc}
\toprule
\shead & & \multicolumn{3}{c}{model minus simulator \hu{(points)}} \\
\shead model & simulator & difference & tempo-scaled & gravity-scaled \\
\bandrule
\mshort{ltx23} & 92\%\,\nn{13/13} & \vci{+8}{+0}{+21} & \snum{+33}\,\nn{3} & \snum{+33}\,\nn{3} \\
\mshort{ltx25d} & 68\%\,\nn{40/40} & \vci{+32}{+16}{+50} & \snum{+42}\,\nn{24} & \snum{+31}\,\nn{35} \\
\mshort{cosmos} & 65\%\,\nn{34/34} & \vci{+35}{+24}{+47} & \snum{+40}\,\nn{15} & \snum{+39}\,\nn{23} \\
\mshort{h3} & 11\%\,\nn{46/46} & \vci{+87}{+72}{+100} & \snum{+43}\,\nn{46} & \snum{+36}\,\nn{11} \\
\mshort{wan22} & 59\%\,\nn{27/36} & \vci{-22}{-44}{-3} & \snum{-8}\,\nn{24} & \snum{+29}\,\nn{7} \\
\grouprule
\mshort{wan30} & 84\%\,\nn{37/37} & \vci{+16}{+5}{+29} & \snum{+14}\,\nn{37} & \snum{+29}\,\nn{28} \\
\mshort{sd20} & 50\%\,\nn{26/26} & \vci{+50}{+33}{+67} & \snum{+68}\,\nn{25} & \snum{+31}\,\nn{26} \\
\mshort{sd20mini} & 30\%\,\nn{37/37} & \vci{+70}{+55}{+85} & \snum{+76}\,\nn{37} & \snum{+43}\,\nn{37} \\
\bottomrule
\end{stab}
\end{table}

\subsection{Locality: local edits and shared-front tracks}
\label{sec:supp-res-locality}

Only MiniMax-H3 changes speed at an edit in the physical direction while the motion before it stays unchanged, by a fifth of the physical change at the bump.
In three open-weight models an edit ahead of the ball changes its path before the ball arrives beyond the null edit, and on shared-front tracks most models follow the drawn back-section phase with a lead of their own.

\paragraph{Local edits, open-weight models.}
All arms of a track share the seed, so edit-induced differences can be read relative to the rerun and null-edit floors (Table~\ref{tab:supp-local-open}).
MiniMax-H3 slows by 2.4\% at a bump (simulator $-11.4\%$) and speeds up by 1.0\% at a dip after normalizing by the pre-edit tempo (2.4\% raw, simulator $+7.8\%$), with the time before the edit changed by at most 0.7\%.
Its arrival shift is 0.42 of the physical one, and its arrival is 0.7--0.8\% late after both edits.
LTX-2.5 reaches either edit 3--6\% sooner and slows at the dip ($-6.1\%$, $[-8.6,-0.8]$), where the simulated ball speeds up, while its reading at the bump ($-2.1\%$, $[-10.6,+7.5]$) includes zero.
Wan~2.2 (A14B) slows by 29\% at the bump, 2.6 times the physical size, from before the bump on (time to the edit $+15.3\%$), and all ten of its pairs come from the one seed that reaches the stop within the clip.
The null edit of LTX-2.5 (distilled) changes $dv$ by $+4.1\%$ $[+2.1,+5.3]$ and is flagged, so its readings are not specific to the edit, and LTX-2.3 and Cosmos 3 Nano have too few pairs for a size reading.
Before the ball reaches it, a bump or dip changes the path 1.7 to 4.0 times as much as the null edit of the same size, and the intervals separate from the null for LTX-2.3 and LTX-2.5 (bump and dip) and MiniMax-H3 (dip).

\newcommand{\vcir}[4]{\makecell[c]{\snum{#1} \gnote{(\snum{#4})}\\[-2.4pt]\ci{#2}{#3}}}
\newcommand{\sep}{$^\ast$}
\begin{table*}[t]
\centering
\scriptsize
\caption{Local edits, open-weight models (percent unless marked), pairs of the same track and seed. $dv$: speed change at the edit relative to the base video, normalized by the speed before the edit, with the unnormalized (raw) change at the dip in grey. Arrival: the normalized arrival shift of the bump over its physical value (1.07--1.10 on exact renders). $T_{\mathrm{pre}}$: change of the time before the edit, bump / dip. Path: change of the path before the edit over that under the null edit, bump / dip, with $^\ast$ where the interval separates from the null. Null: null-edit $dv$, with $^\dagger$ marking a flag. Floors: rerun 0 for every model except LTX-2.3 (0.15\%), measurement 1.0\% ($dv$) and 0.4\% ($dt$).}
\label{tab:supp-local-open}
\setlength{\tabcolsep}{4.4pt}
\begin{stab}{hState}{lccccccc}
\toprule
\shead model & pairs & $dv$ bump & $dv$ dip \hu{(raw)} & arrival / physical & $T_{\mathrm{pre}}$ & path / null & null $dv$ \\
\bandrule
\oraclerow \oraclelabel & & \snum{-11.0} to \snum{-11.7} & \snum{+7.7} to \snum{+8.0} & 1 & 0, 0 & \nd & 0 \\
\oraclerule
\mname{ltx23} & \gnote{10, 16} & \vci{+4.1}{0.0}{+11.9} & \vcir{-1.1}{-8.9}{+3.7}{-3.9} & \vci{-1.57}{-4.67}{0.33} & \snum{-0.2}, \snum{-0.2} & 2.1\sep, 2.2\sep & \snum{+2.5} \\
\mname{ltx25d} & \gnote{30, 31} & \vci{-3.9}{-13.9}{+0.9} & \vcir{+1.9}{-2.2}{+6.3}{-1.4} & \vci{0.34}{-1.76}{2.74} & \snum{-5.2}, \snum{-3.5} & 2.8, 2.1 & \snum{+4.1}$^\dagger$ \\
\mname{ltx25} & \gnote{19, 27} & \vci{-2.1}{-10.6}{+7.5} & \vcir{-6.1}{-8.6}{-0.8}{-4.4} & \vci{2.93}{0.48}{5.11} & \snum{-3.1}, \snum{-5.8} & 3.6\sep, 4.0\sep & \snum{-0.5} \\
\mname{cosmos} & \gnote{11, 11} & \vci{-2.6}{-27.0}{+7.4} & \vcir{-0.1}{-7.4}{+3.0}{-1.9} & \vci{4.26}{2.22}{6.63} & \snum{-0.9}, \snum{-0.5} & 3.0, 2.2 & \snum{-1.2} \\
\mname{h3} & \gnote{35, 36} & \tilec[qStateL]{\vci{-2.4}{-3.6}{-0.7}} & \vcir{+1.0}{-0.3}{+2.9}{+2.4} & \vci{0.42}{0.08}{0.78} & \snum{-0.7}, \snum{-0.1} & 2.0, 1.7\sep & \snum{-0.7} \\
\mname{wan22} & \gnote{10, 9} & \vci{-29.3}{-37.2}{-12.6} & \vcir{-3.9}{-32.8}{-0.7}{-0.8} & \vci{1.90}{-0.15}{4.08} & \snum{+15.3}, \snum{-0.9} & 1.1, 2.0 & \snum{-8.2} \\
\bottomrule
\end{stab}
\end{table*}

\paragraph{Local edits, closed models.}
Without seed control the arms are independent samples that also differ in tempo.
On the three closed-model tracks Seedance 2.0 and Seedance 2.0 mini speed up at the dip by $+9.3\%$ $[+3.4,+29.1]$ and $+12.3\%$ $[+5.6,+13.1]$ (simulator $+7.4\%$), in the physical direction and by at least the physical amount, with the time before the dip changed by $-3.2$ and $+0.6\%$.
Their slowdown at the bump ($-5.0\%$ $[-12.4,+0.5]$ and $-1.5\%$ $[-1.8,+4.9]$, simulator $-10.6\%$) has intervals that include zero.
Wan~3.0 has no reading, since its arms differ in the time to the edit by up to 14\%.

\paragraph{Shared-front tracks.}
LTX-2.5 (distilled), LTX-2.5 and Wan~3.0 follow the drawn back-section phase with slopes whose intervals contain 1, Seedance 2.0 mini follows it with a slope of 0.80, and Seedance 2.0 and LTX-2.3 have slopes near 1 with intervals that also contain 0 (Table~\ref{tab:supp-sharedfront}).
Their leads have the sign and size of the track-family phase, and LTX-2.5 (distilled), Seedance 2.0 mini and LTX-2.3 lead by 17--25\degree{} more.
MiniMax-H3 does not follow the back section as a whole (slope $-0.11$): the exploratory decomposition splits its response into a part locked to the drawn phase (0.41 of the physical amplitude, leading by 17\degree) and a larger part with the same phase on every track (0.55).
Such a fixed part arises only with temporal smoothing, which is why the shared-front tracks place MiniMax-H3, Cosmos 3 Nano, LTX-2.5 and Seedance 2.0 with the smoothing family.
In Wan~3.0, Seedance 2.0 mini and LTX-2.5 the lead grows from the front to the back section, where the ball is faster, by 9--22\degree, while both fitted families predict that it shrinks.
Wan~2.2 (A14B) shows no measurable back-section response.
The twelve front sections are drawn identically, and the motion over them spreads by 4.8--27.6\,mm across tracks against a floor of 1.6--1.7\,mm, the same kind of change that a null edit produces.

\newcommand{\slopebar}[4]{\ivbar[#1]{#2}{#3}{#4}{1}{-0.5}{1.5}}
\begin{table*}[t]
\centering
\scriptsize
\caption{Shared-front tracks. Slope of the back-section phase on the drawn back phase (1: follows the drawing, 0: continues the front) with its resultant length $R$, drawn on $-0.5$ to 1.5 with the physical 1 as a tick, and the track-family phase at $f=3$ for comparison. Exploratory: locked part $|G|$ and its lead, fixed part $|C|$ (in brackets the physical oracle at the model's tempo), front-section lead and its change to the back section. Front: spread of the motion over the identical front sections in mm (floor 1.6--1.7). Medians with 95\% intervals. $n$: estimable videos.}
\label{tab:supp-sharedfront}
\setlength{\tabcolsep}{3.2pt}
\begin{stab}{hState}{lccccccccccc}
\toprule
\shead & & \multicolumn{3}{c}{back-section phase slope} & track & \multicolumn{5}{c}{exploratory} & front \\
\shead model & $n$ & & slope & $R$ & phase & $|G|$ & lead of $G$ & $|C|$ & front lead & back $-$ front & \hu{(mm)} \\
\bandrule
\oraclerow \oraclelabel & & & 1.00 & & 0 & 1 & 0 & 0 & 0 & 0 & 0 \\
\oraclerule
\mname{ltx23} & \gnote{30} & \slopebar{mIcecyanD}{-0.21}{1.06}{1.18} & \vci{1.06}{-0.21}{1.18} & 0.62 & \snum{+38} & \vci{0.90}{0.49}{1.29} & \vci{+59}{+37}{+80} & \vci{0.16}{0.06}{0.64} \gnote{(0.07)} & \snum{+54} & \vci{+12}{-23}{+73} & 19.5 \\
\mname{ltx25d} & \gnote{29} & \slopebar{mGlacierD}{0.97}{1.05}{1.19} & \vci{1.05}{0.97}{1.19} & 0.83 & \snum{+35} & \vci{0.80}{0.63}{1.10} & \vci{+60}{+51}{+70} & \vci{0.14}{0.05}{0.34} \gnote{(0.22)} & \snum{+55} & \vci{-4}{-25}{+23} & 27.6 \\
\mname{ltx25} & \gnote{34} & \slopebar{mPeriD}{0.87}{0.95}{1.01} & \vci{0.95}{0.87}{1.01} & 0.85 & \snum{+36} & \vci{0.86}{0.67}{1.04} & \vci{+27}{+19}{+37} & \vci{0.32}{0.23}{0.45} \gnote{(0.09)} & \snum{+19} & \vci{+9}{-1}{+17} & 14.4 \\
\mname{cosmos} & \gnote{21} & \slopebar{mSlateD}{-0.50}{-0.28}{1.50} & \vci{-0.28}{-0.50}{1.50} & 0.52 & \snum{+41} & \vci{0.51}{0.36}{0.80} & \vci{+23}{+1}{+50} & \vci{0.48}{0.30}{0.75} \gnote{(0.17)} & \snum{+39} & \vci{-6}{-48}{+68} & 11.3 \\
\mname{h3} & \gnote{36} & \slopebar{mLavD}{-0.24}{-0.11}{0.06} & \vci{-0.11}{-0.24}{0.06} & 0.68 & \snum{-3} & \vci{0.41}{0.35}{0.47} & \vci{+17}{+5}{+26} & \vci{0.55}{0.47}{0.62} \gnote{(0.04)} & \snum{+63} & \vci{-22}{-86}{+21} & 9.3 \\
\mname{wan22} & \gnote{32} & \slopebar{mOrchidD}{-0.50}{-0.50}{1.50} & \vci{-0.50}{-0.50}{1.50} & 0.30 & \snum{+15} & \vci{0.09}{0.03}{0.38} & \nd & \vci{0.05}{0.02}{0.35} \gnote{(0.09)} & \snum{+102} & \nd & 4.8 \\
\grouprule
\mname{wan30} & \gnote{18} & \slopebar{mRoseD}{0.97}{1.07}{1.18} & \vci{1.07}{0.97}{1.18} & 0.98 & \snum{+27} & \vci{0.79}{0.67}{0.92} & \vci{+32}{+24}{+39} & \vci{0.19}{0.08}{0.35} \gnote{(0.01)} & \snum{+18} & \vci{+17}{+5}{+28} & 9.6 \\
\mname{sd20} & \gnote{18} & \slopebar{mCoralD}{0.00}{0.92}{1.50} & \vci{0.92}{0.00}{1.50} & 0.69 & \snum{+19} & \vci{0.67}{0.37}{1.02} & \vci{+28}{+12}{+43} & \vci{0.30}{0.09}{0.64} \gnote{(0.08)} & \snum{+21} & \vci{+9}{-145}{+103} & 25.1 \\
\mname{sd20mini} & \gnote{18} & \slopebar{mPeachD}{0.71}{0.80}{0.89} & \vci{0.80}{0.71}{0.89} & 0.88 & \snum{+36} & \vci{0.54}{0.44}{0.67} & \vci{+53}{+36}{+67} & \vci{0.17}{0.05}{0.31} \gnote{(0.04)} & \snum{+30} & \vci{+22}{+3}{+43} & 17.5 \\
\bottomrule
\end{stab}
\end{table*}

\subsection{Energy: ramp pairs, barriers and chutes}
\label{sec:supp-res-energy}

The generated ball climbs above its release height, crosses barriers higher than its release point at every height tested, and keeps a crest speed that stays flat as the barrier rises.
The fitted preferred-speed rules of Sec.~\ref{sec:supp-res-response} reproduce none of the models on the track, the local edit and the ramp pair together.

\paragraph{Ramp pairs.}
Of the 72 readable ascents, 63 rise above the release height (61 by more than 10\%, against a single-video floor of 0.055--0.091), where the simulated ball turns around at 0.93--0.98 (Table~\ref{tab:supp-ramp-results}).
Thirty-seven reach the top stop, 12 are still climbing at the end of the clip, 10 stop on the slope and 13 turn back, 7 of those above the release height.
Seven balls leave the track at the top and keep rising, 10--365\,mm above the top stop (two in LTX-2.5 (distilled), three in LTX-2.5, two in Wan~2.2 (A14B)).
The mean height ratio is 1.15--2.36 across models against 0.96 for the physical oracle (0.97 for the two angles of the closed models).
The 18 unreadable climbs also carry a height ratio and are left out of every count.
The ratio rises with the angle because a steep ascent ends at a higher stop and a shallow one outlasts the clip, and the physical dependence on the angle (0.929 to 0.980) is below the resolution at model tempo.

\newcommand{\hrbar}[4]{\ivbar[#1]{#2}{#3}{#4}{0.962}{0}{4}}
\newcommand{\angcell}[5]{\gpad{\makecell[c]{#1\\[-1.6pt]\outbar{#2}{#3}{#4}{#5}}}}
\begin{table*}[t]
\centering
\perarmhead\\[-0.5pt]
\perarm{352}{131.3}
\vspace{3pt}
\caption{Energy. Top: rows 6 and 7 of Fig.~\ref{fig:supp-perarm}, the height reached over the release height by uphill angle (one mark per video) and the share of barriers crossed by height. Bottom: ramp pairs, readable climbs. Mean height ratio over ascents with 95\% interval (drawn on 0 to 4 with the physical 0.962 as a tick), ascents above the release height as dots, and per angle the median ratio over one block per climb: \okey{qStateD}{top stop}, \okey{qStateM}{still climbing}, \okey{qIntM}{stopped on the slope}, \okey{qGoalM}{turned back} (top-stop ratio in the header). Physical oracle: nominal turning height with the 64-member band. Last column: flat ascents reaching the end stop. The closed models ran 60, 15 and 0\degree.}
\label{tab:supp-ramp-results}
\scriptsize
\setlength{\tabcolsep}{3.8pt}
\begin{stab}{hState}{lcccccccc}
\toprule
\shead & \multicolumn{2}{c}{mean height ratio} & above release & \multicolumn{4}{c}{per uphill angle \hu{(top-stop ratio)}} & 0\degree \\
\shead model & & & & 60\degree{} \hu{(2.52)} & 30\degree{} \hu{(1.61)} & 15\degree{} \hu{(1.46)} & 5\degree{} \hu{(1.32)} & end stop \\
\bandrule
\oraclerow \oraclelabel & & 0.962 \gnote{(closed 0.973)} & 0 & 0.980 \gnote{(0.92--1.00)} & 0.975 \gnote{(0.90--1.00)} & 0.965 \gnote{(0.87--1.00)} & 0.929 \gnote{(0.76--1.00)} & reaches \\
\oraclerule
\mname{ltx23} & \hrbar{mIcecyanD}{1.15}{1.23}{1.30} & \vci{1.23}{1.15}{1.30} & \pips{6}{7} & \nd & \angcell{1.48}{0}{0}{0}{2} & \angcell{1.21}{0}{0}{3}{0} & \angcell{1.00}{0}{1}{1}{0} & \pips{0}{3} \\
\mname{ltx25d} & \hrbar{mGlacierD}{1.54}{2.36}{3.70} & \vci{2.36}{1.54}{3.70} & \pips{11}{12} & \angcell{2.62}{3}{0}{0}{0} & \angcell{1.44}{1}{0}{1}{1} & \angcell{1.33}{1}{0}{0}{2} & \angcell{1.27}{0}{0}{3}{0} & \pips{1}{3} \\
\mname{ltx25} & \hrbar{mPeriD}{1.14}{1.60}{2.06} & \vci{1.60}{1.14}{2.06} & \pips{5}{6} & \angcell{1.61}{1}{0}{0}{1} & \angcell{1.95}{2}{0}{0}{0} & \angcell{1.20}{0}{0}{1}{0} & \angcell{1.25}{0}{1}{0}{0} & \pips{1}{2} \\
\mname{cosmos} & \hrbar{mSlateD}{0.80}{1.15}{1.49} & \vci{1.15}{0.80}{1.49} & \pips{6}{10} & \angcell{1.56}{1}{0}{0}{1} & \angcell{1.42}{1}{0}{0}{2} & \angcell{1.34}{2}{0}{0}{1} & \angcell{0.73}{1}{0}{0}{1} & \pips{1}{1} \\
\mname{h3} & \hrbar{mLavD}{1.48}{1.56}{1.64} & \vci{1.56}{1.48}{1.64} & \pips{12}{12} & \angcell{2.48}{2}{1}{0}{0} & \angcell{1.24}{1}{2}{0}{0} & \angcell{1.41}{2}{0}{1}{0} & \angcell{1.22}{0}{3}{0}{0} & \pips{1}{3} \\
\mname{wan22} & \hrbar{mOrchidD}{1.55}{1.76}{1.95} & \vci{1.76}{1.55}{1.95} & \pips{9}{11} & \angcell{3.30}{3}{0}{0}{0} & \angcell{1.60}{3}{0}{0}{0} & \angcell{1.28}{1}{1}{0}{0} & \angcell{0.73}{1}{2}{0}{0} & \pips{1}{3} \\
\grouprule
\mname{wan30} & \hrbar{mRoseD}{1.84}{1.94}{2.00} & \vci{1.94}{1.84}{2.00} & \pips{5}{5} & \angcell{2.40}{2}{0}{0}{1} & \notrun[10mm] & \angcell{1.39}{2}{0}{0}{0} & \notrun[10mm] & \pips{3}{3} \\
\mname{sd20} & \hrbar{mCoralD}{1.51}{1.66}{1.80} & \vci{1.66}{1.51}{1.80} & \pips{3}{3} & \angcell{2.46}{1}{0}{0}{0} & \notrun[10mm] & \angcell{1.25}{1}{1}{0}{0} & \notrun[10mm] & \pips{0}{3} \\
\mname{sd20mini} & \hrbar{mPeachD}{1.61}{1.82}{1.94} & \vci{1.82}{1.61}{1.94} & \pips{6}{6} & \angcell{2.40}{2}{0}{0}{1} & \notrun[10mm] & \angcell{1.44}{3}{0}{0}{0} & \notrun[10mm] & \pips{2}{3} \\
\bottomrule
\end{stab}
\end{table*}

\paragraph{Barriers.}
No model shows a threshold (Table~\ref{tab:supp-barrier-results}).
The ensemble crosses with probability 1.00, 1.00, 0.42, 0.16, 0 and 0 at hump heights of 0.70, 0.85, 0.95, 0.971, 1.05 and 1.20 of the release drop.
Six models cross every barrier they were given (the closed models ran three heights), including the two that stand 5\% and 20\% above the release drop, and across models the ball crosses 43 of the 49 impassable humps it reaches and fails at 7 of the 72 passable ones.
Over the humps higher than the release drop the crossing ball of eight models keeps 0.33--0.87 of its speed at the foot, independent of the height, where the physical oracle falls from 0.53 through 0.35, 0.15 and 0.03 to a turn-around.
The ball slows on the hump by an amount unrelated to the height it climbs.
The outcome of Wan~2.2 (A14B) follows the seed: seed 0 crosses all six barriers, seed 1 turns back low on all six, seed 2 stops on the hump or does not reach it.

\newcommand{\bx}[3]{\gpad{\makecell[c]{\pips{#1}{#2}\\[-1.6pt]#3}}}
\newcommand{\impass}[1]{\tilew{qStateL!80}{9.6mm}{#1}}
\newcommand{\impassl}[1]{\tilebox{qStateL!80}{9.6mm}{1.1pt}{#1}}
\begin{table*}[t]
\centering
\scriptsize
\caption{Barriers. Per height ratio: crossings among the readable videos as dots (filled: crossed) over the median crest-speed ratio. Tinted: the barriers the stated law does not cross. Forbidden: crossings of the barriers the stated law does not cross (1.05, 1.20), with 0.95 and 0.971 counted by their nominal outcome. Missed: failures on barriers the stated law crosses. 50\%: logistic crossing point. The closed models ran 0.85, 1.05 and 1.20.}
\label{tab:supp-barrier-results}
\setlength{\tabcolsep}{4.2pt}
\begin{stab}{hState}{lccccccccc}
\toprule
\shead & \multicolumn{6}{c}{barrier height over release drop} & & & \\
\shead model & 0.70 & 0.85 & 0.95 & 0.971 & 1.05 & 1.20 & forbidden & missed & 50\% \\
\bandrule
\oraclerow \oraclelabel & pass, 0.53 & pass, 0.35 & 27/64, 0.15 & 10/64, 0.03 & \impassl{turn back} & \impassl{turn back} & 0 & 0 & 0.971--1.05 \\
\oraclerule
\mname{ltx23} & \bx{3}{3}{0.10} & \bx{3}{3}{0.52} & \bx{3}{3}{0.59} & \bx{3}{3}{0.54} & \impass{\bx{3}{3}{0.45}} & \impass{\bx{3}{3}{0.41}} & 6/6 & 0/12 & none \\
\mname{ltx25d} & \bx{3}{3}{0.63} & \bx{3}{3}{0.51} & \bx{3}{3}{0.52} & \bx{3}{3}{0.52} & \impass{\bx{3}{3}{0.61}} & \impass{\bx{3}{3}{0.61}} & 6/6 & 0/12 & none \\
\mname{ltx25} & \bx{2}{2}{0.16} & \bx{2}{2}{0.47} & \bx{2}{2}{0.16} & \bx{2}{2}{0.15} & \impass{\bx{2}{2}{0.47}} & \impass{\bx{1}{2}{0.87}} & 3/4 & 0/8 & 1.20 \\
\mname{cosmos} & \bx{2}{2}{0.55} & \bx{3}{3}{0.70} & \bx{2}{3}{0.91} & \bx{2}{3}{0.82} & \impass{\bx{2}{2}{0.52}} & \impass{\bx{2}{3}{0.40}} & 4/5 & 2/11 & 1.32 \gnote{(extrap.)} \\
\mname{h3} & \bx{3}{3}{0.82} & \bx{3}{3}{0.62} & \bx{3}{3}{0.58} & \bx{3}{3}{0.56} & \impass{\bx{3}{3}{0.56}} & \impass{\bx{3}{3}{0.68}} & 6/6 & 0/12 & none \\
\mname{wan22} & \bx{1}{3}{0.83} & \bx{1}{3}{0.78} & \bx{1}{3}{0.87} & \bx{1}{3}{0.70} & \impass{\bx{1}{3}{0.96}} & \impass{\bx{1}{3}{1.07}} & 2/6 & 5/9 & by seed \\
\grouprule
\mname{wan30} & \notrun[9mm] & \bx{3}{3}{0.27} & \notrun[9mm] & \notrun[9mm] & \impass{\bx{2}{2}{0.33}} & \impass{\bx{2}{2}{0.33}} & 4/4 & 0/3 & none \\
\mname{sd20} & \notrun[9mm] & \bx{2}{2}{0.46} & \notrun[9mm] & \notrun[9mm] & \impass{\bx{3}{3}{0.53}} & \impass{\bx{3}{3}{0.45}} & 6/6 & 0/2 & none \\
\mname{sd20mini} & \notrun[9mm] & \bx{3}{3}{0.42} & \notrun[9mm] & \notrun[9mm] & \impass{\bx{3}{3}{0.46}} & \impass{\bx{3}{3}{0.43}} & 6/6 & 0/3 & none \\
\bottomrule
\end{stab}
\end{table*}

\paragraph{Chutes.}
The four chutes release the ball in the same state, and the stated law lands it at the same point whatever the drawn end curvature.
Only Wan~3.0 lands where the drawn curvature points (Table~\ref{tab:supp-chute-results}): its landing moves by $-2.95$\,mm per unit of $\kappa$ (permutation $p=0.009$), about 35\,mm from $\kappa=-4$ to $+8$, and its exit direction steepens with $\kappa$ ($-2.36\degree$ per unit, $p=0.014$).
The ball leaves the tip carrying the direction the chute had before its end.
MiniMax-H3 also steepens its exit with $\kappa$ ($p=0.02$), but its landing order stays within the spread between chutes ($p=0.10$).
The other models show no ordering by $\kappa$ ($p\ge0.13$), and LTX-2.3, LTX-2.5 (distilled), Wan~2.2 (A14B) and Cosmos 3 Nano have 4--6 readable videos.
The Seedance models let the ball pause at the tip and then drop, so their fitted exit direction is not reported.

\begin{table}[t]
\centering
\scriptsize
\caption{Chutes. Landing slope (mm per unit $\kappa$ in m$^{-1}$) with bootstrap interval and permutation $p$, mean landing offset in mm from the physical landing (720\,mm), and the exit-direction slope ($\degree$ per unit $\kappa$, with $p$). $n$: readable videos. Tinted: the model whose landing follows the drawn curvature.}
\label{tab:supp-chute-results}
\setlength{\tabcolsep}{2.4pt}
\begin{stab}{hState}{lccccc}
\toprule
\shead model & $n$ & landing slope & $p$ & offset & exit slope \hu{($p$)} \\
\bandrule
\oraclerow \oraclelabel & & 0 & & 0 & 0 \\
\oraclerule
\mshort{ltx23} & \gnote{4} & \vci{-0.99}{-1.73}{-0.25} & 0.50 & \snum{+3.9} & \snum{+1.40} \gnote{(0.51)} \\
\mshort{ltx25d} & \gnote{6} & \vci{+1.91}{0.87}{2.95} & 0.13 & \snum{-59.7} & \snum{+0.07} \gnote{(1.0)} \\
\mshort{cosmos} & \gnote{6} & \vci{+2.45}{-1.16}{6.07} & 0.50 & \snum{-40.2} & \snum{-0.63} \gnote{(0.83)} \\
\mshort{h3} & \gnote{12} & \vci{-1.13}{-2.51}{-0.19} & 0.10 & \snum{-59.3} & \snum{-0.79} \gnote{(0.020)} \\
\mshort{wan22} & \gnote{5} & \vci{+4.02}{1.38}{6.66} & 0.58 & \snum{-3.9} & \snum{+1.54} \gnote{(0.08)} \\
\grouprule
\mshort{wan30} & \gnote{8} & \tilec[qStateL]{\vci{-2.95}{-4.18}{-1.67}} & \tile[qStateL]{0.009} & \snum{-14.6} & \snum{-2.36} \gnote{(0.014)} \\
\mshort{sd20} & \gnote{8} & \vci{+1.80}{-2.89}{6.00} & 0.59 & \snum{-2.8} & \nd \\
\mshort{sd20mini} & \gnote{9} & \vci{+0.13}{-3.38}{3.37} & 0.95 & \snum{-3.6} & \nd \\
\bottomrule
\end{stab}
\end{table}

\paragraph{The fitted mechanisms beyond the track family.}
Each model's two fitted mechanisms from Sec.~\ref{sec:supp-res-response} were run on the local-edit, shared-front, ramp and barrier geometries at the model's tempo and read by the same estimator (Table~\ref{tab:supp-mech-carry}, exploratory).
On the local edits both families predict a local response in the physical direction and an unchanged motion before the edit.
Only MiniMax-H3 has that form, at a size near the slope-target prediction, and the pre-edit changes of LTX-2.5 and Wan~2.2 (A14B) come from neither family.
On ramps and barriers the two families separate through gravity.
The slope-target law has a target speed with a positive floor, so its ball slows on an ascent without stopping and crosses every barrier.
The smoothed cruise keeps gravity, so at the fitted rates its ball turns back low on the 60\degree{} ascent and crosses only the lowest barriers, and it still does so when the preferred speed is raised by half.
All five models placed in the smoothed-cruise family on the track family behave like the slope-target law on ramps and barriers, and LTX-2.3 and LTX-2.5, placed with the slope target, agree with it there and depart from both families at the local edits.
No single preferred-speed rule thus reproduces one model's behaviour on the undulating track, at the local edit and on the ramp pair together.
Three observations lie outside both families: the ball leaving the track at the top, the direction carried out of a curved chute, and outcomes set by the seed.
The mechanism fits are the most economical description of the track response, and locality and energy call for something beyond them.

\newcommand{\otc}[3]{#1 \gnote{/ #2 / #3}}
\begin{table*}[t]
\centering
\scriptsize
\caption{Observed readings against the two fitted mechanisms: the observed value in dark, then the predictions of the slope target and of the smoothed cruise in grey. Bump and dip: $dv$ in percent. $|C|$: fixed part on shared-front tracks. Ramps: mean height ratio and the outcome on the 60\degree{} ascent (top: reaches the top stop, turn: turns back at the given ratio). Barriers: crossings (the predictions count the six heights). In brackets the family of the track-family fit.}
\label{tab:supp-mech-carry}
\setlength{\tabcolsep}{4pt}
\begin{stab}{hState}{lllllll}
\toprule
\shead & \multicolumn{6}{c}{observed \hu{/ slope target / smoothed cruise}} \\
\shead model \hu{(family)} & bump $dv$ & dip $dv$ & fixed $|C|$ & ramps: mean & 60\degree{} ascent & barriers crossed \\
\bandrule
\mname{ltx23} \gnote{(target)} & \otc{\snum{+4.1}}{\snum{-3.5}}{\snum{-2.7}} & \otc{\snum{-1.1}}{\snum{+2.3}}{\snum{+1.8}} & \otc{0.16}{0.05}{0.34} & \otc{1.23}{1.81}{1.18} & \otc{\nd}{top}{turn 0.36} & \otc{18/18}{6}{6} \\
\mname{ltx25d} \gnote{(cruise)} & \otc{\snum{-3.9}}{\snum{-5.5}}{\snum{-8.0}} & \otc{\snum{+1.9}}{\snum{+3.0}}{\snum{+3.3}} & \otc{0.14}{0.02}{0.35} & \otc{2.36}{1.81}{1.02} & \otc{top}{top}{turn 0.53} & \otc{18/18}{6}{1} \\
\mname{ltx25} \gnote{(target)} & \otc{\snum{-2.1}}{\snum{-3.4}}{\snum{-6.9}} & \otc{\snum{-6.1}}{\snum{+1.8}}{\snum{+1.9}} & \otc{0.32}{0.02}{0.28} & \otc{1.60}{1.81}{1.01} & \otc{top, turn}{top}{turn 0.46} & \otc{11/12}{6}{4} \\
\mname{cosmos} \gnote{(target)} & \otc{\snum{-2.6}}{\snum{-2.6}}{\snum{-4.2}} & \otc{\snum{-0.1}}{\snum{+1.9}}{\snum{+0.3}} & \otc{0.48}{0.01}{0.52} & \otc{1.15}{1.81}{1.78} & \otc{top, turn}{top}{top} & \otc{13/16}{6}{6} \\
\mname{h3} \gnote{(cruise)} & \otc{\snum{-2.4}}{\snum{-4.4}}{\snum{-9.3}} & \otc{\snum{+1.0}}{\snum{+2.9}}{\snum{+5.6}} & \otc{0.55}{0.06}{0.40} & \otc{1.56}{1.68}{0.71} & \otc{top}{top}{turn 0.60} & \otc{18/18}{6}{1} \\
\mname{wan22} \gnote{(target)} & \otc{\snum{-29.3}}{\nd}{\nd} & \otc{\snum{-3.9}}{\nd}{\nd} & \otc{0.05}{0.10}{0.16} & \otc{1.76}{1.66}{0.80} & \otc{top}{top}{turn 0.53} & \otc{by seed}{6}{1} \\
\grouprule
\mname{wan30} \gnote{(cruise)} & \otc{\snum{-7.0}}{\snum{-5.3}}{\snum{-7.3}} & \otc{\snum{-2.7}}{\snum{+3.6}}{\snum{+3.6}} & \otc{0.19}{0.01}{0.11} & \otc{1.94}{1.76}{1.00} & \otc{top}{top}{turn 0.61} & \otc{7/7}{6}{3} \\
\mname{sd20} \gnote{(cruise)} & \otc{\snum{-5.0}}{\snum{-7.2}}{\snum{-8.3}} & \otc{\snum{+9.3}}{\snum{+4.9}}{\snum{+4.9}} & \otc{0.30}{0.04}{0.48} & \otc{1.66}{1.74}{0.72} & \otc{top}{top}{turn 0.52} & \otc{8/8}{6}{2} \\
\mname{sd20mini} \gnote{(cruise)} & \otc{\snum{-1.5}}{\snum{-3.5}}{\snum{-8.5}} & \otc{\snum{+12.3}}{\snum{+2.2}}{\snum{+4.4}} & \otc{0.17}{0.04}{0.47} & \otc{1.82}{1.74}{0.73} & \otc{top}{top}{turn 0.51} & \otc{9/9}{6}{2} \\
\bottomrule
\end{stab}
\end{table*}

\subsection{Outcome: prompt pairs and tray interventions}
\label{sec:supp-res-outcome}

The named outcome strongly controls the destination: with the same first frame and seed and a prompt that leaves the outcome unnamed, tray arrival falls in all eight models tested, the steering beyond the ball's own exit state shrinks to at most 27 points, and a drawn tray by itself barely attracts the ball.

\paragraph{Prompt pairs.}
This paragraph and Table~\ref{tab:supp-promptpairs} use the pairs that are trackable under both prompts, while Table~\ref{tab:supp-destination} covers all named-prompt videos.
Arrival falls by 33 to 96 percentage points.
The unpaired arrival over all trackable videos of each prompt shows the same drop, and so do the pairs in which the ball is visible at the end of both clips (there Wan~3.0 shrinks to $-14$ points, $[-30,0]$).
The neutral prompt also slows the motion of seven models (tempo 1.22--2.35 times that of the named prompt, with LTX-2.3 unchanged on its three pairs), and part of the drop is motion that is still under way at the end of the clip.
On the pairs whose neutral motion is complete (the ball has arrived, is at rest, or would have arrived at its own launch tempo), the drop stays large for Cosmos 3 Nano, LTX-2.3, Wan~2.2 (A14B) and Seedance 2.0 mini and becomes small for LTX-2.5 (distilled) and Wan~3.0.
MiniMax-H3 shows that the named outcome also controls when the motion ends.
At its own launch tempo its ball could not reach the tray within 5\,s in 43 of 46 named-prompt videos, and 42 of them end in the tray.
With the neutral prompt 40 of 45 are still moving at the end.
The steering beyond the ball's exit state, $+43$ to $+75$ points for MiniMax-H3 and the Seedance models with the named prompt, falls to $-18$, 0 and $+11$ points, and the negative value comes from motions still under way at the end of the clip.
Under the neutral prompt no model steers by more than $+27$ points (LTX-2.5 (distilled), $[-3,+52]$).
Wan~3.0 keeps the most of its destination: it reaches the tray in 67\% of its neutral videos and steers by $+22$ points ($[0,+43]$) without a named outcome.
The route stays wrong under both prompts: the exact order appears in one named-prompt video each of the two Seedance models and in none of the others.

\newcommand{\arrcell}[3][qGoalD]{\gpad{\makecell[c]{\ivband{81}{98}\dumbbell[#1]{#3}{#2}{}\\[-1.6pt]#2\% / #3\%}}}
\newcommand{\misscell}[3]{\gpad{\makecell[c]{\tribar{#1}{#2}{#3}{qTimeM}{qIntM}{qIntL}\\[-1.6pt]#1 / #2 / #3}}}
\begin{table*}[t]
\centering
\perarmhead\\[-0.5pt]
\perarm{480}{67.3}
\vspace{3pt}
\caption{Effect of naming the destination. Top: tray arrival under named and neutral prompts. Bottom: matched prompt pairs, reporting arrival change, steering relative to the physical rollout from the model's exit state, tempo change and the status of neutral-prompt misses. Intervals are 95\% scene-bootstrap intervals.}
\label{tab:supp-promptpairs}
\scriptsize
\setlength{\tabcolsep}{3.6pt}
\begin{stab}{hGoal}{lcccccccc}
\toprule
\shead & & tray arrival & & & \multicolumn{2}{c}{steering} & & \\
\shead model & pairs & neutral / named & difference & completed & neutral & named & tempo & neutral misses \\
\bandrule
\oraclerow \oraclelabel & & 81--98\% \gnote{of members} & & & & & & \\
\oraclerule
\mname{ltx23} & 8 & \arrcell[mIcecyanD]{25}{100} & \vci{-75}{-100}{-50} & \vcin{-67}{-100}{-33}{6} & \vcin{+25}{0}{+50}{8} & \vcin{+33}{0}{+50}{3} & 1.01\,\nn{3} & \misscell{1}{5}{0} \\
\mname{ltx25d} & 27 & \arrcell[mGlacierD]{56}{100} & \vci{-44}{-63}{-28} & \vcin{-17}{-36}{0}{18} & \vcin{+27}{-3}{+52}{26} & \vcin{+44}{+20}{+69}{16} & 2.35\,\nn{15} & \misscell{7}{5}{0} \\
\mname{cosmos} & 15 & \arrcell[mSlateD]{7}{100} & \vci{-93}{-100}{-80} & \vcin{-92}{-100}{-75}{12} & \vcin{+0}{-50}{+50}{6} & \vcin{+50}{+17}{+83}{6} & 1.38\,\nn{5} & \misscell{1}{9}{4} \\
\mname{h3} & 46 & \arrcell[mLavD]{2}{98} & \vci{-96}{-100}{-89} & \nd\,\nn{2} & \vcin{-18}{-30}{-7}{45} & \vcin{+43}{+31}{+56}{46} & 1.22\,\nn{45} & \misscell{40}{5}{0} \\
\mname{wan22} & 34 & \arrcell[mOrchidD]{3}{47} & \vci{-44}{-58}{-28} & \vcin{-42}{-75}{-8}{12} & \vcin{-20}{-40}{0}{15} & \vcin{-14}{-39}{+17}{22} & 1.41\,\nn{12} & \misscell{20}{8}{5} \\
\grouprule
\mname{wan30} & 9 & \arrcell[mRoseD]{67}{100} & \vci{-33}{-44}{-17} & \vcin{-14}{-38}{0}{7} & \vcin{+22}{0}{+43}{9} & \vcin{+0}{0}{0}{9} & 1.25\,\nn{9} & \misscell{1}{0}{2} \\
\mname{sd20} & 9 & \arrcell[mCoralD]{33}{100} & \vci{-67}{-89}{-33} & \nd\,\nn{3} & \vcin{+0}{-27}{+27}{9} & \vcin{+62}{+40}{+100}{8} & 1.26\,\nn{9} & \misscell{5}{0}{1} \\
\mname{sd20mini} & 9 & \arrcell[mPeachD]{33}{100} & \vci{-67}{-100}{-25} & \vcin{-40}{-100}{0}{5} & \vcin{+11}{-45}{+75}{9} & \vcin{+75}{+50}{+100}{8} & 1.38\,\nn{9} & \misscell{1}{2}{3} \\
\bottomrule
\end{stab}\\[3pt]
{\scriptsize\color{sInkB} Arrival: named (filled) to neutral (open) on 0 to 100\%, over the band of physical arrival. Completed: pairs whose neutral motion is complete. Neutral misses: \okey{qTimeM}{still moving} / \okey{qIntM}{at rest} / \okeyf{qIntL}{out of view or not visible} at the end. Brackets: pairs or videos.}
\end{table*}

\paragraph{Shifted tray.}
With the neutral prompt only the drawn tray can attract the ball.
The capture contrast is 0.42 for Seedance 2.0 and 0.10 for Seedance 2.0 mini with intervals that exclude zero, far below 1, and its interval includes zero for Wan~3.0 and MiniMax-H3 (Table~\ref{tab:supp-shiftedtray}).
The high arrival in the beyond group comes from the models' own short landings: the ball leaves the deflector slowly and lands 125--154\,mm on the incoming side of the physical point already in the in-tray group, and the beyond trays sit exactly there.
The landing then moves a further 26--74\,mm toward the tray.
MiniMax-H3 drops the ball almost vertically below the deflector in both groups.
Cosmos 3 Nano, LTX-2.3, LTX-2.5 (distilled) and Wan~2.2 (A14B) leave one or two readable pairs (a second ball, struck deflectors, or balls that have not landed by the end).

\newcommand{\capcell}[4][sInk]{\gpad{\makecell[c]{\ivbar[#1]{#3}{#2}{#4}{0}{-0.3}{1}\\[-0.4pt]\snum{#2}\,\ci{#3}{#4}}}}
\begin{table}[t]
\centering
\scriptsize
\caption{Shifted tray (neutral prompt). Capture contrast over pairs of the same base scene and seed (0: ignores the tray, 1: follows its centre), drawn on $-0.3$ to 1 with the physical 0 as a tick, tray arrival in the in-tray and beyond groups (physical oracle 100\% and 0\%), and the median landing offset from the physical point in the two groups (mm, positive toward the deflector).}
\label{tab:supp-shiftedtray}
\setlength{\tabcolsep}{3pt}
\begin{stab}{hGoal}{lccccc}
\toprule
\shead model & pairs & capture contrast & in & beyond & offset \hu{(mm)} \\
\bandrule
\oraclerow \oraclelabel & & 0 & 100\% & 0\% & 0/0 \\
\oraclerule
\mname{h3} & 12 & \capcell[mLavD]{-0.11}{-0.26}{0.04} & 0\% & 17\% & \snum{+317}/\snum{+305} \\
\grouprule
\mname{wan30} & 6 & \capcell[mRoseD]{0.41}{-0.16}{0.45} & 83\% & 67\% & \snum{+146}/\snum{+215} \\
\mname{sd20} & 6 & \capcell[mCoralD]{0.42}{0.21}{0.54} & 83\% & 83\% & \snum{+125}/\snum{+199} \\
\mname{sd20mini} & 6 & \capcell[mPeachD]{0.10}{0.06}{0.41} & 50\% & 100\% & \snum{+154}/\snum{+180} \\
\bottomrule
\end{stab}
\end{table}

\paragraph{Two trays and no tray.}
When a second, identical tray sits under the launch lip, the stated law rolled from the models' own slow and steep exit states would mostly send the ball into it, and the models put it there in 0--3 videos (Table~\ref{tab:supp-twotrays}).
Their balls end elsewhere: MiniMax-H3 is still moving at the end of all nine such videos, Wan~2.2 (A14B) rests on the first deflector or on the board, and Cosmos 3 Nano often leaves the ball on the launch ramp.
Wan~3.0 puts four of six balls in the physical tray, while its own exit state would send five of six into the second one: its destination agrees with the scene's physical outcome, consistent with its steering under the neutral prompt.
With the tray removed, no model stops the ball where the tray stood, and only one Cosmos 3 Nano video draws the tray back.

\newcommand{\finalcell}[3]{\gpad{\makecell[c]{\tribar{#1}{#2}{#3}{qGoalD}{qGoalM}{qIntL}\\[-1.6pt]#1 / #2 / #3}}}
\begin{table}[t]
\centering
\scriptsize
\caption{Two trays (neutral prompt). Final position of trackable videos, \okey{qGoalD}{physical tray} / \okey{qGoalM}{second tray} / \okeyf{qIntL}{elsewhere}, and the share of rollouts of the stated law from the model's own exit state that end in the second tray, unscaled and scaled by the launch tempo. Physical oracle: at least 63/64 in the physical tray, 0/64 touch the second.}
\label{tab:supp-twotrays}
\setlength{\tabcolsep}{4pt}
\begin{stab}{hGoal}{lccc}
\toprule
\shead model & $n$ & \makecell[c]{final position\\[-1pt]\hu{physical / second / elsewhere}} & rollout to second \\
\bandrule
\oraclerow \oraclelabel & 64 & ${\ge}63$ / 0 / \nd & \\
\oraclerule
\mname{ltx23} & 1 & \finalcell{0}{1}{0} & 1/1 \\
\mname{ltx25d} & 10 & \finalcell{3}{3}{4} & 10/10; 8/10 \\
\mname{cosmos} & 6 & \finalcell{1}{0}{5} & \nd \\
\mname{h3} & 12 & \finalcell{0}{3}{9} & 10/12; 6/11 \\
\mname{wan22} & 12 & \finalcell{0}{3}{9} & 7/9; 8/8 \\
\grouprule
\mname{wan30} & 6 & \finalcell{4}{0}{2} & 4/6; 5/6 \\
\mname{sd20} & 6 & \finalcell{1}{1}{4} & 5/5; 4/6 \\
\mname{sd20mini} & 6 & \finalcell{3}{2}{1} & 5/6; 1/6 \\
\bottomrule
\end{stab}
\end{table}

\paragraph{Near misses and permuted layouts.}
A deflector 10\,mm from the path is touched by every closed model in at least one of three seeds: Wan~3.0 in 3 of 3, Seedance 2.0 in 2 and Seedance 2.0 mini in 1, each a real rebound off the deflector end visible in all frames around the closest point.
Wan~2.2 (A14B) touches the near-miss deflector in 3 of 7 videos (one per placement) and MiniMax-H3 in 1 of 9, whose path passes 70--134\,mm below the deflectors placed above it.
With three seeds these counts establish that the models' paths come within 10\,mm of the deflector end.
On permuted layouts no trackable video reproduces either order (0 of 61), so the permutation reads at the floor.
The first deflector alone is hit in 0/11 to 8/17 videos per model.

\subsection{Timing: clip length}
\label{sec:supp-res-timing}

With the same scene, first frame, prompt and seed, the clip length controls the pace of the motion: from 3 to 8\,s the same event stretches nearly in proportion, events fall at fixed fractions of the clip, and the extra time is spent at rest or on a repeat of the event.
Both prompts of this experiment name the outcome.

\paragraph{Tempo and event times.}
On the tracks the duration exponent is 0.85--1.04 in three models (Table~\ref{tab:supp-length}), where a fixed physical rate gives 0 and a motion fitted to the clip gives 1.
The intervals of MiniMax-H3 (1.04, $[0.95,1.14]$) and LTX-2.3 (0.85, $[0.53,1.04]$) include 1, and that of LTX-2.5 (distilled) (0.91, $[0.90,0.92]$) lies just below it.
MiniMax-H3 reaches the end stop at 90--93\% of the clip at every length, and its arrival times scale by 1.61 and 0.59 for clip ratios of 1.6 and 0.6.
LTX-2.3 arrives at 55--59\% of the clip and LTX-2.5 (distilled) at 47--52\%, and both then rest.
Wan~2.2 (A14B) runs twice as fast in 3\,s as in 5\,s (tempo ratio 0.49, $[0.38,0.53]$, 10 units) and keeps its 5\,s pace at 8\,s (0.85, $[0.70,1.12]$, 7 units).
A single exponent would average these two opposite steps, so the two ratios describe its tracks, and in 5\,s only one of its twelve track videos reaches the stop.
On deflectors the first tray contact gives $\gamma$ of 1.06 for MiniMax-H3, 0.76 for Wan~2.2 (A14B) and 0.44 for LTX-2.5 (distilled).
The tempo differences between lengths are 11--53 times the tempo floor.

\newcommand{\tempocell}[2][sInk]{\gpad{\makecell[c]{\lenspark[#1]{#2}{0}{4.5}\\[-1.6pt]\supplistslash{#2}}}}
\newcommand{\supplistslash}[1]{\foreach \x [count=\i] in {#1}{\ifnum\i>1\,/\,\fi\x}}
\newcommand{\gamcell}[5][sInk]{\gpad{\makecell[c]{\ivbar[#1]{#3}{#2}{#4}{0}{-0.1}{1.2}\\[-1.6pt]\snum{#2}\,\ci{#3}{#4}\,\nn{#5}}}}
\begin{table*}[t]
\centering
\perarmhead\\[-0.5pt]
\perarm{544}{0}
\vspace{3pt}
\caption{Clip length. Top: row 9 of Fig.~\ref{fig:supp-perarm}, the arrival at the end stop as a fraction of the clip by clip length. Bottom: tempo at 3 / 5 / 8\,s (drawn on 0 to 4.5), duration exponent $\gamma$ with interval and scene-seed units (drawn on $-0.1$ to 1.2 with the physical 0 as a tick), the arrival at the end stop as a share of the clip over readable videos (videos), and the paired ratio of arrival times 8\,s over 5\,s and 3\,s over 5\,s (physical oracle 1, a motion fitted to the clip 1.6 and 0.6). Wan~2.2 (A14B): the tempo ratios 3\,s over 5\,s and 8\,s over 5\,s in place of $\gamma$.}
\label{tab:supp-length}
\scriptsize
\setlength{\tabcolsep}{4.6pt}
\begin{stab}{hTime}{lccccccc}
\toprule
\shead & & & \multicolumn{3}{c}{arrival at the end stop \hu{(share of clip)}} & \multicolumn{2}{c}{arrival-time ratio} \\
\shead model & $\alpha$ at 3 / 5 / 8\,s & $\gamma$, tracks & 3\,s & 5\,s & 8\,s & 8/5 & 3/5 \\
\bandrule
\oraclerow \oraclelabel & equal at all lengths & 0 & & & & 1 & 1 \\
\oraclerule
\mname{ltx23} & \tempocell[mIcecyanD]{1.05,1.53,2.64} & \gamcell[mIcecyanD]{0.85}{0.53}{1.04}{11} & 59\%\,\nn{4} & 55\%\,\nn{8} & 56\%\,\nn{3} & 1.92 & \vci{0.44}{0.38}{1.50} \\
\mname{ltx25d} & \tempocell[mGlacierD]{0.88,1.42,2.36} & \gamcell[mGlacierD]{0.91}{0.90}{0.92}{12} & 52\%\,\nn{9} & 47\%\,\nn{9} & 52\%\,\nn{11} & \vci{1.69}{1.31}{1.95} & \vci{0.75}{0.50}{0.85} \\
\mname{h3} & \tempocell[mLavD]{1.48,2.37,4.09} & \gamcell[mLavD]{1.04}{0.95}{1.14}{12} & 91\%\,\nn{12} & 90\%\,\nn{10} & 93\%\,\nn{10} & \vci{1.61}{1.53}{1.65} & \vci{0.59}{0.57}{0.60} \\
\mname{wan22} & \tempocell[mOrchidD]{1.29,2.76,2.64} & tempo 0.49; 0.85 & 80\%\,\nn{7} & 65\%\,\nn{1} & 58\%\,\nn{5} & \nd & 0.91 \\
\bottomrule
\end{stab}
\end{table*}

\paragraph{Track fingerprints across lengths.}
The undulation response stretches with the motion (Table~\ref{tab:supp-length-fingerprint}).
Apart from the two readable 8\,s videos of LTX-2.3 and the phase of Wan~2.2 (A14B), the changes of gain and phase between lengths stay within twice their floors and those of accuracy within three times, so the pattern of the track family holds at every length.
The relaxation rate, fitted in video time, changes most (13--49 times its floor).
Converted to physical time (exploratory), it is the same at 5 and 8\,s for both LTX models.
At 3\,s the converted rate and the trend ratio depart from the longer clips, which may come from the truncation of short clips, for which no floor exists.

\newcommand{\fpcell}[6][sInk]{\gpad{\lenspark[#1]{#2}{#3}{#4}}\hspace{0.5em}#5\if\relax\detokenize{#6}\relax\else\ \gnote{(#6)}\fi}
\begin{table*}[t]
\centering
\scriptsize
\caption{Track readings at 3 / 5 / 8\,s: gain, phase (\degree), relaxation rate $b$ (s$^{-1}$, video time) with the largest length difference over its floor in brackets, and $b\alpha$ (physical time, exploratory). Each line runs over the three lengths on a common scale per column ($|H|$ 0 to 2, phase $-20$ to $+50\degree$, $b$ 0 to 10, $b\alpha$ 0 to 40).}
\label{tab:supp-length-fingerprint}
\setlength{\tabcolsep}{7pt}
\begin{stab}{hTime}{lllll}
\toprule
\shead model & $|H|$ & phase & $b$ & $b\alpha$ \\
\bandrule
\mname{ltx23} & \fpcell[mIcecyanD]{0.71,0.90,1.82}{0}{2}{0.71/0.90/1.82}{8.8} & \fpcell[mIcecyanD]{40,39,14}{-20}{50}{\snum{+40}/\snum{+39}/\snum{+14}}{1.0} & \fpcell[mIcecyanD]{2.9,3.9,2.9}{0}{10}{2.9/3.9/2.9}{13} & \fpcell[mIcecyanD]{2.6,8.8,9.0}{0}{40}{2.6/8.8/9.0}{} \\
\mname{ltx25d} & \fpcell[mGlacierD]{0.45,0.70,0.52}{0}{2}{0.45/0.70/0.52}{1.7} & \fpcell[mGlacierD]{40,34,49}{-20}{50}{\snum{+40}/\snum{+34}/\snum{+49}}{1.4} & \fpcell[mGlacierD]{3.5,4.1,3.5}{0}{10}{3.5/4.1/3.5}{33} & \fpcell[mGlacierD]{3.0,8.2,8.6}{0}{40}{3.0/8.2/8.6}{} \\
\mname{h3} & \fpcell[mLavD]{0.35,0.47,0.45}{0}{2}{0.35/0.47/0.45}{1.2} & \fpcell[mLavD]{7,-5,-1}{-20}{50}{\snum{+7}/\snum{-5}/\snum{-1}}{1.8} & \fpcell[mLavD]{5.1,1.5,1.4}{0}{10}{5.1/1.5/1.4}{49} & \fpcell[mLavD]{8.3,4.0,5.7}{0}{40}{8.3/4.0/5.7}{} \\
\mname{wan22} & \fpcell[mOrchidD]{0.71,0.46,0.68}{0}{2}{0.71/0.46/0.68}{0.9} & \fpcell[mOrchidD]{-16,17,-7}{-20}{50}{\snum{-16}/\snum{+17}/\snum{-7}}{3.5} & \fpcell[mOrchidD]{5.4,3.4,9.9}{0}{10}{5.4/3.4/9.9}{49} & \fpcell[mOrchidD]{5.1,8.2,37.8}{0}{40}{5.1/8.2/37.8}{} \\\bottomrule
\end{stab}
\end{table*}

\paragraph{Deflectors across lengths and the time after the event.}
MiniMax-H3 and LTX-2.5 (distilled) already bring every settled ball into the tray at 3\,s, and longer clips bring no more balls to the tray and no better order (exact order at 3, 5 and 8\,s in 5/9, 4/10 and 6/11 videos of MiniMax-H3 and in 2/10, 5/10 and 3/9 of LTX-2.5 (distilled)).
Wan~2.2 (A14B) lands 12/12 in 3\,s and 12/12 in 8\,s, and in 7 of the 8\,s videos the ball then returns to the launch point and the event starts again.
In the extra time of 8\,s clips the LTX models leave the ball at rest for 3.5--3.8\,s on tracks, LTX-2.5 (distilled) lifts the ball back onto a deflector and drops it into the tray a second time in 4 of 9 deflector videos, Wan~2.2 (A14B) loses the ball after the stop in 4 of 7 track videos, and MiniMax-H3 has about 0.5\,s left after the event.

\subsection{Photographed first frames}
\label{sec:supp-res-photo}

Replacing the rendered first frame with a photograph of the built scene leaves the readings of six models within their seed spread, and each of the three changes can be localized to a specific failure mode.

\begin{scard}[hInt]{Decision rule}{photograph against control}
For each model and configuration the photograph row is paired with its control, the rendered twin for open-weight models and the model's own reading on the nominal scene for closed models.
A reading changes with the visual domain only when three conditions hold together:
\fentry{f}{interval}{the interval of the mean paired difference over configurations excludes zero,}
\fentry{m}{seed spread}{the pooled difference exceeds the seed spread $2s\sqrt{1/n_1+1/n_2}$,}
\fentry{l}{floor}{and most pairs exceed the floor of the photograph rows.}
For a closed model a difference is attributed to the photograph only when it leaves the range produced by changing only the control (the twin minus nominal differences of the open-weight models).
The floors are tight (gain 0.04--0.31, phase 1.5--25\degree, residual 0.5--4.6\degree) and the seed spread is usually larger, so the decisions rest mainly on the seed spread.
\end{scard}

\paragraph{Results.}
Cosmos 3 Nano, LTX-2.3 on tracks, LTX-2.5 (distilled), Wan~2.2 (A14B), Seedance 2.0 mini and Wan~3.0 keep their readings (Table~\ref{tab:supp-photo}).
The tempo and gain differences of Wan~3.0 exceed its seed spread but lie within the range that the change of control alone produces (tempo $+0.04$ to $+1.01$, gain $+0.02$ to $+0.38$), and its clip duration stays the same ($+0.05$\,s).
Three changes remain.
LTX-2.5 pushes the camera in on the photographs (fiducial drift in 9 of 12 videos against 0 of 12 twins), so most of its photograph rows are not estimable.
The twin with the same measured geometry is readable in 11 of 12, implicating photographic appearance rather than the measured geometry.
MiniMax-H3 errs by 17.5\degree{} more at contacts on photographs, with the same order and the same tray arrival.
Seedance 2.0 shows a gain higher by 0.66, beyond both its seed spread and the range of the weaker control.
LTX-2.3 treats the deflectors as loose objects in every domain and adds new objects on photographs (Sec.~\ref{sec:supp-failures}), which leaves it almost no trackable deflector videos (0 of 12 photographs, 1 of 9 twins).

The patterns of the main sets hold on photographs for every readable model.
On three-deflector layouts the physical order appears in 0 or 1 of 6 photograph videos per model, and every trackable ball ends in the tray except with Wan~2.2 (A14B) (8 of 12, fewer on twins and nominal scenes).
On tracks the relaxation rate is positive for every model and domain (1--46\,s$^{-1}$), the median phase leads by 8--76\degree{} for the eight models other than Cosmos 3 Nano, whose phase scatters with a seed spread of 59\degree, the median gain is 0.59--1.36, and the trend ratio stays between $-0.28$ and 0.49, far from 1.

\newcommand{\chg}[3]{\tilec[qGoalL]{\vci{#1$^\ast$}{#2}{#3}}}
\begin{table*}[t]
\centering
\setlength{\tabcolsep}{1.2pt}
\begin{tabular}{@{}cccc@{}}
\thumb[0.243\linewidth]{supp/thumbs/photo_track.jpg} & \thumb[0.243\linewidth]{supp/thumbs/twin_track.jpg} &
\thumb[0.243\linewidth]{supp/thumbs/photo_deflectors.jpg} & \thumb[0.243\linewidth]{supp/thumbs/twin_deflectors.jpg}\\
\thumbcap{track, photograph} & \thumbcap{track, rendered twin} & \thumbcap{deflectors, photograph} & \thumbcap{deflectors, rendered twin}\\
\end{tabular}
\vspace{2pt}
\caption{Photographed first frames. Top: a photographed first frame and its rendered twin for a track and for a deflector layout. Bottom: mean difference photograph minus control over configurations, with 95\% intervals below. Tracks: share of estimable videos, tempo, gain, phase (\degree). Deflectors: share of trackable videos, exact order, tray arrival, largest residual (\degree). $^\ast$, tinted: a change under the three-part rule. Control: rendered twin (open-weight models) or the nominal scene (closed models).}
\label{tab:supp-photo}
\scriptsize
\setlength{\tabcolsep}{3.4pt}
\begin{stab}{hInt}{llcccccccc}
\toprule
\shead & & \multicolumn{4}{c}{tracks} & \multicolumn{4}{c}{deflectors} \\
\shead model & control & estimable & $\alpha$ & $|H|$ & phase & trackable & order & tray & residual \\
\bandrule
\mname{ltx23} & \gnote{twin} & \vci{+0.08}{-0.25}{+0.50} & \vci{-0.12}{-0.36}{+0.31} & \vci{+0.90}{+0.78}{+1.28} & \vci{+49}{+6}{+64} & \vci{-0.08}{-0.25}{+0.00} & \nd & \nd & \nd \\
\mname{ltx25d} & \gnote{twin} & \vci{-0.25}{-0.75}{+0.00} & \vci{+0.30}{+0.09}{+0.62} & \vci{-0.27}{-0.31}{-0.24} & \vci{+2}{-18}{+23} & \vci{+0.08}{+0.00}{+0.25} & \vci{-0.33}{-0.75}{+0.00} & \snum{+0.00} & \vci{+18.2}{+16.5}{+19.6} \\
\mname{ltx25} & \gnote{twin} & \chg{-0.42}{-0.67}{-0.17} & \vci{+0.06}{+0.05}{+0.07} & \vci{+0.17}{+0.14}{+0.19} & \vci{+1}{-1}{+4} & \multicolumn{4}{c}{\notrun[30mm]} \\
\mname{cosmos} & \gnote{twin} & \vci{-0.17}{-0.33}{+0.00} & \vci{-0.16}{-0.51}{+0.04} & \vci{-0.25}{-0.74}{+0.08} & \vci{-31}{-101}{+110} & \vci{+0.08}{-0.17}{+0.33} & \vci{+0.08}{+0.00}{+0.25} & \snum{+0.00} & \makecell[c]{\snum{-46}\\[-2.4pt]\gnote{\tiny (1 config.)}} \\
\mname{h3} & \gnote{twin} & \snum{+0.00} & \vci{-0.10}{-0.20}{+0.02} & \vci{-0.01}{-0.09}{+0.13} & \vci{+3}{-31}{+13} & \vci{+0.17}{+0.00}{+0.33} & \snum{+0.00} & \snum{+0.00} & \chg{+17.5}{+2.8}{+32.2} \\
\mname{wan22} & \gnote{twin} & \vci{+0.00}{-0.33}{+0.33} & \vci{-0.04}{-0.18}{+0.14} & \vci{-0.22}{-1.01}{+0.06} & \vci{-16}{-55}{+28} & \vci{+0.08}{+0.00}{+0.25} & \vci{-0.17}{-0.50}{+0.00} & \vci{+0.29}{+0.08}{+0.54} & \vci{+28}{-25}{+81} \\
\grouprule
\mname{wan30} & \gnote{nominal} & \snum{+0.00} & \vci{+0.26}{+0.13}{+0.43} & \vci{+0.35}{+0.22}{+0.40} & \vci{+20}{-7}{+33} & \vci{+0.17}{+0.00}{+0.50} & \vci{-0.17}{-0.50}{+0.00} & \snum{+0.00} & \vci{+10.9}{-3.5}{+32.7} \\
\mname{sd20} & \gnote{nominal} & \snum{+0.00} & \vci{-0.02}{-0.27}{+0.24} & \chg{+0.66}{+0.30}{+0.85} & \vci{+15}{-10}{+24} & \vci{+0.08}{+0.00}{+0.25} & \vci{-0.04}{-0.38}{+0.25} & \snum{+0.00} & \vci{-7.7}{-14.7}{-2.4} \\
\mname{sd20mini} & \gnote{nominal} & \vci{-0.08}{-0.25}{+0.00} & \vci{+0.16}{-0.28}{+0.63} & \vci{+0.02}{-0.14}{+0.41} & \vci{-35}{-60}{+20} & \vci{+0.08}{+0.00}{+0.25} & \vci{-0.17}{-0.33}{+0.00} & \snum{+0.00} & \vci{+8.2}{-12.1}{+31.2} \\
\bottomrule
\end{stab}
\end{table*}

\subsection{Regeneration probe}
\label{sec:supp-res-regeneration}

Given a motion that is drawn correctly and wrong in its dynamics, the two open-weight models keep its timing, add back their own response to the drawn shape, and change the motion where a drawn contact is violated.
What they put in its place moves toward their own version of the scene and barely toward the stated law.

\paragraph{The lowest level copies.}
At the lowest noise level (LTX-2.5 (distilled) 0.7, Wan~2.2 (A14B) 0.90) the output stays within 4--15\% of the distance between source and physics from the source (41\% for the transparent deflector under Wan~2.2 (A14B)), and $\lambda$ is near zero for every version (Table~\ref{tab:supp-lambda}).
Even so the regeneration adds 17--21\,mm of deviation to an exact source, ten times the 1.8\,mm floor, so at this level $\lambda\approx0$ reads as copying and the tendencies are read at the higher levels.

\paragraph{What is kept and what is replaced.}
The versions that break a drawn contact are pulled away from the source: the ball meets the transparent deflector from LTX 0.8 on ($\lambda=1.01$) and from Wan 0.94 on ($\lambda$ 1.09--1.12), and the restitution-0.3 source is pulled to $\lambda=0.32$ at LTX 0.9 and replaced at Wan 0.94 and 0.95 ($\lambda$ 0.99 and 1.40).
The replacement keeps the dynamics wrong.
The order stays wrong in every one of these videos and the first-contact restitution stays at 0.16--0.45 against 0.85 (Table~\ref{tab:supp-regen-readings}).
Split along the two directions (exploratory, seeds 0 and 1, each output against the model's own image-to-video video of the same scene and seed), the Wan~2.2 (A14B) regenerations of the restitution-0.3 source move 0.66 of the way toward that video and 0.10 toward physics (medians of 10 videos at 0.94 and 0.95).
Per level the split is 0.40 and 0.23 at 0.94 (4 videos) and 0.73 and 0.08 at 0.95 (6 videos).
A $\lambda$ above 1 is the model's own version overshooting physics along the source-to-physics direction.
The versions that keep every drawn contact are accepted.
Their $\lambda$ stays below 0.1 except at LTX's highest level (slowed source 0.24, $+90\degree$ source 0.10), where the output drifts toward the model's own version.
The slowed motion keeps its tempo (2.01--2.10 under Wan and 1.81--1.90 under LTX, source 2.0), the constant-speed motion keeps its timing ($\lambda\le0.02$) and regains a response to the drawn undulation (gain 0.27--0.62 against 0.02 in the source), and the $+90\degree$ phase returns toward each model's own lead on the track family (LTX $+30\degree$ at 0.9, Wan $+6$ and $+13\degree$ at 0.94 and 0.95), away from the physical $0\degree$.

\newcommand{\lam}[4]{\heattile{qRespM}{#1}{1.2}{14.5mm}{\vcin{#1}{#2}{#3}{#4}}}
\DeclareRobustCommand{\lamkey}{\tikz[baseline=-0.6ex]{\shade[left color=white,right color=qRespM!46!white] (0,-0.62mm) rectangle (9mm,0.62mm);\draw[qIntM,line width=0.25pt] (0,-0.62mm) rectangle (9mm,0.62mm);}}
\newcommand{\lamgroup}[1]{\multicolumn{7}{@{}l}{\gnote{\itshape #1}}}
\begin{table*}[t]
\centering
\scriptsize
\caption{Regeneration: $\lambda$ per source version and noise level (0: stays on the source, 1: pulled onto physics), median with 95\% interval and readable videos. The tint deepens with $\lambda$, from 0 to 1.2 and above (0\,\lamkey\,1.2).}
\label{tab:supp-lambda}
\setlength{\tabcolsep}{6pt}
\begin{stab}{hResp}{lcccccc}
\toprule
\shead & \multicolumn{3}{c}{\mname{ltx25d}} & \multicolumn{3}{c}{\mname{wan22}} \\
\shead source version & 0.7 & 0.8 & 0.9 & 0.90 & 0.94 & 0.95 \\
\bandrule
\lamgroup{keeps every drawn contact} \\
constant speed & \lam{0.01}{0.00}{0.02}{9} & \lam{0.00}{0.00}{0.01}{9} & \lam{0.00}{-0.06}{0.03}{9} & \lam{0.01}{0.00}{0.02}{9} & \lam{0.02}{0.01}{0.02}{6} & \lam{0.02}{-0.02}{0.47}{6} \\
phase $+90\degree$ & \lam{0.00}{-0.02}{0.01}{9} & \lam{0.02}{0.01}{0.03}{9} & \lam{0.10}{0.04}{0.13}{9} & \lam{0.01}{0.01}{0.02}{9} & \lam{0.07}{0.03}{0.14}{7} & \lam{0.08}{0.04}{0.14}{8} \\
slowed $2\times$ & \lam{-0.00}{-0.01}{0.00}{18} & \lam{0.06}{0.01}{0.19}{18} & \lam{0.24}{0.18}{0.41}{15} & \lam{0.00}{0.00}{0.01}{18} & \lam{0.00}{-0.01}{0.02}{17} & \lam{0.01}{-0.03}{0.02}{16} \\
\grouprule
\lamgroup{breaks a drawn contact} \\
restitution 0.3 & \lam{0.01}{-0.02}{0.02}{9} & \lam{0.03}{-0.05}{0.14}{9} & \lam{0.32}{0.17}{0.92}{7} & \lam{0.01}{0.01}{0.09}{8} & \lam{0.99}{0.11}{1.70}{6} & \lam{1.40}{0.52}{1.76}{9} \\
transparent deflector & \lam{0.07}{0.07}{0.14}{5} & \lam{1.01}{0.99}{1.28}{5} & \lam{1.01}{0.97}{1.11}{4} & \lam{0.28}{0.25}{0.62}{5} & \lam{1.09}{1.01}{1.38}{6} & \lam{1.12}{1.02}{1.35}{8} \\
\bottomrule
\end{stab}
\end{table*}

\begin{table}[t]
\centering
\scriptsize
\caption{The defining reading of each version after regeneration, at the three levels of each model, against the source and the physical oracle.}
\label{tab:supp-regen-readings}
\setlength{\tabcolsep}{1.6pt}
\begin{stab}{hResp}{lc>{\columncolor{sOracle}}ccc}
\toprule
\shead version: reading & source & oracle & \mglyph{ltx25d}\,LTX 0.7/0.8/0.9 & \mglyph{wan22}\,Wan 0.90/0.94/0.95 \\
\bandrule
slowed: $\alpha$ & 2.0 & 1.0 & 1.86/1.90/1.81 & 2.01/2.04/2.10 \\
constant speed: $|H|$ & 0.02 & 1.0 & 0.47/0.60/0.62 & 0.27/0.50/0.52 \\
phase $+90\degree$: phase & \snum{+86} & 0 & \snum{+61}/\snum{+37}/\snum{+30} & \snum{+70}/\snum{+6}/\snum{+13} \\
restitution 0.3: $e_n$ & 0.30 & 0.85 & 0.28/0.28/0.45 & 0.16/0.35/0.29 \\
transparent: exact order & 0 & 1 & 0/0/0 & 0/0/0 \\
\bottomrule
\end{stab}
\end{table}

\paragraph{From the exact source, and the switch of Wan~2.2 (A14B).}
Started from the exact source, both models drift away from the physical motion as the noise rises, by about 105--124\,mm at the highest level (Table~\ref{tab:supp-regen-drift}).
LTX-2.5 (distilled) drifts toward its own image-to-video version ($\kappa$ 0.25--0.29, the same for the same and for another seed, so the direction is the model's general rendering of the scene), and at 0.9 half of its seed-0 and seed-1 outputs across all versions have become that version (21 of 42).
Most drift of Wan~2.2 (A14B) points elsewhere ($\kappa$ 0.04--0.08).
Its response to the noise level is a switch: the deviation stays at 36--40\% of its own image-to-video deviation from 0.92 to 0.94 and jumps to 98\% at 0.95, where 16 of 21 deflector outputs become its own version while the track outputs mostly still follow the source (2 of 23).
The level of the switch depends on the scene, 0.94 for one irregular layout and 0.95 for another.

\newcommand{\dcell}[4][sInk]{\gpad{\makecell[c]{\ivbar[#1]{#3}{#2}{#4}{0}{0}{220}\\[-1.6pt]#2\,\ci{#3}{#4}}}}
\newcommand{\owncell}[5][qIntD]{\gpad{\makecell[c]{\setlength{\shw}{5mm}\fracbar[#1]{#2}{#3}\,\fracbar[#1]{#4}{#5}\\[-1.6pt]#2/#3; #4/#5}}}
\begin{table}[t]
\centering
\scriptsize
\caption{Exact sources: distance from the physical motion (mm, drawn on 0 to 220 with the physical motion at 0) and the component $\kappa$ toward the model's own image-to-video version with the same and with another seed. Last column: outputs of all versions that became the model's own version ($\kappa\ge0.5$), tracks and deflectors, one bar each.}
\label{tab:supp-regen-drift}
\setlength{\tabcolsep}{3pt}
\begin{stab}{hResp}{lcccc}
\toprule
\shead level & $d_{\mathrm{phys}}$ & $\kappa$ same & $\kappa$ other & own version \\
\bandrule
\mglyph{ltx25d}\,LTX 0.7 & \dcell[mGlacierD]{17}{10}{31} & 0.01 & 0.01 & \owncell[mGlacierD]{0}{24}{0}{23} \\
\mglyph{ltx25d}\,LTX 0.8 & \dcell[mGlacierD]{76}{33}{94} & 0.03 & 0.04 & \owncell[mGlacierD]{3}{24}{6}{22} \\
\mglyph{ltx25d}\,LTX 0.9 & \dcell[mGlacierD]{124}{109}{175} & \vci{0.25}{0.21}{0.66} & \vci{0.29}{0.03}{0.32} & \owncell[mGlacierD]{12}{24}{9}{18} \\
\grouprule
\mglyph{wan22}\,Wan 0.90 & \dcell[mOrchidD]{21}{14}{42} & 0.00 & 0.00 & \owncell[mOrchidD]{0}{24}{2}{18} \\
\mglyph{wan22}\,Wan 0.94 & \dcell[mOrchidD]{77}{50}{104} & 0.03 & 0.03 & \owncell[mOrchidD]{1}{21}{9}{19} \\
\mglyph{wan22}\,Wan 0.95 & \dcell[mOrchidD]{105}{41}{211} & \vci{0.04}{0.00}{0.13} & \vci{0.08}{0.02}{0.52} & \owncell[mOrchidD]{2}{23}{16}{21} \\
\bottomrule
\end{stab}
\end{table}

\section{Case grids and per-model charts}
\label{sec:supp-cases}

The grids show the simulator and every model on one scene of each level and of the main interventions, and the charts give every model's readings of every experiment on one page.
Each grid cell is a stroboscopic composite: the ball at equal time steps, from light to full as time passes, on the board of the video, with the simulated path dashed and a small tick at each physical contact.
Cells are chosen by rule.
On a track level the cell shows the model's estimable video of the scene nearest its median reading of the level (the ramp exponent on the straight level, gain and phase at 3 cycles per metre on the others).
On a deflector level it shows a video with the model's usual tray outcome and the first-contact restitution nearest its median, and in the interventions one seed (for the closed models, one generation index) that is trackable in every arm with the model's usual outcome.
A model with no readable video of a scene shows its most typical flagged video, tagged as an altered scene.
Videos in which the camera moves are compensated frame by frame with the measured board homography.
Fig.~\ref{fig:cases} of the main paper shows four of the models on five of these scenes.

\begin{figure*}[p]
\centering
\includegraphics[width=\linewidth,height=7.6in,keepaspectratio]{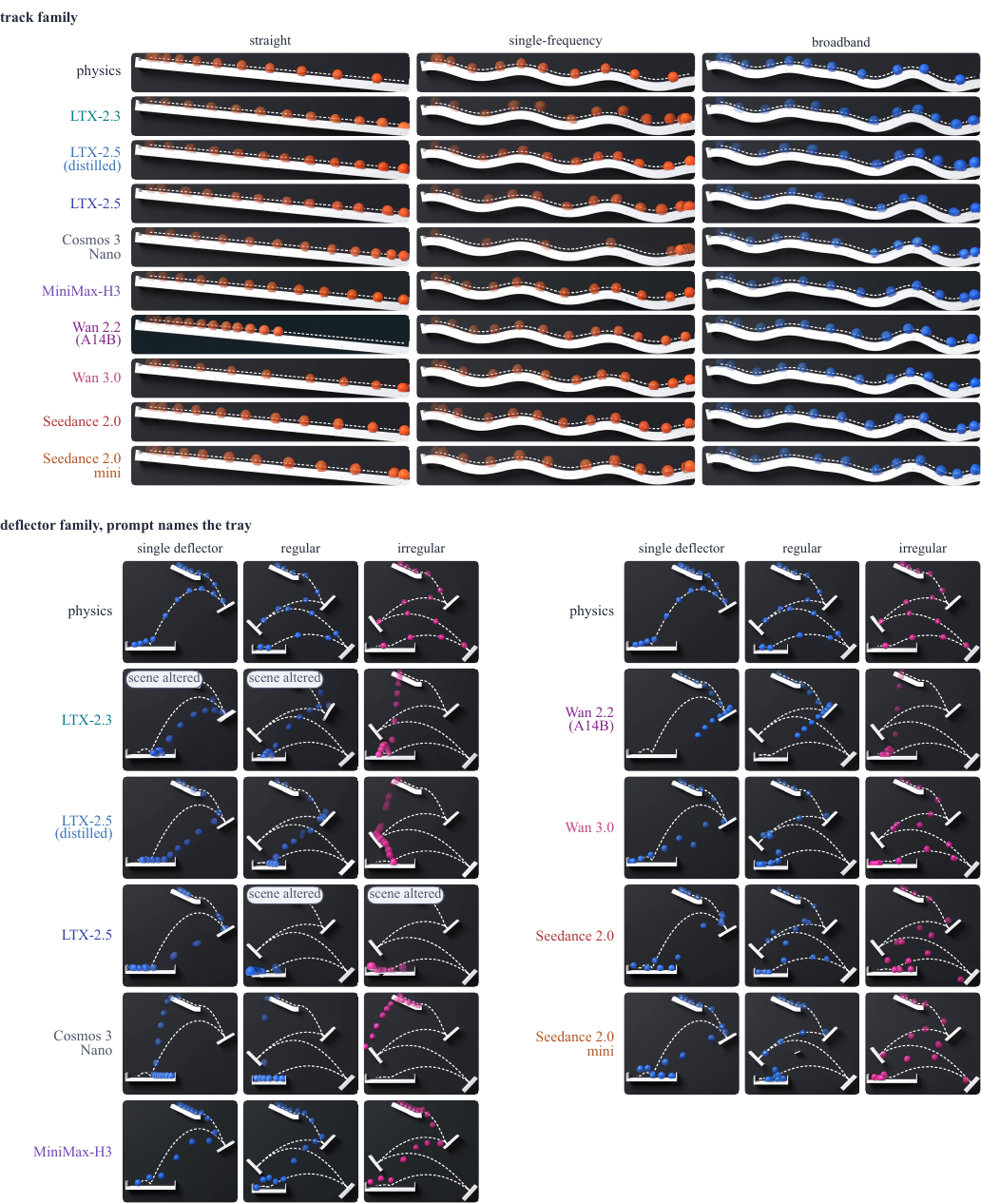}
\caption{\textbf{Main geometry families, all models.} Top: representative straight, single-frequency and broadband track scenes. Bottom: single-, regular- and irregular-deflector scenes under the named-outcome prompt. Dashed curves show physical trajectories; tagged cells mark models with no readable video of that scene.}
\label{fig:supp-cases-families}
\end{figure*}

\begin{figure*}[p]
\centering
\includegraphics[page=1,width=\linewidth,height=7.9in,keepaspectratio]{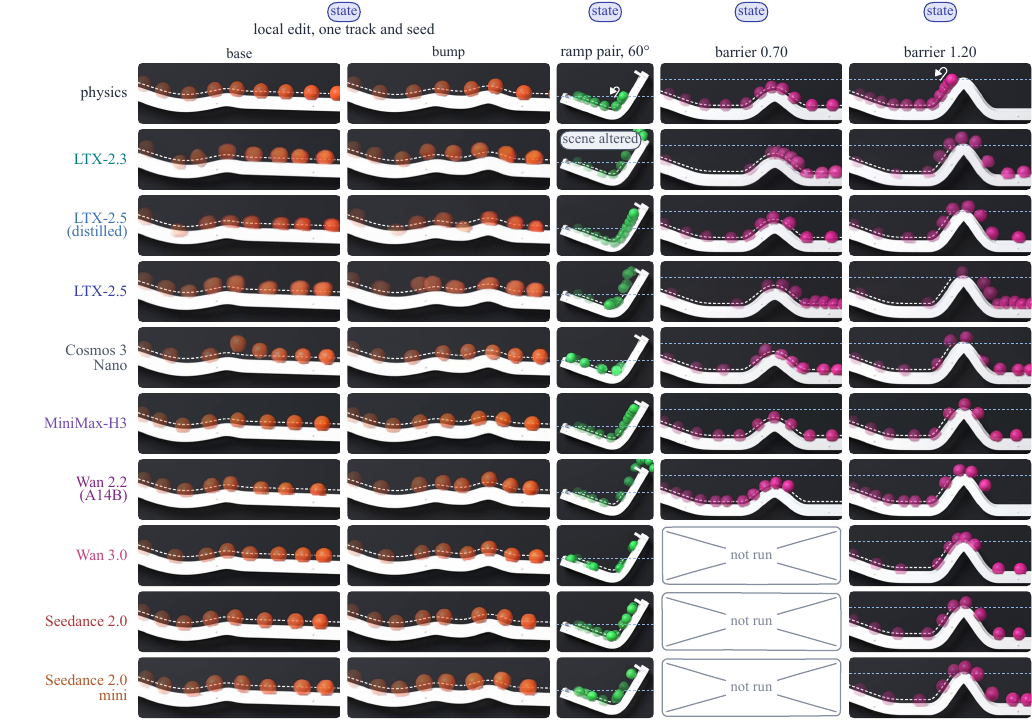}
\caption{\textbf{Interventions on state, all models.} Columns: the local edit (base and bump) on the one track that all nine models ran, the ramp pair with a 60\degree{} ascent, and the barriers at 0.70 and 1.20 of the release drop (the closed models ran 0.85, 1.05 and 1.20). Rows: the simulator and the nine models. The straight dashed line marks the release height and the arrow the physical turning point.}
\label{fig:supp-cases-interventions}
\end{figure*}

\begin{figure*}[p]
\centering
\includegraphics[page=2,width=\linewidth,height=7.6in,keepaspectratio]{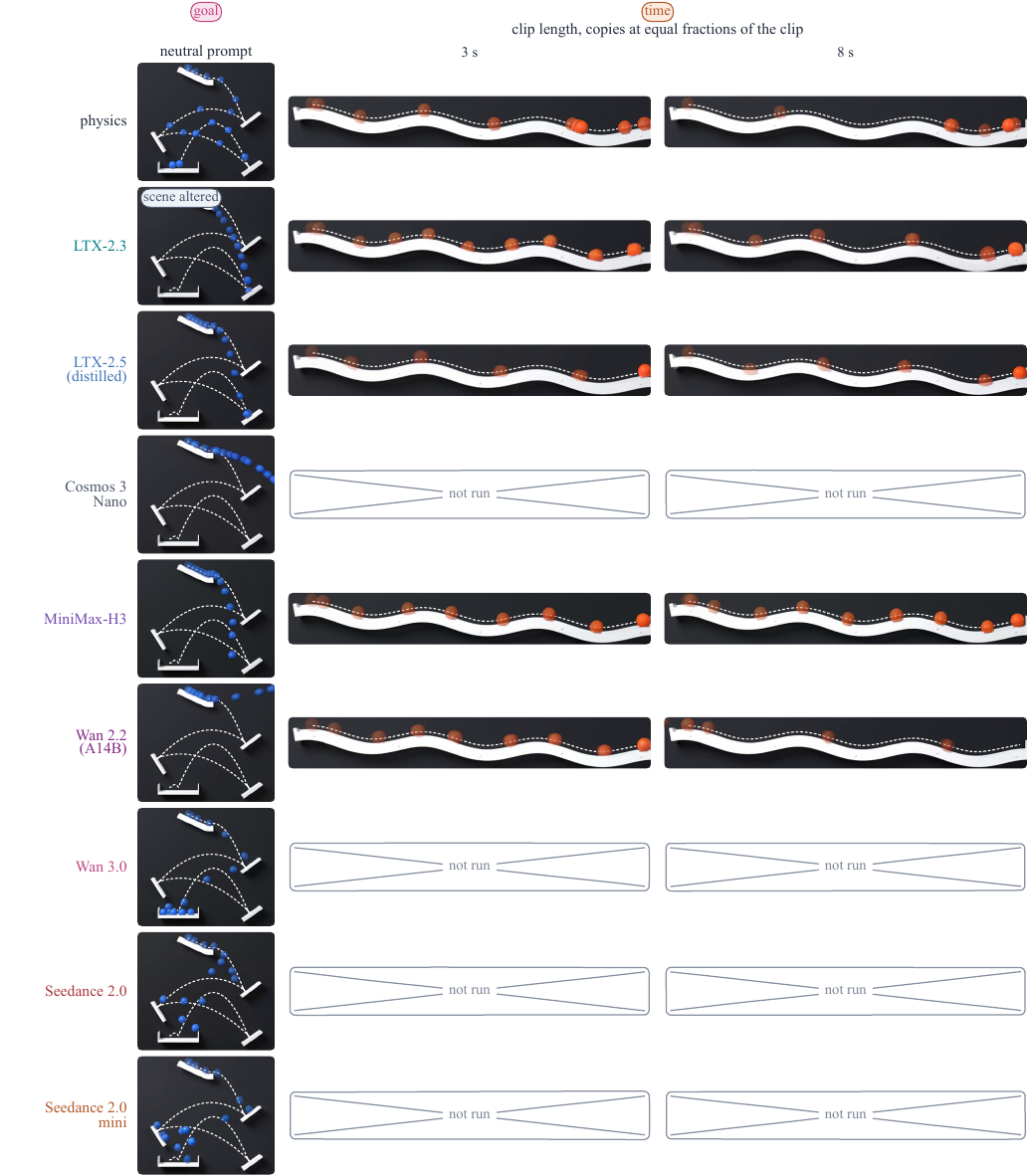}
\caption{\textbf{Interventions on goal and time.} Left: an irregular three-deflector scene under the prompt that leaves the outcome unnamed, from the prompt pairs, for the simulator and the eight models that ran them. Right: one single-frequency track at 3 and 8\,s for the four open-weight models of the clip-length arm, with the ball drawn at equal fractions of each clip, so that a motion scaled to the clip gives the same spacing at both lengths. The tagged cell (LTX-2.3) shows an altered scene.}
\label{fig:supp-cases-goal}
\end{figure*}

\begin{figure*}[p]
\centering
\includegraphics[width=\linewidth,height=7.9in,keepaspectratio]{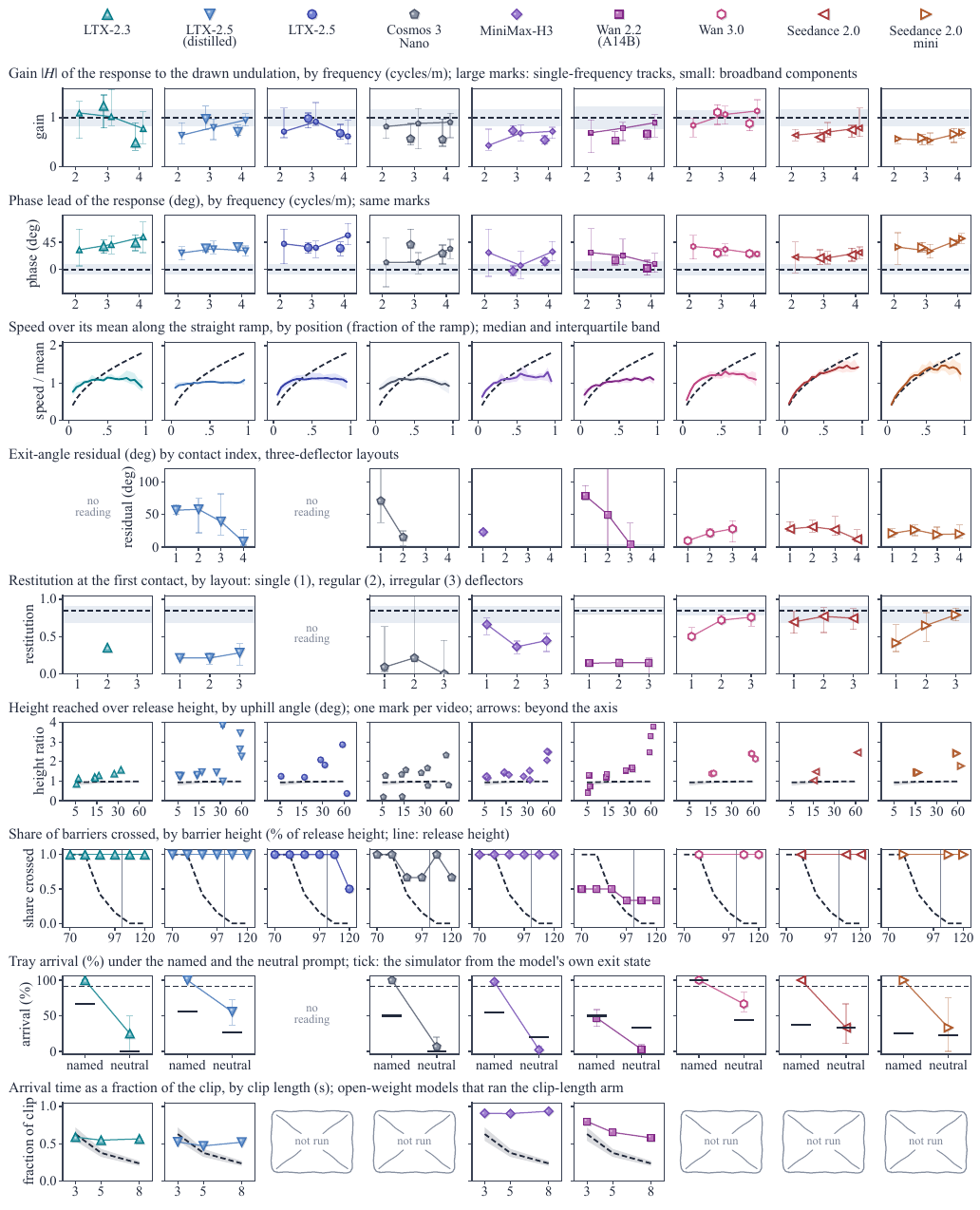}
\caption{\textbf{Per-model readings of every experiment.} One column per model and one row per reading, named above each row, with the physical value dashed. In the first two rows the grey band is the largest floor of the three undulating track levels at the model's frame rate. In the last row the physical band is the track-family transit time (1.62--2.15\,s) divided by the clip length, dashed at its midpoint.}
\label{fig:supp-perarm}
\end{figure*}

\clearpage

\section{Statistics}
\label{sec:supp-statistics}

The unit of every analysis is a scene with its seeds, and every interval comes from resampling scenes.
Paired designs remove what two videos share, permutation tests answer ordering questions, and a reading counts as a property of a model only when its interval clears the floor at the model's frame rate and output size.

\subsection{Units and intervals}

Per-video readings are summarized by their median over the estimable videos of a cell, phases by their circular mean with the resultant length $R$, and binary outcomes (exact order, tray arrival) by their share.
Intervals are 95\% percentile intervals from 2000 bootstrap resamples of scenes, in which the seeds of a scene are drawn together, so that the spread between scenes, the larger source of variation, enters every interval.
Within a video, the gain and phase come from a regression whose coefficients carry a Newey--West covariance.
Ratios of coefficients have Fieller intervals, and the gain, phase and $2f/1f$ intervals come from 4000 draws of the coefficients.
Every table carries its denominators (Sec.~\ref{sec:supp-integrity-rule}), so that small cells can be read as such.
Closed models contribute three to six scenes to each intervention family, and their readings are stated as intervals over those scenes.

\subsection{Paired designs}

\paragraph{Same seed.}
For the open-weight models the arms of an intervention share scene and seed, so each pair differs only in the changed element.
Local edits pair every arm with the base video of the same track and seed and take the median over pairs, with a bootstrap over tracks.
The rerun arm gives the sampler's own floor, which is zero for every open-weight model except LTX-2.3.
Prompt pairs match scene, first frame and seed, and differences in arrival are paired.
The steering compares the two prompts through their means per scene, since the videos with a wrong route differ between prompts.
Capture contrasts pair the in-tray and beyond scenes of the same base and seed, with a bootstrap over base scenes.
Clip-length differences (3 minus 5\,s, 8 minus 5\,s) are paired by scene and seed, and the duration exponent is the least-squares slope of $\log\alpha$ on $\log T_{\mathrm{clip}}$ after removing the mean of each scene and seed, with a bootstrap over scenes.

\paragraph{Independent samples.}
Closed models do not reproduce a seed, so an intervention compares the mean of an arm with the mean of the base over the same scenes, and the difference carries the variation between samples, including their tempo.
Photographed first frames compare each photograph row with its control row per configuration, with a bootstrap over configurations and the seed-spread test of Sec.~\ref{sec:supp-res-photo}, in which $s$ pools the seed deviations of the model around the centres of all its rows.
The destination counterfactual compares the model's arrival with the arrival of the stated law from the same exit states, with a bootstrap over scenes.

\subsection{Permutation and stratified resampling}

On chutes the four curvatures of one seed are exchangeable units for an open-weight model, so the $p$-value of the landing slope comes from permuting the curvature labels within each seed (across all videos for closed models).
The interval in Table~\ref{tab:supp-chute-results} resamples seeds within each curvature and leaves out the spread between chutes, so it is narrower, and the permutation $p$ carries the decision.
On ramp pairs and barriers the intervals resample seeds within each scene and keep the design of angles and heights.

\subsection{Reading intervals against floors}

A reading is interpreted as a property of a model when its interval lies outside the floor at the model's frame rate and size (Sec.~\ref{sec:supp-floors}).
Single-pair floors of the local edits (2.4--5.6\% in $dv$) are of the size of the dip effect itself, so these readings rest on medians over about 36 pairs, whose error on the simulated motion stays within a few tenths of a percent.
The null edit is flagged when the interval of its median difference from the base lies wholly beyond the larger of the rerun floor and the measurement floor (1.0\% for $dv$, 0.4\% for $dt$), a rule that flags none of 120 noise-only runs and all 120 runs in which the null arm carries the bump response.
Clip-length differences are read as the larger of $|3\,\mathrm{s}-5\,\mathrm{s}|$ and $|8\,\mathrm{s}-5\,\mathrm{s}|$ over its floor.
Readings marked exploratory (the locked and fixed parts of the shared-front response, the front and back leads, the relaxation rate in physical time, the decomposition of regenerated motion, and the mechanism predictions on other families) were added after the design was fixed.
They inform the interpretation and carry no claim on their own.

\end{document}